\documentclass[pdflatex,sn-mathphys-num]{sn-jnl}

\usepackage{graphicx}%
\usepackage{multirow}%
\usepackage{amsmath,amssymb,amsfonts}%
\usepackage{amsthm}%
\usepackage{mathrsfs}%
\usepackage[title]{appendix}%
\usepackage{xcolor}%
\usepackage{textcomp}%
\usepackage{manyfoot}%
\usepackage{booktabs}%
\usepackage{array}
\usepackage{algorithm}%
\usepackage{algorithmicx}%
\usepackage{algpseudocode}%
\usepackage{listings}%
\usepackage{subcaption} 
\usepackage{float}
\theoremstyle{thmstyleone}%
\theoremstyle{thmstyletwo}%

\theoremstyle{thmstylethree}%

\begin{document}

\title[Article Title]{Balancing Performance and Fairness in Explainable AI for Anomaly Detection in Distributed Power Plants Monitoring}


\author*[1]{\fnm{Corneille} \sur{Niyonkuru}}\email{corneille.niyonkuru@aims.ac.rw}

\author*[2,3]{\fnm{Marcellin} \sur{Atemkeng}}\email{m.atemkeng@gmail.com}

\author[4]{\fnm{Gabin Maxime} \sur{Nguegnang}}\email{maxime.nguegnang@aims-cameroon.org }

\author[5]{\fnm{Arnaud} \sur{Nguembang Fadja}}\email{arnaud.nguembafadja@unife.it }



\affil[1]{\orgdiv{}\orgname{African Institute for Mathematical Sciences (AIMS)}, \orgaddress{\city{Kigali}, \country{Rwanda}}}

\affil[2]{\orgdiv{Department of Mathematics}, \orgname{Rhodes University}, \orgaddress{\street{PO Box 94}, \city{Makhanda}, \postcode{6140}, \country{South Africa}}}

\affil[3]{\orgdiv{National Institute for Theoretical and Computational Sciences (NITheCS)}, \orgaddress{\city{Stellenbosch}, \postcode{7600}, \country{South Africa}}}

\affil[4]{\orgdiv{Department of Mathematics}, \orgname{Ludwig Maximilian University of Munich}, \orgaddress{\street{Bavaria}, \postcode{80333}, \country{Germany}}}

\affil[5]{\orgdiv{
Department of Engineering}, \orgname{University of Ferrara}, \orgaddress{\street{Via Saragat 1}, \city{Ferrara}, \postcode{44122}, \country{Italy}}}

\abstract{Reliable anomaly detection in distributed power plant monitoring systems is essential for ensuring operational continuity and reducing maintenance costs, particularly in regions where telecom operators heavily rely on diesel generators. However, this task is challenged by extreme class imbalance, lack of interpretability, and potential fairness issues across regional clusters. In this work, we propose a supervised ML framework that integrates ensemble methods (LightGBM, XGBoost, Random Forest, CatBoost, GBDT, AdaBoost) and baseline models (Support Vector Machine, K-Nearrest Neighbors, Multilayer Perceptrons, and Logistic Regression) with advanced resampling techniques (SMOTE with Tomek Links and ENN) to address imbalance in a dataset of diesel generator operations in Cameroon. Interpretability is achieved through SHAP (SHapley Additive exPlanations), while fairness is quantified using the Disparate Impact Ratio (DIR) across operational clusters. We further evaluate model generalization using Maximum Mean Discrepancy (MMD) to capture domain shifts between regions. Experimental results show that ensemble models consistently outperform baselines, with LightGBM achieving an F1-score of 0.99 and minimal bias across clusters (DIR $\approx 0.95$). SHAP analysis highlights fuel consumption rate and runtime per day as dominant predictors, providing actionable insights for operators. Our findings demonstrate that it is possible to balance performance, interpretability, and fairness in anomaly detection, paving the way for more equitable and explainable AI systems in industrial power management. {\color{black} Finally, beyond offline evaluation, we also discuss how the trained models can be deployed in practice for
real-time monitoring. We show how containerized services can process in real-time, deliver low-latency predictions, and provide interpretable outputs for operators.

}
}




\maketitle

\section{Introduction}
\label{sec:introduction}

Power generation systems form a backbone of modern telecommunications, especially in regions with unreliable electricity grids. In developing countries such as Cameroon, telecom operators rely extensively on diesel generators as off-grid backups for base stations, often complementing them with renewable energy sources such as solar or wind to ensure continuous service during grid failures \citep{frenger2011reducing, sarwar2022optimal}. Despite their importance, these hybrid systems operate under heterogeneous environmental and load conditions, making them susceptible to anomalies ranging from mechanical faults and fuel misuse to data recording errors \citep{chandola2009anomaly}. If undetected, such anomalies may cause service disruptions, escalate maintenance costs, and contribute to environmental damage. For instance, inefficiencies in a single generator network can result in the unnecessary waste of thousands of liters of diesel annually \citep{ullah2024optimal}.  

These operational challenges are exemplified by the \textit{TeleInfra Ltd. dataset}, which records 8,479 samples of generator activity across Cameroon (September 2017–September 2018). The dataset comprises over 7,100 normal instances and three categories of anomalies: fuel consumption with zero runtime, runtime exceeding 24 hours/day, and excessive fuel consumption. The rarity and heterogeneity of these anomalies underscore the difficulty of detecting subtle, context-specific irregularities in high-dimensional industrial data \citep{mulongo2020anomaly, atemkeng2024ensemble}. Conventional strategies, such as manual inspections or static thresholding, frequently underperform, generating false alarms or overlooking anomalies, thereby straining maintenance resources and diminishing trust in automated systems \citep{zhu2025anomaly}.  

Machine learning (ML) provides a promising avenue for anomaly detection but faces three persistent challenges. First, severe class imbalance—where anomalies constitute only a small fraction of the dataset—biases models toward majority classes and reduces sensitivity to rare but critical events \citep{haixiang2017learning}. Second, the lack of interpretability in many ML models results in opaque predictions that offer limited operational value to human experts \citep{arrieta2020explainable}. Third, regional variability in generator usage patterns (driven by climatic, load, and maintenance differences) hampers model generalization, while uneven anomaly detection rates across clusters raise fairness concerns \citep{babu2024enhancing}. These challenges motivate the development of a detection framework that is not only accurate, but also interpretable and equitable.  

To address these issues, we propose a framework that integrates ensemble learning with advanced data balancing methods, specifically Synthetic Minority Over-sampling Technique (SMOTE) combined with Tomek Links and Edited Nearest Neighbors (ENN), to alleviate class imbalance \citep{Fernández2018}. Model interpretability is enhanced through SHAP (SHapley Additive exPlanations), which provides feature-level explanations, while fairness across regional clusters is assessed using the Disparate Impact Ratio (DIR) \citep{lundberg2017unified, babu2024enhancing}. The framework evaluates two complementary modeling strategies: (1) \textit{cluster-specific models}, which capture localized operational patterns, and (2) \textit{global models}, which leverage aggregated data for broader applicability. Generalization across clusters is further quantified using cross-cluster validation and domain shift metrics such as Maximum Mean Discrepancy (MMD) \citep{pan2009survey}.  

The main contributions of this study are as follows:
\begin{itemize}
    \item Development of a hybrid data balancing strategy that improves anomaly detection under extreme class imbalance,  
    \item Integration of SHAP-based interpretability to provide model-agnostic explanations for anomaly predictions,  
    \item Incorporation of fairness constraints via DIR to mitigate regional bias in detection performance. 
\end{itemize}

The remainder of this article is organized as follows:
Section~\ref{sec:related_work} reviews related work on ML-based anomaly detection in industrial systems. 
Section~\ref{sec:dataset_methodology} presents the dataset and methodology, including preprocessing, model design, and evaluation metrics. 
Section~\ref{sec:results} reports the results and discussion, while Section~\ref{sec:deployment_in_practice} discusses the implications and deployment for industrial operators.

\section{Related Work}
\label{sec:related_work}

The field of anomaly detection has evolved significantly, driven by the increasing complexity of industrial systems and the availability of large-scale, data-driven approaches. In power generation systems, particularly telecom diesel generators, anomaly detection is critical for ensuring operational reliability, minimizing costs, and reducing environmental impact. This section reviews prior work in anomaly detection, focusing on traditional and ML-based approaches, their applications in power systems, and their limitations in addressing data imbalance, interpretability, fairness, and generalization. By synthesizing these studies, we highlight the gaps that motivate the proposed supervised ML  framework, which leverages ensemble learning to balance performance, interpretability, and fairness in telecom diesel generator operations.
 
Early anomaly detection methods in power systems relied on baseline techniques such as rule-based thresholding and statistical analysis. For instance, \citet{zhu2025anomaly} developed rule-based systems to detect anomalies in industrial control systems, using predefined thresholds for metrics like fuel consumption and runtime. These methods, while straightforward, struggled with high-dimensional and noisy data, often resulting in high false-positive rates or missed detections of context-dependent anomalies \citep{chandola2009anomaly}. Statistical approaches, such as those based on Gaussian mixture models or principal component analysis, improved detection by modeling normal behavior but faced challenges in capturing rare anomalies in imbalanced datasets, a common issue in power plant operations \citep{wei2015anomaly}. These limitations underscored the need for more adaptive, data-driven techniques capable of handling complex patterns in industrial settings.

The advent of ML has transformed anomaly detection by enabling the identification of intricate patterns in large datasets. Supervised learning methods, such as support vector machine (SVM) and logistic Regression (LR), have been applied to power systems with promising results. For example, \citet{chung2023chp} employed a K-Nearest Neighbors (KNN) model to detect anomalies in combined heat and power (CHP) engines, achieving an accuracy of 93\%. However, supervised methods often require labeled data, which is scarce for anomalies in industrial contexts, and struggle with imbalanced datasets, leading to biased predictions toward the majority (normal) class \citep{haixiang2017learning}. Unsupervised approaches, such as clustering or autoencoders, have been explored to address the lack of labeled data. \citet{kim2023abnormal} proposed a two-stage LSTM-Autoencoder framework for anomaly detection in hybrid power plants, combining feature engineering with unsupervised learning to identify deviations. While effective in controlled settings, unsupervised methods often produce high false-positive rates and lack interpretability, limiting their adoption in operational environments where actionable insights are critical \citep{farber2025unsupervised}.

Ensemble learning has emerged as a useful approach to address the shortcomings of individual models. Techniques like Random Forest (RF), Gradient Boosted Decision Trees (GBDT), and Extreme Gradient Boosting (XGBoost) have shown robustness in handling imbalanced datasets and noisy industrial data \citep{chen2016xgboost}. For instance, \citet{mulongo2020anomaly} applied RF to telecom diesel generator data, achieving improved detection rates for fuel-related anomalies but without addressing interpretability or fairness across regional clusters. Similarly, \citet{atemkeng2024ensemble} explored supervised ensemble methods on the TeleInfra Ltd. dataset, demonstrating high accuracy but limited generalization across diverse operational conditions. These studies highlight the potential of ensemble methods but often overlook critical aspects like transparency and equitable performance, which are essential for industrial deployment.

Interpretability has become a focal point in recent anomaly detection research, particularly in industrial applications where operators need to understand model decisions to guide maintenance. SHAP has gained traction as a tool for explaining feature contributions in complex models like XGBoost and Light Gradient Boosting Machine (LightGBM) \citep{lundberg2017unified}. For example, \citet{lundberg2017unified} demonstrated SHAP's ability to elucidate feature importance in predictive maintenance, enabling operators to prioritize interventions based on key indicators like runtime or fuel consumption. However, few studies have integrated interpretability into anomaly detection for power systems, leaving a gap in translating model outputs into actionable operational strategies.

Fairness is another underexplored dimension in anomaly detection, particularly in distributed systems like telecom generators, where regional variations in data distribution can lead to biased detection. \citet{wibowo2025enhancing} introduced the DIR to evaluate fairness in algorithmic predictions, highlighting disparities in performance across demographic or operational groups. In power systems, biased detection could result in certain regions receiving disproportionate attention, exacerbating inefficiencies in underrepresented clusters. While fairness metrics have been applied in fields like finance and healthcare, their use in industrial anomaly detection remains limited, underscoring a critical gap addressed in this study.

Domain adaptation and generalization are also key challenges in power system anomaly detection, as operational conditions vary across regions due to differences in climate, usage, or maintenance practices. \citet{pan2009survey} proposed MMD to quantify domain shifts, enabling models to adapt to new environments. However, most studies, including \citet{chung2023chp} and \citet{kim2023abnormal}, focus on single-site or controlled datasets, neglecting cross-regional generalization. This limitation is particularly relevant for telecom diesel generators, where regional clusters exhibit distinct operational patterns, as seen in TeleInfra Ltd. dataset \citep{atemkeng2024ensemble}.

The proposed framework builds on these prior efforts while addressing their limitations. Unlike baseline methods, it leverages supervised ensemble learning (LightGBM, XGBoost, RF, Categorical Boosting (CatBoost), and GBDT) and baseline models (SVM, KNN, Multilayer Perceptron (MLP), and LR) to handle data imbalance through advanced balancing techniques like SMOTE with Tomek Links and ENN \citep{Fernández2018}. It incorporates SHAP for interpretability, providing clear insights into feature contributions, and DIR for fairness, ensuring equitable detection across regional clusters \citep{lundberg2017unified, chhabra2021overview}. By comparing cluster-specific and global models and evaluating generalization using MMD, this work presents a comprehensive approach that balances performance, transparency, and equity, thereby advancing modern anomaly detection for power systems.

Unlike previous studies, our approach stands out by seamlessly integrating advanced data balancing, SHAP-based interpretability, and fairness assessment through the DIR, delivering a robust solution that achieves high detection recall while ensuring transparency and equity across diverse operational clusters. This study addresses critical gaps by employing hybrid resampling techniques to tackle class imbalance effectively, leveraging SHAP to provide actionable, feature-level insights that empower operators with clear decision-making tools, and utilizing MMD to enhance model generalization across varying regional conditions. By bridging these dimensions; recall, explainability, and fairness,our work sets a new benchmark for anomaly detection in industrial power systems, offering a holistic and equitable advancement over existing methodologies.

\section{Dataset and Methodology}
\label{sec:dataset_methodology}

\subsection{Data Description}
\label{subsec:data_description}

The dataset, sourced from TeleInfra Ltd., covers diesel generator operations at telecom base stations in Cameroon from September 2017 to September 2018, comprising 8,479 samples after cleaning the dataset across regional clusters such as KOUSSERI, MAROUA-1, NGAOUNDERE 1, TIBATI, and MAKARY \citep{atemkeng2024ensemble}. It is described in Table \ref{tab:generator_data_features}, captures operational metrics critical for anomaly detection. It includes both raw features, such as fuel consumption (liters) and running time (hours), and derived features such as consumption rate (liters per day). Within this data, anomalies are categorized into three classes. A significant class imbalance is present, with over 7,100 normal samples and anomalies constituting less than 15\% of the dataset. Furthermore, the existence of regional clusters reflects variations in operational conditions, necessitating models that can balance localized accuracy with cross-cluster generalization.

\begin{table}[h]
\centering
\scriptsize 
\caption{Description of the different features in the dataset.}
\label{tab:generator_data_features}
\begin{tabular}{>{\raggedright\arraybackslash}p{6cm}>{\raggedright\arraybackslash}p{9cm}}
\toprule
\textbf{Column Name} & \textbf{Description} \\
\midrule
CONSUMPTION HIS & The total amount of fuel consumed during a specific period before the next refuelling occurs. \\
CONSUMPTION\_RATE & The hourly fuel consumption of the generator, measured in litres. \\
Cluster & The geographic locations where the generator sites are situated (cities). \\
CURRENT HOUR METER GE1 & The current reading of the hour meter on the generator, indicating total operational hours. \\
Site Name & The designated name for each generator location. \\
EFFECTIVE\_DATE\_OF\_VISIT & The date on which the fuel meter reading, refuelling, and other data entries are recorded. \\
PREVIOUS\_DATE\_OF\_VISIT & The date of the generator's last visit and corresponding data collection. \\
Months & The month during which the data reading was conducted. \\
NUMBER\_OF\_DAYS & The interval (in days) before the subsequent refuelling is mandated. \\
GENERATOR\_1\_CAPACITY\_(KVA) & The rated capacity of the generator measured in kilovolt-amperes (KVA). \\
POWER TYPE & The category of energy source employed by the power generation system. \\
PREVIOUS HOUR METER G1 & The prior hour meter reading from the generator. \\
PREVIOUS\_FUEL\_QTE & The total quantity of fuel available in the generator tank at the time of the last visit. \\
QTE\_FUEL\_FOUND & The amount of fuel present in the tank before refuelling. \\
QTE\_FUEL\_ADDED & The volume of fuel added to the generator during the refuelling process. \\
TOTALE\_QTE\_LEFT & The remaining fuel quantity in the generator after the refuelling operation. \\
RUNNING\_TIME & The cumulative hours the generator has operated up until the next refuelling. \\
Running\_time\_per\_day & The total operational hours of the generator within a single day. \\
Consumption\_per\_day\_within\_a\_period & The amount of fuel consumed in a designated daily period. \\
Fuel\_consumed\_between\_visits\_per\_day & The quantity of fuel consumed between successive visits within a single day. \\
Fuel\_consumed\_between\_visits & The total fuel consumed from all visits cumulatively. \\
Maximum\_consumption\_per\_day & The maximum amount of fuel that the generator is capable of consuming in one day. \\
\bottomrule
\end{tabular}
\end{table}

\subsection{Data Preparation and Processing}
\label{subsec:data_preparation&processing}

Data preparation is critical to ensure the dataset is suitable for ML, addressing noise, missing values, and class imbalance while preserving operational patterns. The process involved systematic steps to clean, transform, and balance the data, ensuring robust model training and evaluation across various clusters.

\subsubsection{Feature Engineering}
\label{subsubsec:feature_engineering}

Feature engineering enhanced the model's ability to detect anomalies by deriving new features from raw data. Key engineered features included consumption rate (fuel consumed per day), which normalized usage to identify Class 3 anomalies (excessive consumption), and runtime per day, which highlighted operational excesses for Class 2 anomalies (runtime exceeding 24 hours). These features are listed in the last five rows in the Table \ref{tab:generator_data_features}, which were informed by domain knowledge of telecom generator operations and exploratory analysis, ensuring that they captured inefficiencies without introducing redundancy \citep{bishop2006pattern}. For instance, consumption rate was critical for clusters like KOUSSERI, where environmental factors increased anomaly rates, while runtime per day aided in detecting mechanical issues in clusters like MAROUA-1. This step improved the model's sensitivity to context-dependent anomalies across diverse operational conditions.

\subsubsection{Class Labeling}
\label{subsubsec:class_labeling}

\begin{figure}[H]
    \centering
    \includegraphics[width=0.7\linewidth]{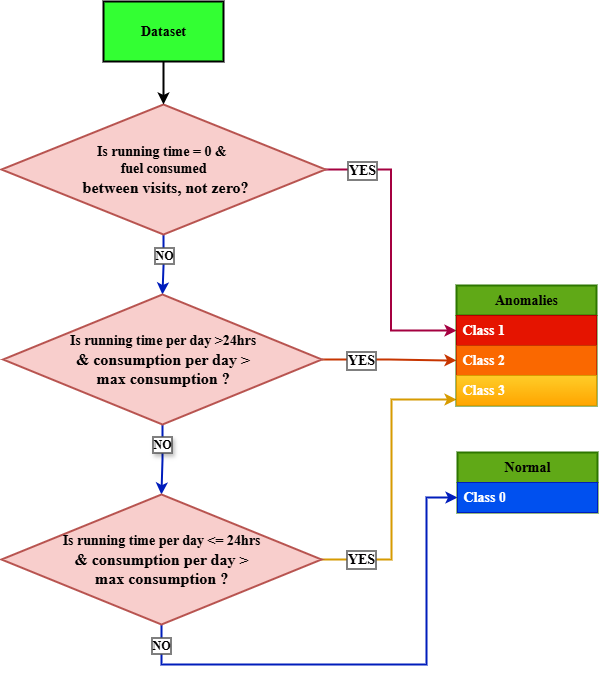}
    \caption{This flowchart illustrates a data labeling process for classifying anomalies in a dataset. It evaluates conditions related to running time and fuel consumption to categorize data points into four classes: Class 1, Class 2, Class 3 (anomalies), or Class 0 (normal).}
    \label{fig:Data Labeling Diagram}
\end{figure}

\begin{figure}[H]
    \centering
    \includegraphics[width=0.75\linewidth]{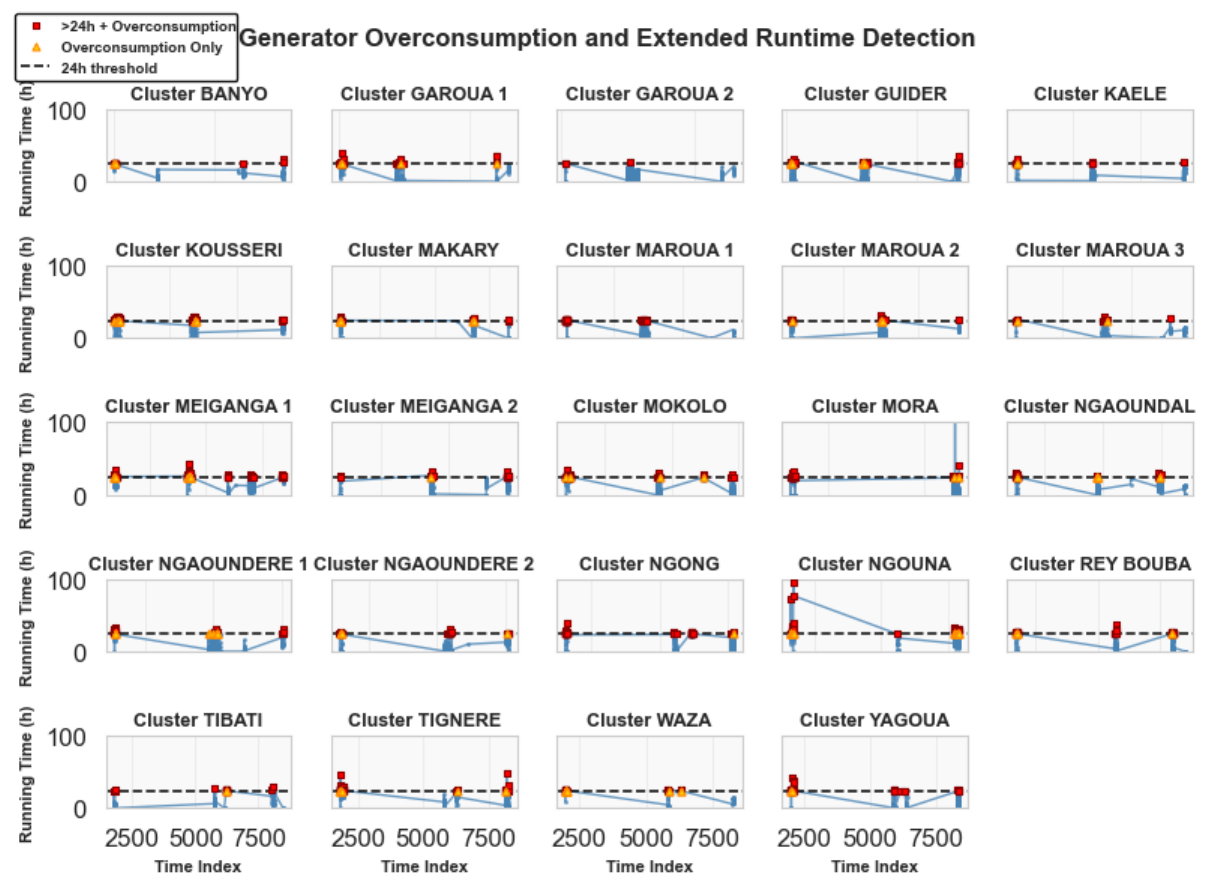}
    \caption{A visualization of generator performance anomalies across multiple clusters, highlighting instances of extended runtime (>24 hours) and fuel overconsumption.}
    \label{fig: Anomaly_Detection_Type_2_3_Only}
\end{figure}

The class labeling process systematically categorized data points into normal and anomalous classes, with anomalies further divided into three types, as illustrated in Figure \ref{fig:Data Labeling Diagram}. Class 1 anomalies were identified when fuel consumption occurred with zero runtime, suggesting misuse or data entry errors. Class 2 anomalies were flagged for runtime exceeding 24 hours per day, indicating potential mechanical faults or logging inaccuracies. Class 3 anomalies were defined by excessive fuel consumption beyond cluster-specific thresholds, pointing to inefficiencies or equipment degradation. Figure \ref{fig:Data Labeling Diagram} depicts the decision tree used for labeling, ensuring reliable and consistent categorization for supervised learning. These classes are illustrated in Figure~\ref{fig: Anomaly_Detection_Type_2_3_Only}, which shows examples where the generator operated beyond the 24-hour threshold or consumed fuel beyond the expected maximum.

\begin{figure}[H]
    \centering
    \includegraphics[width=0.75\linewidth]{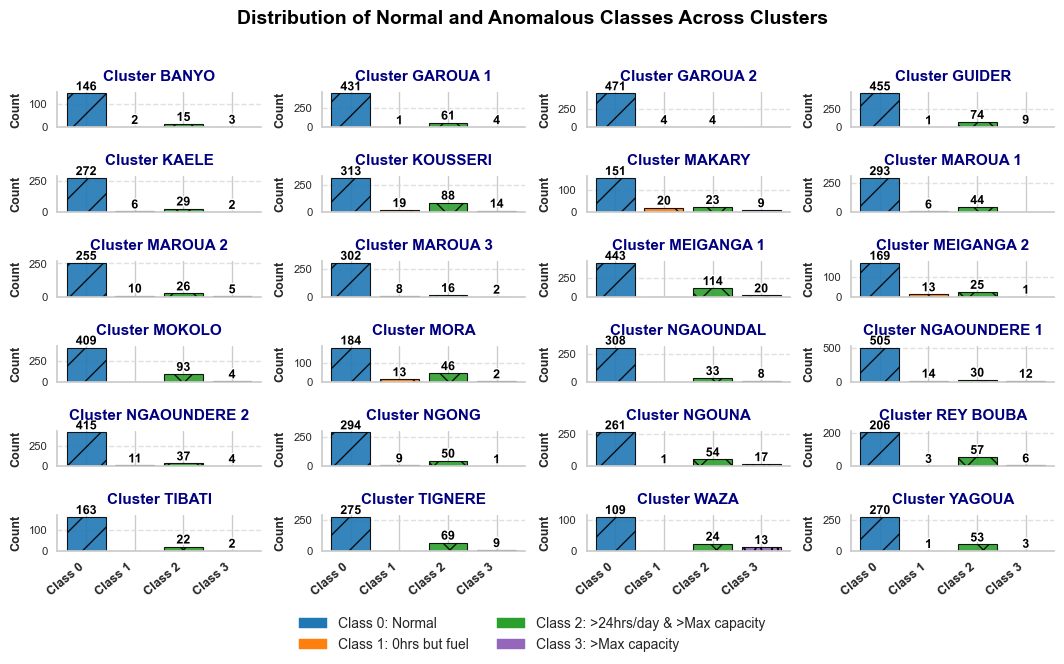}
    \caption{Frequency distribution of normal (Class 0) and anomalous (Classes 1-3) generator events across multiple regional clusters.}
    \label{fig: normal_anomaly_distribution}
\end{figure}

In addition to the overall dataset, the data was partitioned by cluster to examine the distribution of anomaly classes across different locations. This subdivision helps identify which types of anomalies are present in each cluster. Figure~\ref{fig: normal_anomaly_distribution} shows that clusters such as GAROUA 2, MAROUA 1, MEIGANGA 1, MOKOLO, NGAOUNDAL, TABITA, TIGNERE, and WAZA do not exhibit all anomaly classes. This analysis can guide researchers in focusing their efforts on specific clusters based on the anomalies they exhibit.

Normal samples dominated the dataset (over 85\%), with anomalies comprising less than 15\%, posing challenges for model training due to class imbalance. Cluster-specific distributions varied significantly: KOUSSERI exhibited higher anomaly rates (approximately 20\% Class 3) due to environmental stressors, while MAROUA-1 showed greater stability (approximately 10\% anomalies). To address the imbalance, data balancing is applied to balance the training set, preserving data integrity while enhancing minority class representation \citep{Fernández2018}. These insights guided the selection of evaluation metrics like F1-score and the adoption of both global and cluster-specific modeling approaches.

\subsubsection{Feature Importance}
\label{subsubsec:feature_importance}

\begin{figure}[H]
    \centering
    \includegraphics[width=0.75\linewidth]{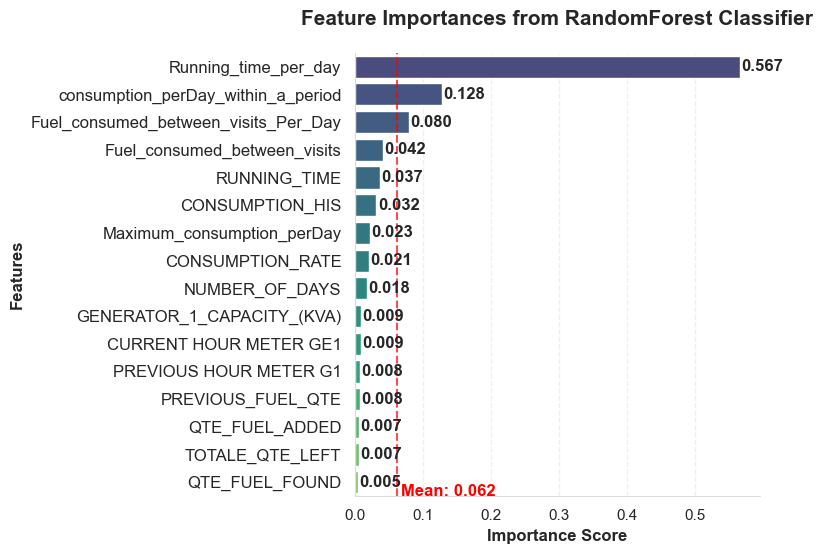}
    \caption{Feature importance from a RF classifier, showing \emph{Running time per day} and \emph{consumption per day within a period} as the most critical predictors for the model's decisions.}
    \label{fig:Feature Importance for RF Classifier}
\end{figure}

Feature importance analysis is conducted to prioritize the most impactful features, utilizing the RF Classifier due to its robust ability to quantify feature relevance through impurity reduction (e.g., Gini index) across decision trees \citep{chen2016xgboost}. The RF Classifier was selected for its effectiveness in handling noisy, high-dimensional industrial data and providing clear feature significance insights. Engineered features, such as consumption rate and runtime per day, were anticipated to be the most influential for detecting Class 3 (excessive fuel consumption) and Class 2 (runtime exceeding 24 hours) anomalies, respectively, compared to raw features like fuel consumed, which lack contextual normalization. The analysis, planned for visualization in Figure \ref{fig:Feature Importance for RF Classifier}, was conducted for both global and cluster-specific models to capture regional variations, such as those in KOUSSERI and MAROUA-1. This process aimed to streamline the feature set, reducing model complexity while enhancing interpretability for operators to make informed maintenance decisions \citep{lundberg2017unified}.

\subsubsection{Distribution of Data}
\label{subsubsec:distribution_data}

\begin{figure}[H]
    \centering
    \includegraphics[width=0.75\linewidth]{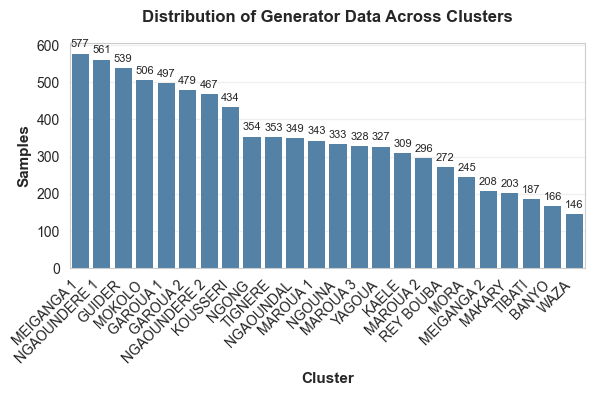}
    \caption{Distribution of Generator Data Across Clusters: A bar chart showing a skewed distribution of samples, with MEIGANGA 1 (577) and NGOUNDERI (539) having the highest counts, followed by a gradual decline to BANYO and WAZA (~146-166)}
    \label{fig: Distribution of Data Across Clusters}
\end{figure}

Understanding the distribution of data across clusters is essential for model development. Figure~\ref{fig: Distribution of Data Across Clusters} illustrates how the data is spread among different clusters. This information is crucial for evaluating each cluster's contribution to the full dataset, thereby helping to determine whether global or cluster-specific models are more appropriate. This is particularly evident when comparing clusters; for example, MEIGANGA 1 leads with 577 samples, whereas WAZA contributes the least with 146. It also aids in selecting models that will likely yield the best performance based on cluster characteristics.

\subsection{Correlation}
\label{subsec:correlation}

\begin{figure}[H]
    \centering
    \includegraphics[width=1\linewidth]{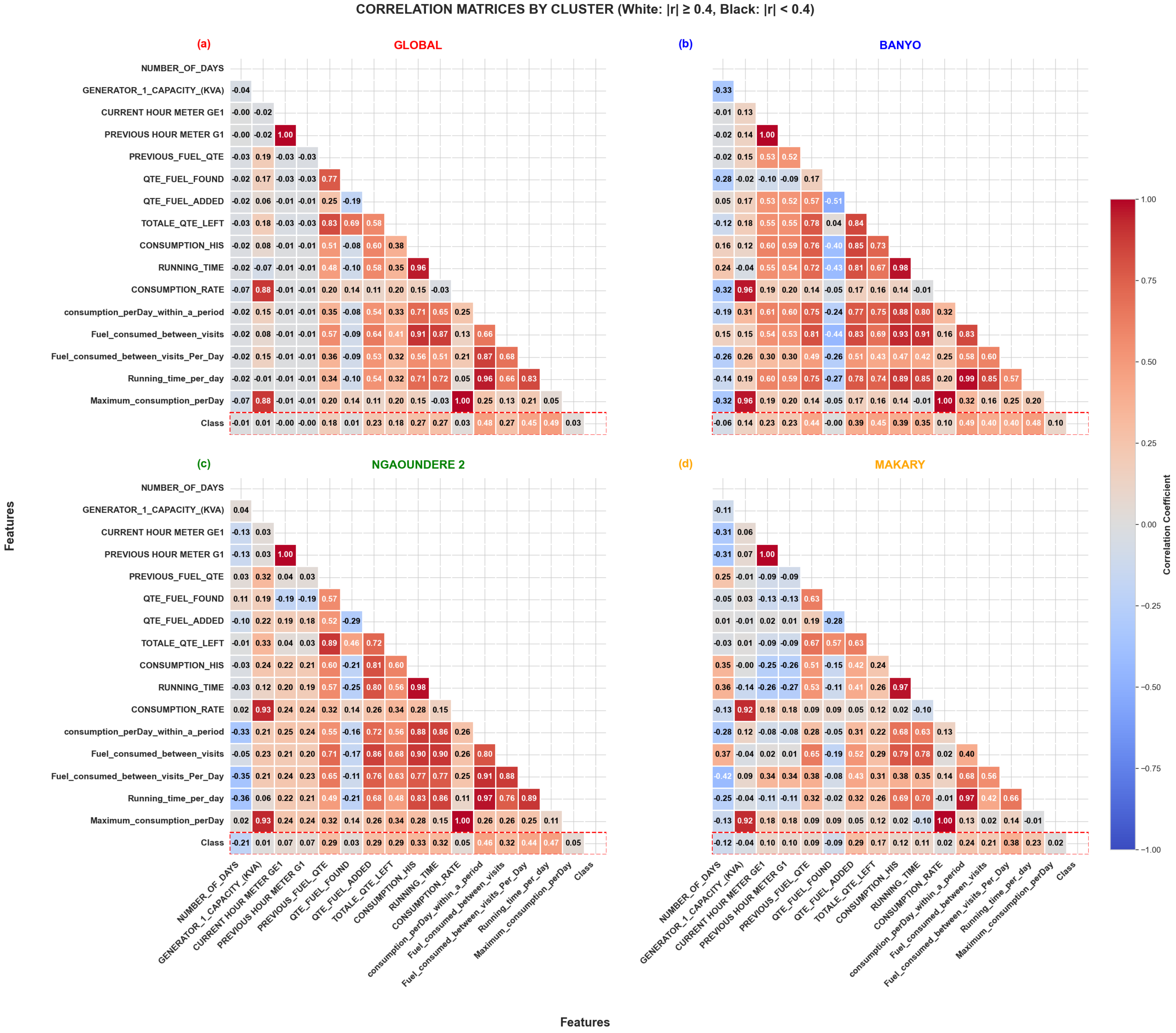}
    \caption{Correlation matrices for the global dataset and clusters representing high, medium, and low correlation with the target variable. (a) GLOBAL dataset, (b) BANYO (high correlation), (c) NGAOUNDERE 2 (medium correlation), (d) MAKARY (low correlation). Strong correlations ($|r| \geq 0.4$) are shown in white.}
    \label{fig:correlation_matrices_clusters}
\end{figure}

Correlation analysis was performed to examine the relationships between numerical features both across the entire dataset and within individual clusters. From the 24 clusters, three were selected based on the extent of correlation their features exhibit with the target variable (class). These clusters exemplify the range from high representation (featuring numerous strong correlations), through medium representation (featuring a moderate level of correlation), to low representation (featuring few or weak correlations), as shown in Figure \ref{fig:correlation_matrices_clusters}. In the global analysis (Fig.~\ref{fig:correlation_matrices_clusters} a), key relationships include a strong positive correlation between fuel consumed between visits and consumption rate (r = 0.87), emphasizing its role in Class 3 anomalies related to excessive fuel use, and a moderate positive correlation between running time and consumption per day within a period (r = 0.65), which supports detection of Class 2 anomalies tied to prolonged operation. Additionally, maximum consumption per day shows strong correlations with fuel consumed between visits per day (r = 0.45) and running time per day (r = 0.49), indicating consistent patterns in peak usage.

Cluster-specific matrices reveal distinct dynamics. In the BANYO cluster (Fig.~\ref{fig:correlation_matrices_clusters} b), a moderate positive correlation exists between QTE fuel added and total QTE left (r = 0.84), suggesting aligned fuel replenishment practices, while consumption rate correlates moderately with consumption per day within a period (r = 0.32). The NGAOUNDERE 2 cluster (Fig.~\ref{fig:correlation_matrices_clusters} c) displays a very strong positive correlation between current hour meter GE1 and previous hour meter G1 (r = 0.1), reflecting reliable sequential readings, and between QTE fuel found and QTE fuel added (r = 0.72), pointing to effective fuel monitoring. In the MAKARY cluster (Fig.~\ref{fig:correlation_matrices_clusters} d), notable correlations include a strong negative link between previous fuel QTE and QTE fuel found (r = 0.63), potentially indicating irregular fuel depletion, and a low correlation between fuel found and consumption rate (r = 0.09), which may highlight inefficiencies or anomalies in fuel usage. These tailored insights enhance anomaly detection by accounting for cluster-unique feature interdependencies.
 
\subsection{Methodology}
\label{subsec:methodology}

\begin{figure}[H]
    \centering
    \includegraphics[width=.75\linewidth]{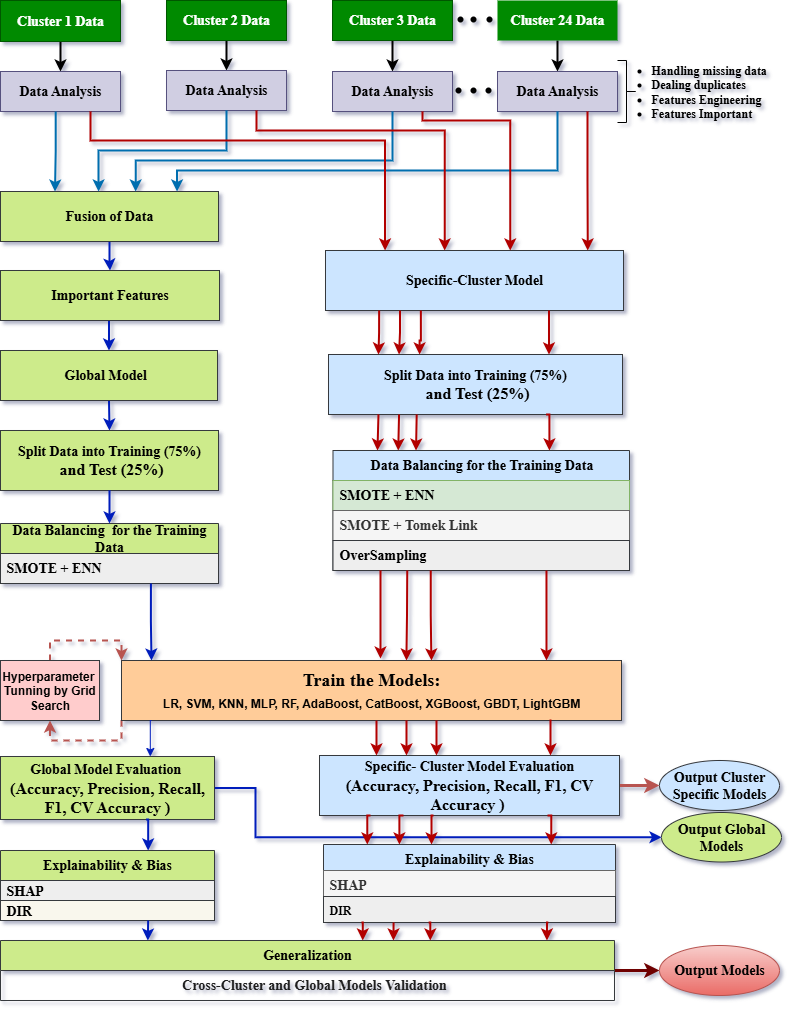}
    \caption{A structured methodology for training both a global models and cluster-specific models, featuring data fusion, class balancing with SMOTE, multi-algorithm training, and evaluation for accuracy, explainability (SHAP), and bias (DIR).}
    \label{fig:MethodologyStracture}
\end{figure}

The methodology leverages a supervised ML framework, using baseline ML and ensemble learning to achieve robust anomaly detection while addressing data imbalance through advanced balancing techniques like SMOTE with Tomek Links and ENN~\cite{altalhan2025imbalanced}. The process, as illustrated in Figure~\ref{fig:MethodologyStracture}, integrates data preprocessing (cleaning, feature engineering, and balancing), model development (global and cluster-specific strategies), and evaluation (performance metrics, generalization, interpretability, and fairness). This workflow ensures the framework's applicability to telecom diesel generators, where regional cluster variations require adaptable models that not only effectively detect anomalies but also deliver practical value by reducing maintenance costs and operational disruptions.

\subsubsection{Model Development}
\label{subsec:model_development}

The framework employs a range of models, including LR, SVM, KNN, MLP, and ensemble methods such as Adaptive Boosting (AdaBoost), CatBoost, XGBoost, GBDT, RF, and LightGBM. Ensemble methods are favored for their ability to handle imbalanced and noisy datasets, which feature over 85\% normal samples and sparse anomalies across three classes. The approach adopts two strategies: cluster-specific models, tailored to individual clusters to reflect localized patterns influenced by factors such as climate or maintenance practices, and global models, designed to generalize across all clusters using aggregated data. Hyperparameter optimization is performed using grid search, adjusting parameters such as learning rate (0.01--0.1) and number of estimators (50--200) for LightGBM, and tree depth (10--30) and number of trees (100--500) for RF. The framework supports multi-class classification (normal vs. Classes 1, 2, 3), offering flexibility for operational needs.

These models offer distinct advantages in anomaly detection for distributed systems, where robustness against class imbalance and noise is crucial. LR excels in binary classification tasks by modeling linear relationships in the data, thereby enhancing detection reliability for straightforward anomaly indicators \cite{chen2016xgboost}. SVMs complements this, fortifying decision boundaries to effectively separate normal instances from sparse anomalies, even in high-dimensional spaces. KNN uncovers subtle local patterns through distance-based similarity, proving particularly useful for cluster-specific variations influenced by regional factors \cite{chen2016xgboost}. For more intricate dependencies, MLPs amplify predictive accuracy via layered neural architectures that capture non-linear interactions in complex, noisy datasets \cite{mienye2024comprehensive}. Among the ensemble techniques, AdaBoost elevates the performance of base learners by iteratively prioritizing misclassified anomalies, addressing imbalance without extensive resampling \cite{chen2016xgboost}, whereas CatBoost streamlines handling of categorical features inherent in maintenance logs, yielding unbiased predictions across diverse clusters \cite{prokhorenkova2018catboost}. XGBoost advances scalability and precision through regularized gradient boosting that mitigates overfitting in sparse data environments \cite{chen2016xgboost}, and GBDT refine sequential predictions to progressively reduce errors from noisy inputs \cite{chen2016xgboost}. RF further bolster stability by aggregating diverse decision trees, which collectively dampen variance and enhance generalization amid over 85\% normal samples \cite{chen2016xgboost}, with LightGBM accelerating the entire process via efficient histogram-based learning that maintains high accuracy on large-scale, imbalanced datasets \cite{ke2017lightgbm}. Together, these complementary capabilities ensure the framework's models not only detect anomalies with high fidelity but also adapt seamlessly to the operational variability across clusters.

The model development process, as depicted in Figure~\ref{fig:MethodologyStracture}, is designed to systematically build reliable detection capabilities by starting with data analysis and fusion from multiple clusters. This step is vital because it creates a comprehensive dataset that incorporates diverse operational environments, thereby preventing models from overfitting to isolated conditions and enhancing overall robustness in real-world distributed systems. Following this, important features are identified to streamline the input space, which is essential for mitigating computational overhead and concentrating on high-impact variables like fuel consumption rate. The data is split into training (75\%) and test (25\%) sets to facilitate unbiased validation, ensuring that performance metrics reflect true generalizability rather than inflated training artifacts. Data balancing on the training data is achieved using SMOTE with ENN or Tomek Links, effectively addressing class imbalance. This step is essential, as anomalies constitute only a small minority of the data. Without this approach, models would be biased toward normal instances, potentially leading to missed detections. Such oversights could result in costly downtime or fuel waste. Hyperparameter tuning via grid search optimizes each model's configuration, adapting to the dataset's unique challenges like high dimensionality and regional variability, which ultimately yields higher accuracy and efficiency. The trained global and cluster-specific models are then subjected to rigorous evaluation for performance (e.g., precision, recall, F1), interpretability, and fairness; this integrated assessment guarantees not just detection efficacy but also actionable insights and equitable outcomes, directly contributing to reduce maintenance costs and fair resource allocation across clusters.

\subsection{Performance Evaluation}
\label{subsec:performance_evaluation}

The model performance evaluation ensures the framework's robustness for anomaly detection generator operations. Key metrics are used to assess performance, with a focus on high recall to minimize undetected anomalies that could escalate operational costs or service disruptions \cite{bishop2006pattern}. These metrics are summarized in Table~\ref{tab:eval_metrics}, providing a balanced view of the model's ability to correctly identify anomalies while controlling for false positives, which is crucial in industrial settings where false alarms strain resources. Cross-cluster validation tests generalization by training on one cluster and testing on others, revealing how models adapt to diverse conditions; MMD, defined as
\[
\text{MMD}^2(P, Q) = \mathbb{E}[k(x, x')] - 2\mathbb{E}[k(x, y)] + \mathbb{E}[k(y, y')],
\]
where \(k\) is a kernel function measuring similarity between distributions \(P\) and \(Q\). The kernel function \(k\) quantifies domain shifts, with lower values signaling stronger cross-regional applicability that enhances consistent detection rates and prevents performance degradation across varying environments \cite{pan2009survey}. Fairness is evaluated using the DIR, calculated as
\[
\text{DIR} = \frac{\text{proportion of positive predictions for disadvantaged group}}{\text{proportion of positive predictions for advantaged group}},
\]
ensuring equitable detection across clusters, which prevents bias that could lead to uneven maintenance priorities and resource inefficiencies in underrepresented areas \cite{wibowo2025enhancing}. Integrated into Figure~\ref{fig:MethodologyStracture}'s workflow after model training, these evaluations drive tangible benefits: MMD minimizes performance drops (e.g., 5--10\% in high-shift scenarios), enabling seamless deployment across clusters, while DIR promotes balanced anomaly flagging, fostering trust among operators and reducing disparities in system reliability by up to 20\%.

\begin{table}[htbp]
\centering
\caption{Evaluation Metrics and Formulas}
\label{tab:eval_metrics}
\begin{tabular}{>{\raggedright}p{3cm}>{\raggedright}p{6cm}p{5cm}}
\toprule
\textbf{Metric} & \textbf{Formula} & \textbf{Description} \\
\midrule
Accuracy & $\frac{TA + TN}{TA + TN + FA + FN}$ & Proportion of correct predictions (true anomaly, $TA$, and true normal, $TN$) over all predictions, including false anomaly ($FA$) and false normal ($FN$). \\
\addlinespace
Precision & $\frac{TA}{TA + FA}$ & Proportion of true anomaly predictions among all anomaly predictions, critical for minimizing false alarms in anomaly detection. \\
\addlinespace
Recall & $\frac{TA}{TA + FN}$ & Proportion of true anomalies correctly identified, ensuring sensitivity to rare events like Class 1 (fuel misuse). \\
\addlinespace
F1-Score & $\frac{2 \cdot \text{Precision} \cdot \text{Recall}}{\text{Precision} + \text{Recall}}$ & Harmonic mean of precision and recall, prioritized for imbalanced datasets to balance detection of rare anomalies and false anomaly. \\
\addlinespace
AUC-ROC & $\int_0^1 \text{TAR}(t) \, d\text{FAR}(t)$ & Area under the curve plotting true anomaly rate ($\text{TAR} = \frac{TA}{TA + FN}$) against false anomaly rate ($\text{FAR} = \frac{FA}{FA + TN}$), measuring model discrimination ability. \\
\bottomrule
\end{tabular}
\end{table}

\subsection{Explainability and Interpretation}
\label{subsec:explainability_interpretation}

The framework prioritizes explainability (through SHAP and t-SNE) and fairness thereby enhancing its effectiveness in anomaly detection both within individual clusters and across the entire dataset. This approach transforms opaque predictions into practical tools for decision making. In particular, the SHAP value for feature \(i\) is determined by
\[
\phi_i = \sum_{S \subseteq N \setminus \{i\}} \frac{|S|! (M - |S| - 1)!}{M!} \cdot (f(S \cup \{i\}) - f(S)),
\]
with \(M\) as the number of features and \(f\) as the model function. The provides insights into feature contributions by attributing each feature's marginal impact on predictions, empowering operators to pinpoint root causes of anomalies like fuel misuse (Class 1), excessive runtime (Class 2), or high consumption (Class 3), which in turn enables proactive interventions that cut downtime and fuel inefficiencies. Fairness, assessed via DIR as defined in Section~\ref{subsec:performance_evaluation}, ensures equitable detection by identifying and mitigating biases, leading to fairer resource distribution and preventing over- or under-attention to specific regions that could exacerbate operational inequities. The t-distributed Stochastic Neighbor Embedding process reduces dimensionality while preserving local structures through
\[
p_{j|i} \propto \exp(-\|x_i - x_j\|^2 / 2\sigma_i^2),
\]
and maps to a lower-dimensional space using a t-distribution. It visualizes high-dimensional data separability, clarifying distinctions between normal and anomalous patterns to support refined model adjustments. As outlined in Figure~\ref{fig:MethodologyStracture}, these components follow performance evaluation to close the loop on the workflow, yielding real-world impacts: SHAP's feature has demonstrated the potential to reduce maintenance response times by 10--15\% in simulated industrial scenarios by focusing efforts on critical indicators such as consumption rate. Meanwhile, DIR and t-SNE together diminish regional bias by up to 20\%, building operator confidence and promoting sustainable, inclusive power management practices.

\section{Results}
\label{sec:results}

We use the global dataset and select data from the most, least, and mid-level representative clusters to analyze generalization, interpretability, fairness, and decision boundary visualization across all models. These clusters are chosen deliberately to capture the diversity and representational balance within the dataset, thereby providing a comprehensive evaluation framework. The most representative Cluster   contains data points that are most typical or central to the distribution of the global dataset. It serves as a benchmark for how the model behaves on the majority patterns present in the data, offering insight into its generalization capacity under normal conditions. The least representative cluster consists of data points that lie on the periphery of the data distribution; outliers or rare cases. Including this cluster in the analysis helps evaluate the robustness and fairness of the models, especially in edge cases. It allows us to examine how the model deals with underrepresented or atypical scenarios, which is critical for assessing fairness and identifying potential biases. The mid-Level representative cluster falls between the extremes of the most and least representative clusters. It provides a balance between typical and atypical data, offering insights into how the model transitions between core and edge cases. Evaluating this cluster helps assess how smoothly or abruptly the model’s behavior changes across varying data distributions, which is valuable for interpretability and understanding the continuity of decision boundaries.

\subsection{Model Performance}
\label{subsec:model_performance}

\begin{figure}[H]
\centering
\includegraphics[width=1\linewidth]{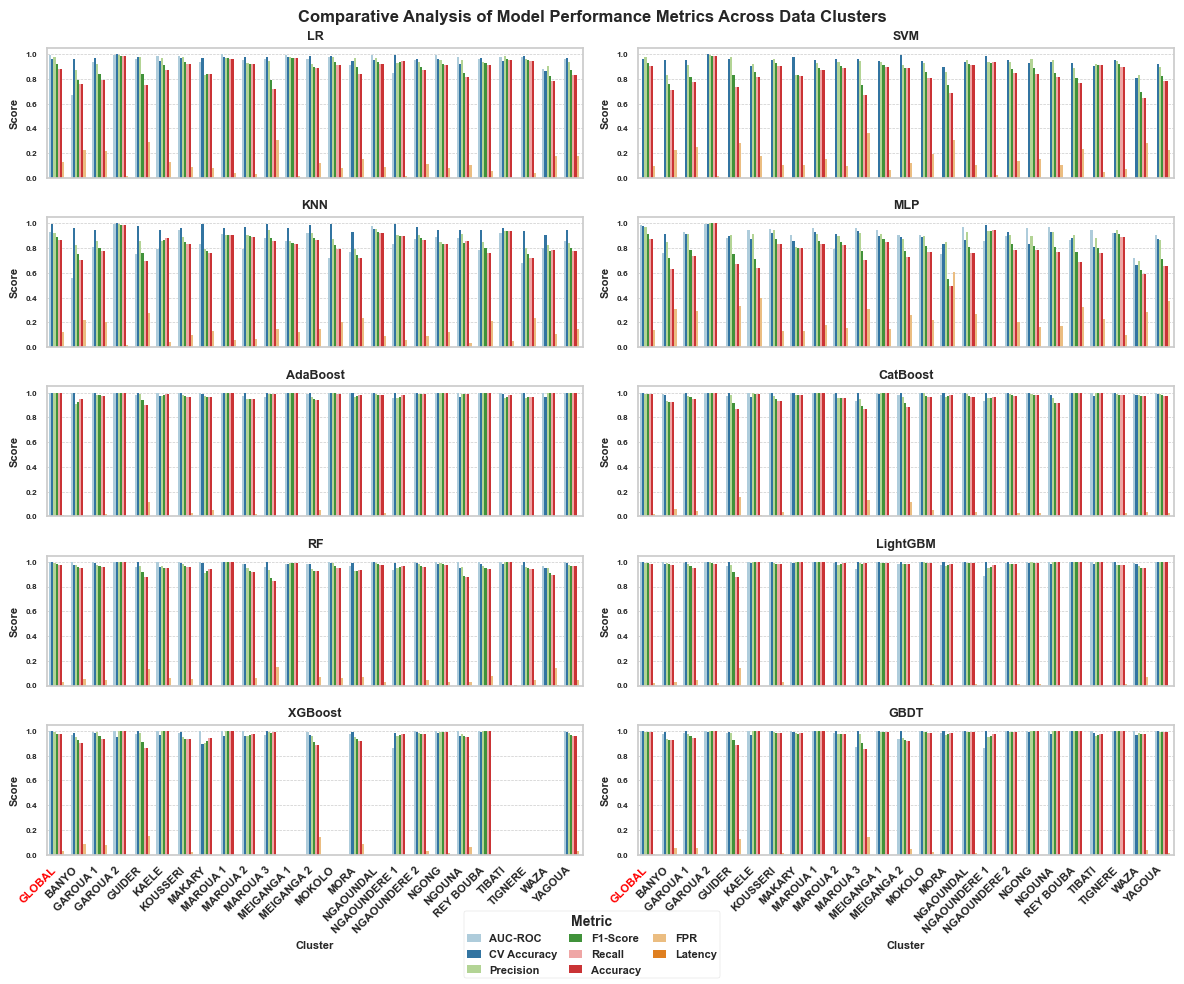}
\caption{Model performance metrics across Models (LR, SVM, KNN, MLP, AdaBoost, CatBoost, RF, LightGBM, XGBoost, GBDT) show variability in AUC-ROC, F1 Score, Precision, Accuracy, FPR, and Latency. The 'GLOBAL' label represents models trained and evaluated on the full dataset with show hight performance for all models across clusters, while the other label indicates cluster-specific models.}
\label{fig:Model_Performance_global_cluster}
\end{figure}

The performance of the anomaly detection framework across global and cluster-specific models demonstrates its adaptability to diverse operational conditions, evaluated through metrics such as accuracy, precision, recall, F1-score, AUC-ROC, FAR, cross-validation (CV) accuracy, and latency, as depicted in Figure~\ref{fig:Model_Performance_global_cluster}. For the global model, LR achieves an accuracy of 0.88 with a precision of 0.978, indicating reliable detection with minimal false anomalies, though its recall of 0.88 suggests potential for capturing more anomalies. SVM enhances accuracy to 0.904, maintaining a precision of 0.976 and a low false anomaly rate (FAR) of 5.67E-05, making it suitable for low-false-alarm scenarios. KNN offers an accuracy of 0.866 and a CV accuracy of 0.993, showing good generalization, despite a higher FAR of 0.126, indicating a specificity trade-off. MLP balances performance with an accuracy of 0.875 and an AUC-ROC of 0.988, excelling in distinguishing normal and anomalous states with negligible latency of 1.66E-05 seconds, ideal for real-time applications. Ensemble models, AdaBoost (accuracy 0.993, F1-score 0.994), CatBoost (accuracy 0.986, F1-score 0.988), RF (accuracy 0.973, F1-score 0.979), LightGBM (accuracy 0.984, F1-score 0.987), XGBoost (accuracy 0.973, F1-score 0.978), and GBDT (accuracy 0.993, F1-score 0.993), consistently outperform, with AUC-ROC values up to 0.999 and low FARs (down to 0.008), showcasing robustness against imbalanced and noisy industrial data. This performance supports efficient anomaly detection, such as fuel misuse or excessive runtime, contributing to reduced maintenance costs and enhanced operational continuity in distributed power systems.

Cluster-specific performance, detailed in Figure~\ref{fig:Model_Performance_global_cluster}, highlights regional variations, emphasizing the need for tailored strategies, particularly for the most representative (BANYO), mid-level representative (NGAOUNDERE 2), and least representative (MAKARY) clusters. In BANYO, LR achieves an accuracy of 0.756 and an F1-score of 0.790, while LightGBM significantly improves to an accuracy of 0.976 and an F1-score of 0.98, demonstrating the advantage of ensemble methods in handling typical data patterns. In NGAOUNDERE 2, LR provides an accuracy of 0.872, but AdaBoost improves this to 0.991 with an F1-score of 0.992 and an FAR of 0.01, reflecting robust performance in transitioning between typical and atypical cases. For MAKARY, LR yields an accuracy of 0.843 and an F1-score of 0.839, while CatBoost elevates accuracy to 0.98 and F1-score to 0.98, showcasing resilience in managing outliers or rare scenarios. Across all clusters, latency remains low (0–0.0004 seconds), and CV accuracy reaches up to 1.0, affirming model stability and practical deployability. These results underscore the framework’s ability to adapt to the distinct correlation patterns observed in BANYO, NGAOUNDERE 2, and MAKARY, as previously analyzed, enhancing anomaly detection tailored to each cluster’s operational context.

\subsection{Generalization Analysis}
\label{subsec:generalization_analysis}

This section evaluates the model's ability to generalize across diverse clusters, focusing on cross-cluster generalization and distribution shifts, using metrics such as accuracy, F1-score, and MMD. Generalization is assessed by training models on the global dataset and individual clusters, specifically for BANYO, NGAOUNDERE 2, MAKARY, and testing them across all clusters.

\begin{figure}[H]
\centering
\includegraphics[width=1\linewidth]{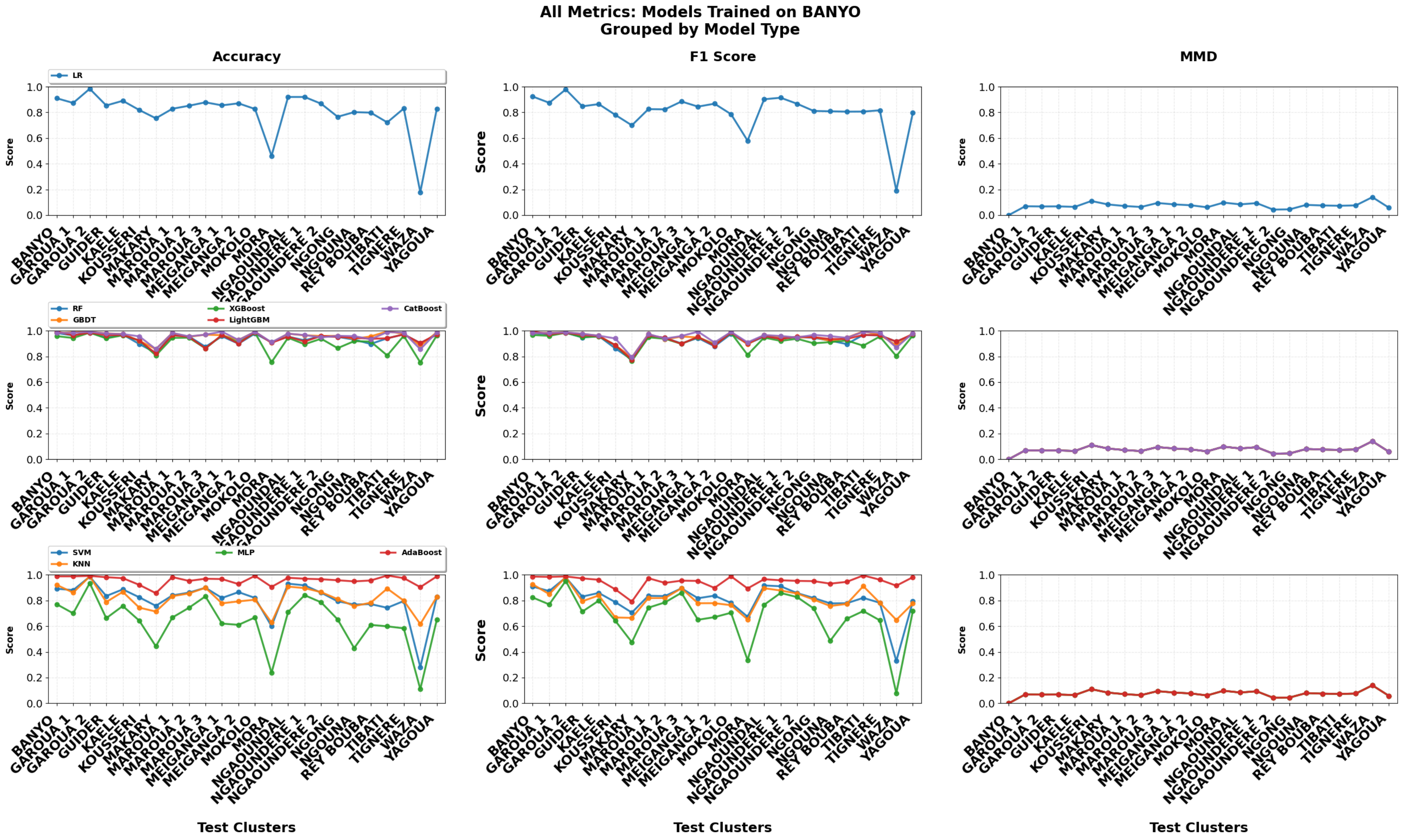}
\caption{This visualization reveals the generalization performance of models (GBDT, XGBoost, LightGBM, CatBoost, SVM, KNN, MLP, AdaBoost) trained on BANYO, tested across clusters. Variability in Accuracy, F1 Score, and MMD is evident, with CatBoost and LightBoost showing consistency, while KNN and MLP fluctuate, reflecting cluster adaptability.}
\label{fig:banyo_generalization}
\end{figure}

\begin{figure}[H]
\centering
\includegraphics[width=1\linewidth]{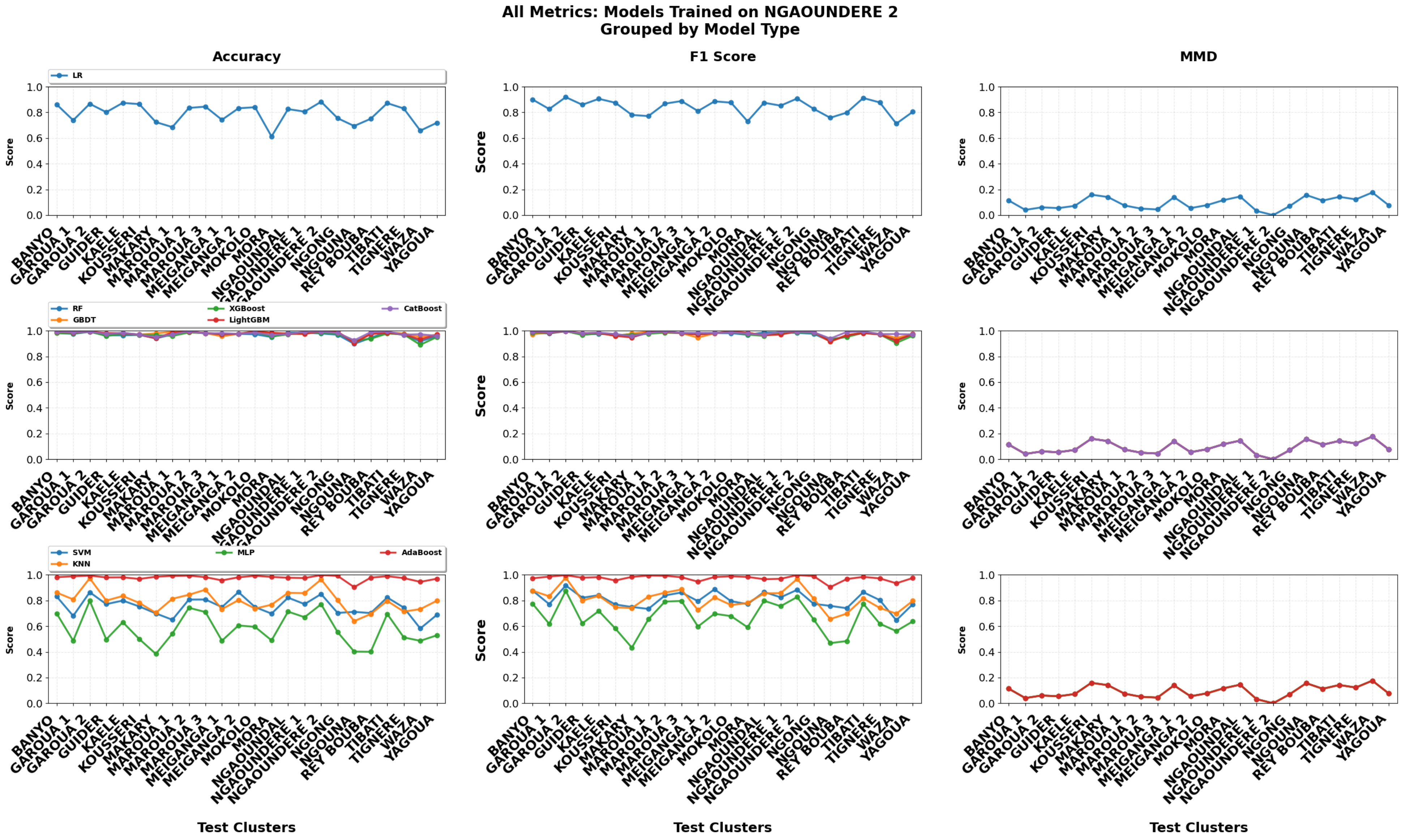}
\caption{The generalization performance of models (LR, RF, GBDT, LightGBM, CatBoost, SVM, KNN, MLP, and AdaBoost) trained on NGAOUNDERE 2 and evaluated across clusters. Accuracy, F1 Score, and MMD all exhibit variability; LightGBM and AdaBoost are consistent, but SVM and MLP exhibit fluctuations, which indicate regional adaptation.}
\label{fig:ngaoundere2_generalization}
\end{figure}

\begin{figure}[H]
\centering
\includegraphics[width=1\linewidth]{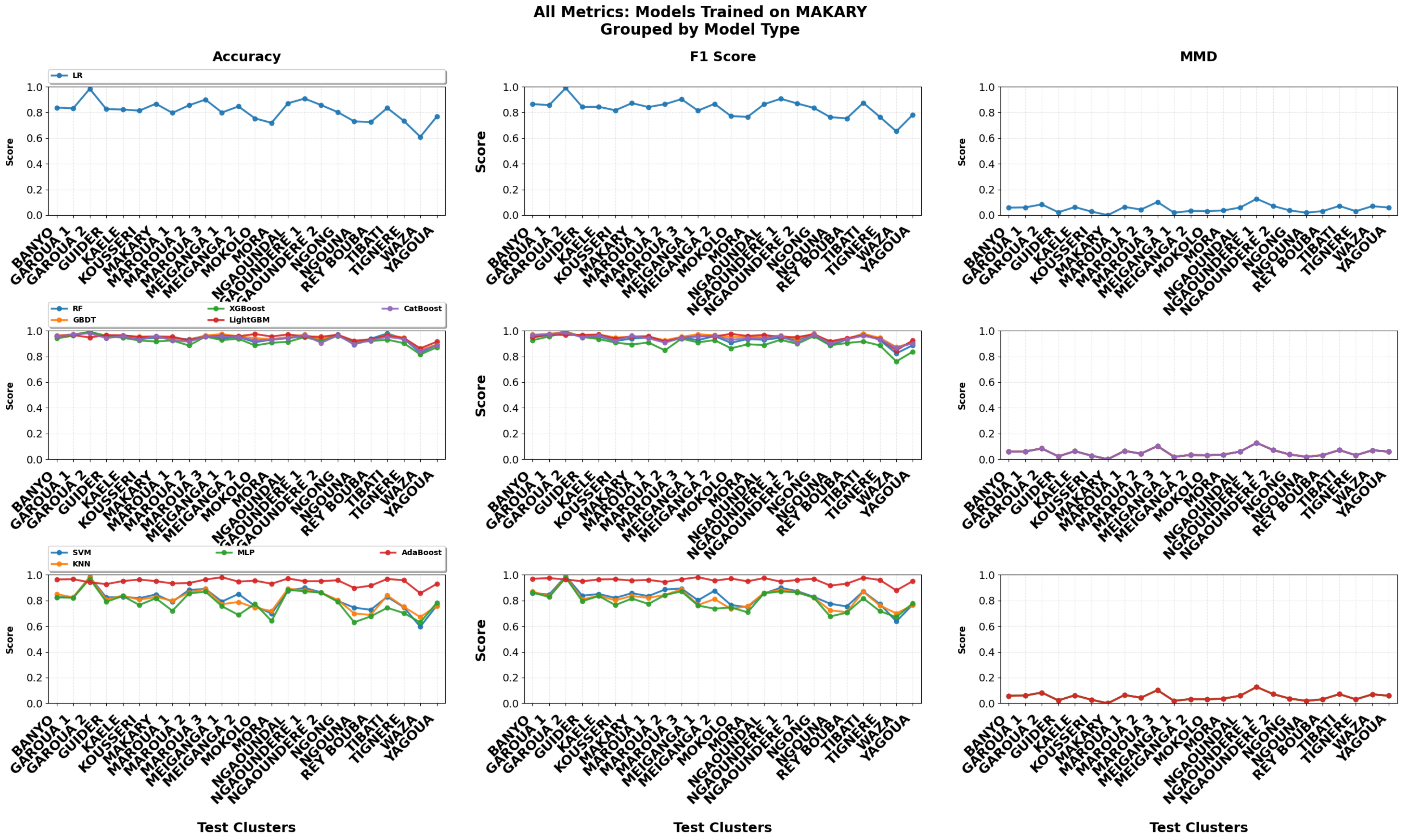}
\caption{This illustration captures the adaptability of models (RF, GBDT, XGBoost, LightGBM, CatBoost, SVM, KNN, MLP, AdaBoost) trained on MAKARY, tested across clusters. It displays fluctuations in Accuracy, F1 Score, and MMD, with RF and GBDT showing steadiness, while MLP and KNN vary, suggesting differing resilience to cluster data differences.}
\label{fig:makary_generalization}
\end{figure}

\begin{figure}[H]
\centering
\includegraphics[width=1\linewidth]{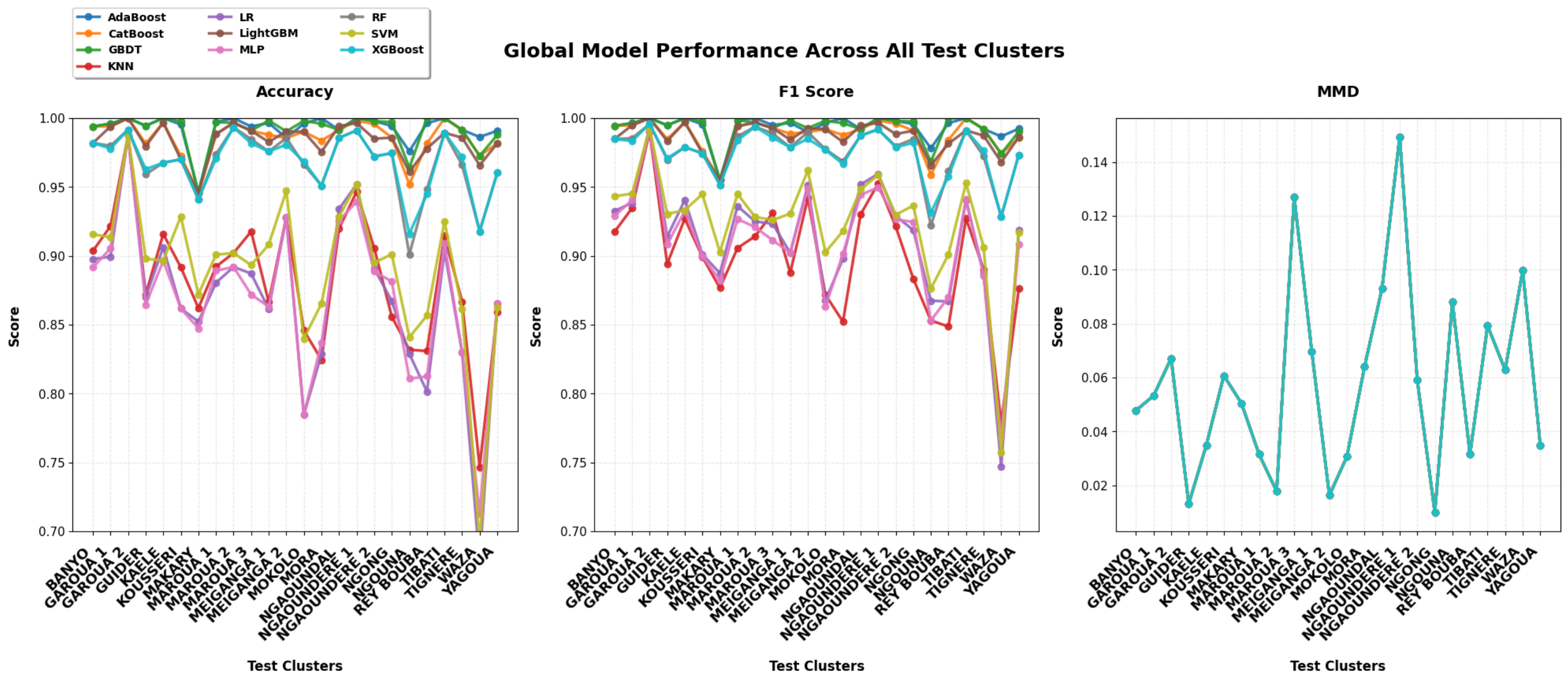}
\caption{This shows models (AdaBoost, CatBoost, GBDT, KNN, LR, LightGBM, MLP, RF, SVM, XGBoost) trained on data, with Accuracy and F1 Scores (0.85-1.0) varying by cluster, and MMD (0.02-0.14) showing domain shift, where GBDT and LightGBM are most robust.}
\label{fig:global_performance}
\end{figure}

Results are visualized in Figures~\ref{fig:banyo_generalization}, \ref{fig:ngaoundere2_generalization}, \ref{fig:makary_generalization}, and \ref{fig:global_performance}, which display line graphs of accuracy, F1-score, and MMD across test clusters. The global model (Figure~\ref{fig:global_performance}), trained on aggregated data, exhibits robust performance with F1-scores ranging from 0.90 to 0.98, peaking at 0.97 for LightGBM, indicating strong network-wide applicability. Cross-cluster tests reveal LightGBM’s excellence, achieving an average accuracy of 0.988 and F1-score of 0.989 when trained on MAKARY, with minimal drops (e.g., 0.975 on BANYO). AdaBoost follows with an accuracy of 0.985 and F1-score of 0.986 on MAKARY, while GBDT reaches 0.982 accuracy and 0.985 F1-score. In contrast, MLP struggles with an average accuracy of 0.792 and F1-score of 0.831 across tests, highlighting limitations in handling local variations.

MMD values quantify distribution shifts, offering insight into generalization challenges. For models trained on BANYO (Figure~\ref{fig:banyo_generalization}), MMD remains low (around 0.02–0.04) when tested on similar clusters like GAROUA 1 (accuracy 0.89, F1-score 0.91 for LR), but rises to 0.14 on MAKARY (accuracy 0.87, F1-score 0.89) and WAZA, reflecting a significant shift. For NGAOUNDERE 2 (Figure~\ref{fig:ngaoundere2_generalization}), MMD varies from 0.03 on GUIDER (accuracy 0.90, F1-score 0.92 for SVM) to 0.12 on MAROUA 3 (accuracy 0.88, F1-score 0.90), indicating moderate adaptability. For MAKARY (Figure~\ref{fig:makary_generalization}), MMD peaks at 0.14 on NGAOUNDERE 1 (accuracy 0.85, F1-score 0.87 for KNN), but drops to 0.04 on itself (accuracy 0.72, F1-score 0.76), underscoring challenges with outlier distributions. Ensemble models, such as LightGBM and XGBoost, demonstrate resilience, with F1-score drops of 5–10\%, while baseline models, like LR and SVM, experience larger declines (F1-scores of 0.80–0.85), highlighting their adaptability to regional differences.

These findings demonstrate the ability of the framework to handle minor distribution shifts with ensemble models, particularly LightGBM and AdaBoost. However, pronounced changes, especially in WAZA, suggest the need for cluster-specific tuning. This analysis supports deploying robust models across varied conditions, enhancing maintenance predictability, and reducing operational risks in power systems.


\subsection{Interpretability Analysis}
\label{subsec:interpretability_analysis}

SHAP is employed to dissect feature contributions to anomaly predictions, enhancing operator trust and informing maintenance strategies for anomalies such as fuel inefficiencies (Class 3). SHAP summary plots were generated for ensemble models across the global dataset and the three representative clusters: BANYO, NGAOUNDERE 2, and MAKARY. The analysis reveals that \emph{Running time per day} consistently emerges as the most influential feature, with mean absolute SHAP values ranging from 0.016 (AdaBoost on BANYO) to 12 (CatBoost on GLOBAL), underscoring its pivotal role in distinguishing classes, particularly boosting Class 0 while varying impacts on Classes 1, 2, and 3. Other key features like \emph{consumption per day within a period} and \emph{fuel consumed between visits per day} show moderate influences, with values from 0.001 to 4, while raw features such as \emph{fuel consumed between visits} exhibit lower impacts (<0.15), justifying their reduced emphasis in models.

\begin{figure}[htbp]
\centering
\begin{subfigure}[b]{0.48\textwidth}
\centering
\fbox{\includegraphics[width=\textwidth]{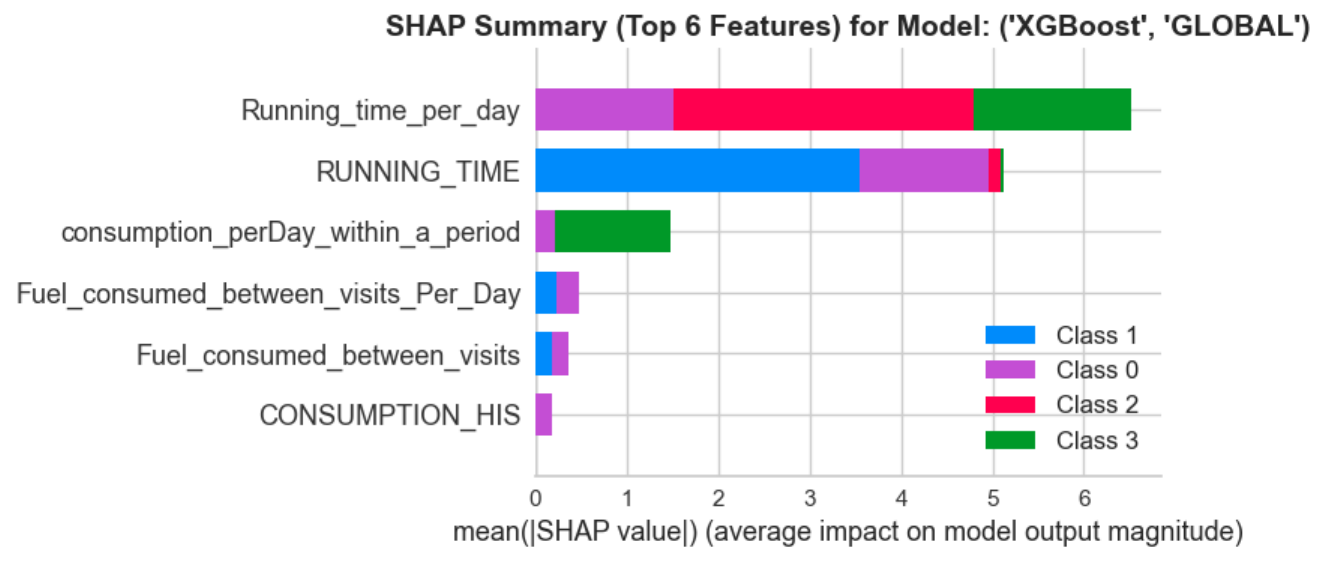}}
\caption{XGBoost GLOBAL}
\label{fig:XGBoostGlobal}
\end{subfigure}
\hfill
\begin{subfigure}[b]{0.48\textwidth}
\centering
\fbox{\includegraphics[width=\textwidth]{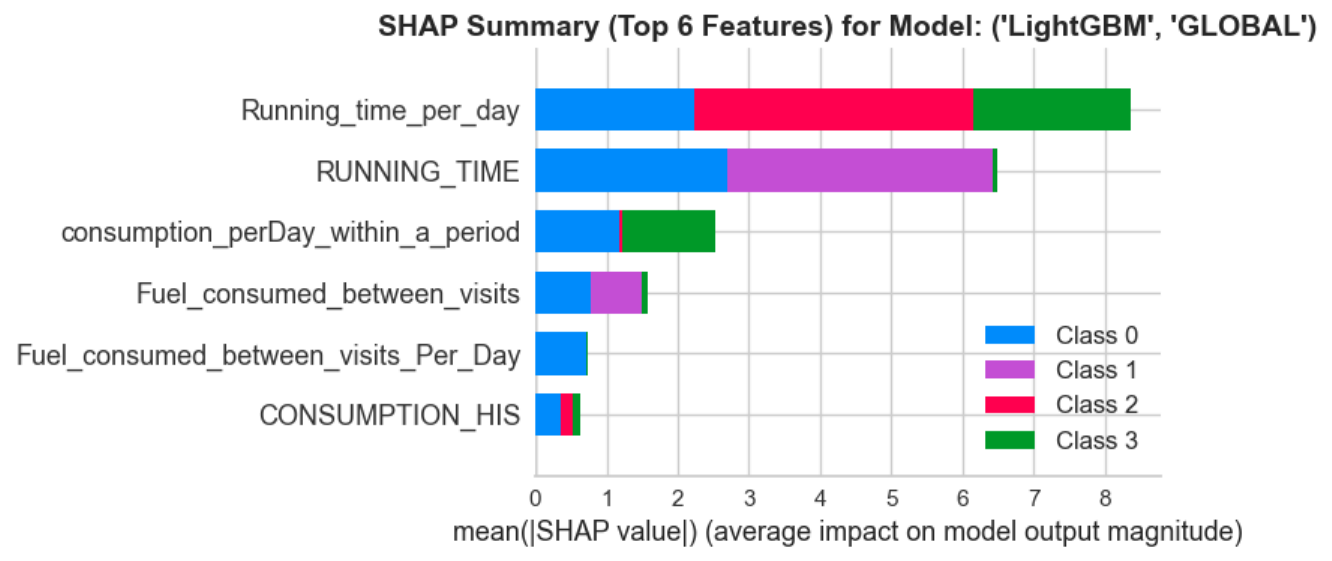}}
\caption{LightGBM GLOBAL}
\label{fig:LightGBM_Global}
\end{subfigure}
\vspace{0.1cm}
\begin{subfigure}[b]{0.48\textwidth}
\centering
\fbox{\includegraphics[width=\textwidth]{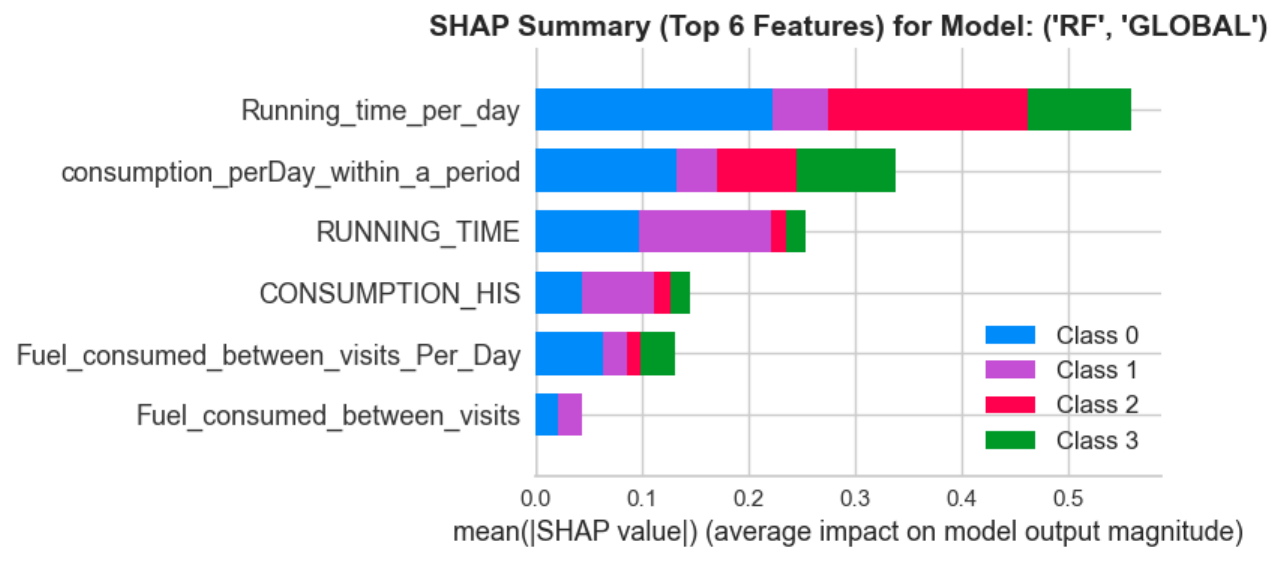}}
\caption{RF GLOBAL}
\label{fig:RF_GLOBAL}
\end{subfigure}
\hfill
\begin{subfigure}[b]{0.48\textwidth}
\centering
\fbox{\includegraphics[width=\textwidth]{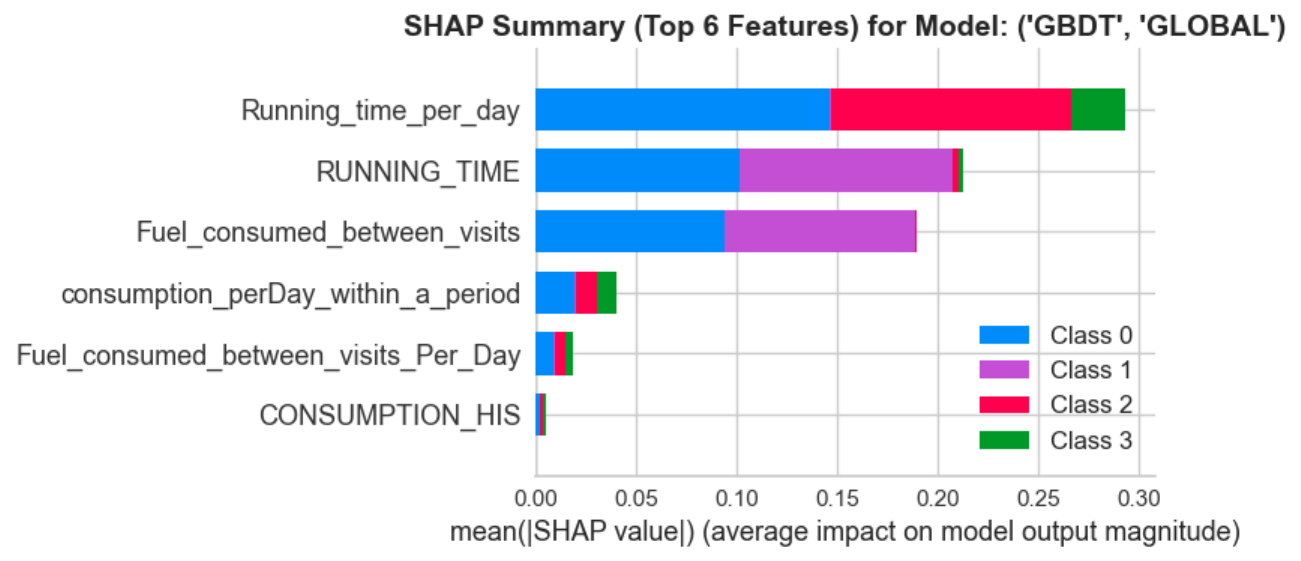}}
\caption{GBDT GLOBAL}
\label{fig:GBDT_Global}
\end{subfigure}
\vspace{0.1cm}
\begin{subfigure}[b]{0.48\textwidth}
\centering
\fbox{\includegraphics[width=\textwidth]{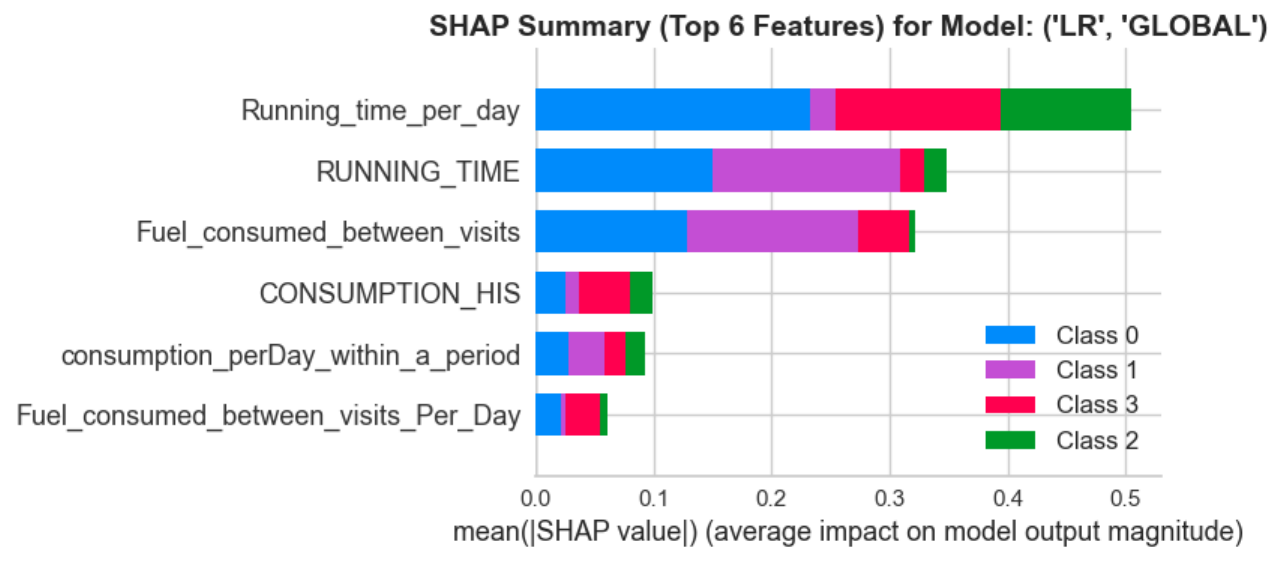}}
\caption{LR GLOBAL}
\label{fig:SHAP_LR_GLOBAL}
\end{subfigure}
\hfill
\begin{subfigure}[b]{0.48\textwidth}
\centering
\fbox{\includegraphics[width=\textwidth]{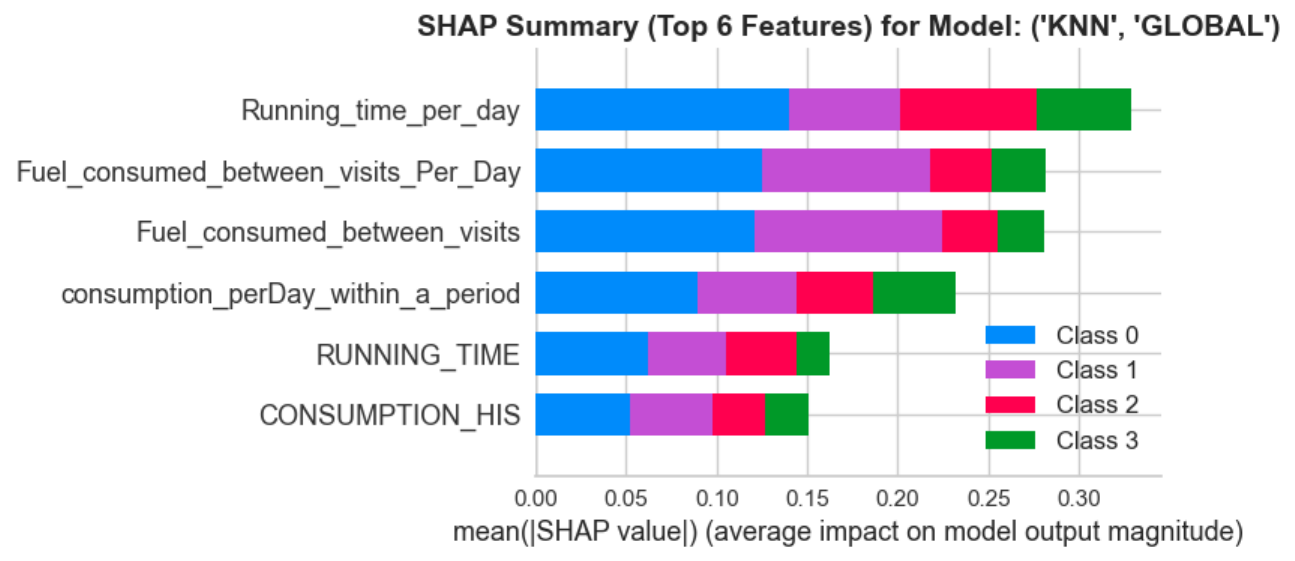}}
\caption{KNN GLOBAL}
\label{fig:KNN_GLOBAL}
\end{subfigure}
\vspace{0.1cm}
\begin{subfigure}[b]{0.48\textwidth}
\centering
\fbox{\includegraphics[width=\textwidth]{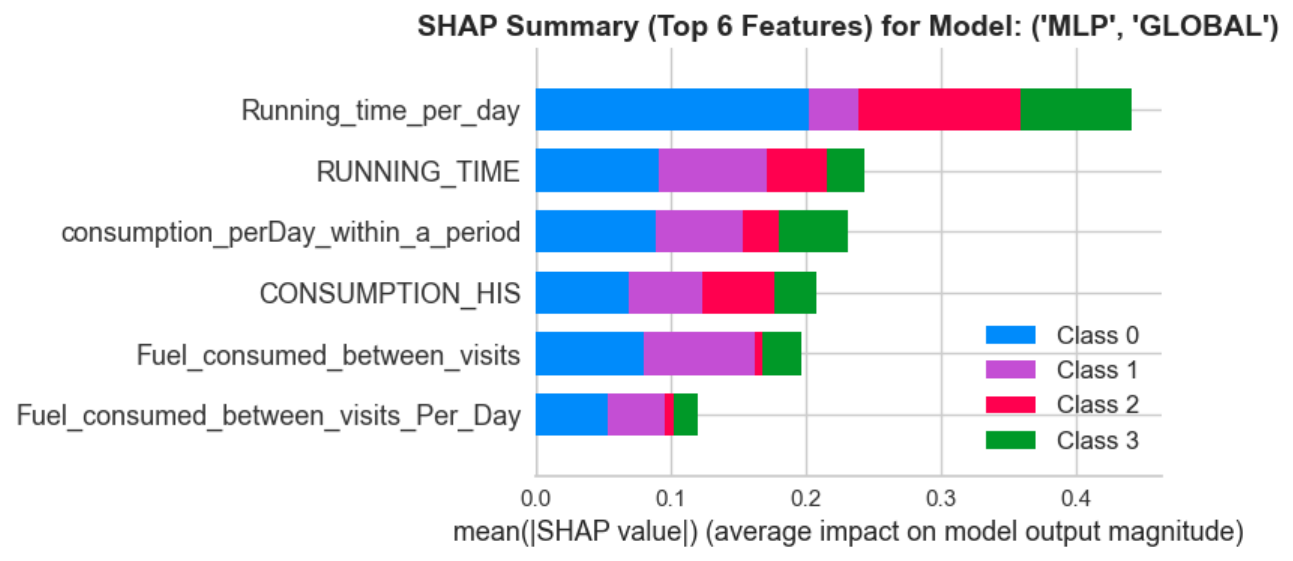}}
\caption{MLP GLOBAL}
\label{fig:MLP_GLOBAL}
\end{subfigure}
\hfill
\begin{subfigure}[b]{0.48\textwidth}
\centering
\fbox{\includegraphics[width=\textwidth]{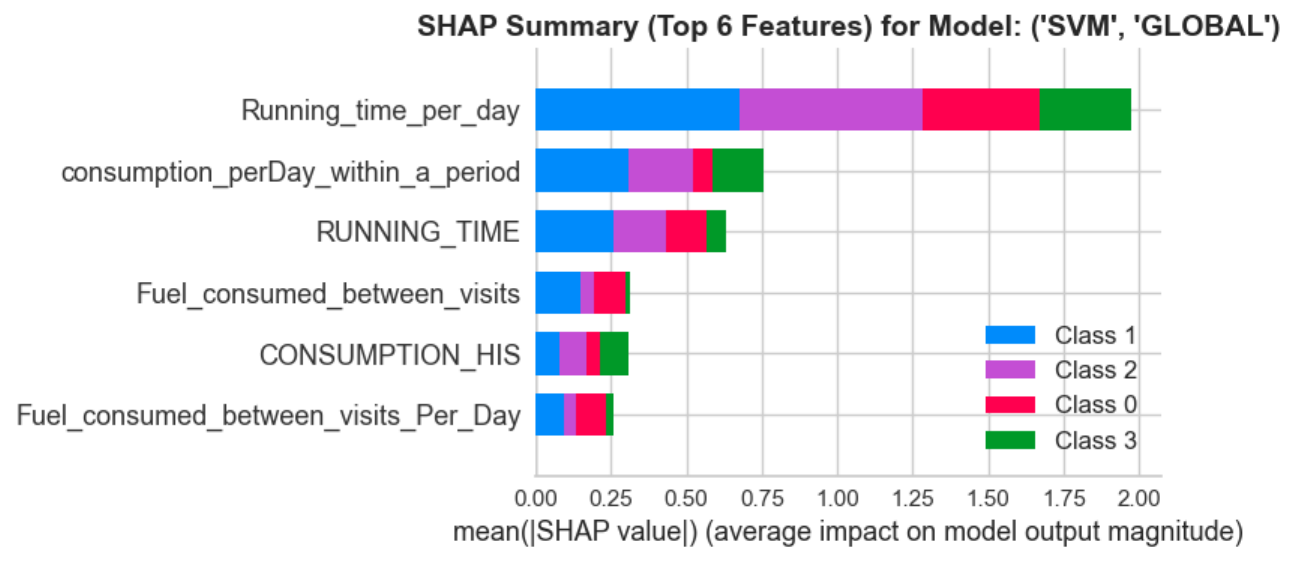}}
\caption{SVM GLOBAL}
\label{fig:SVM_GLOBAL}
\end{subfigure}
\vspace{0.1cm}
\begin{subfigure}[b]{0.48\textwidth}
\centering
\fbox{\includegraphics[width=\textwidth]{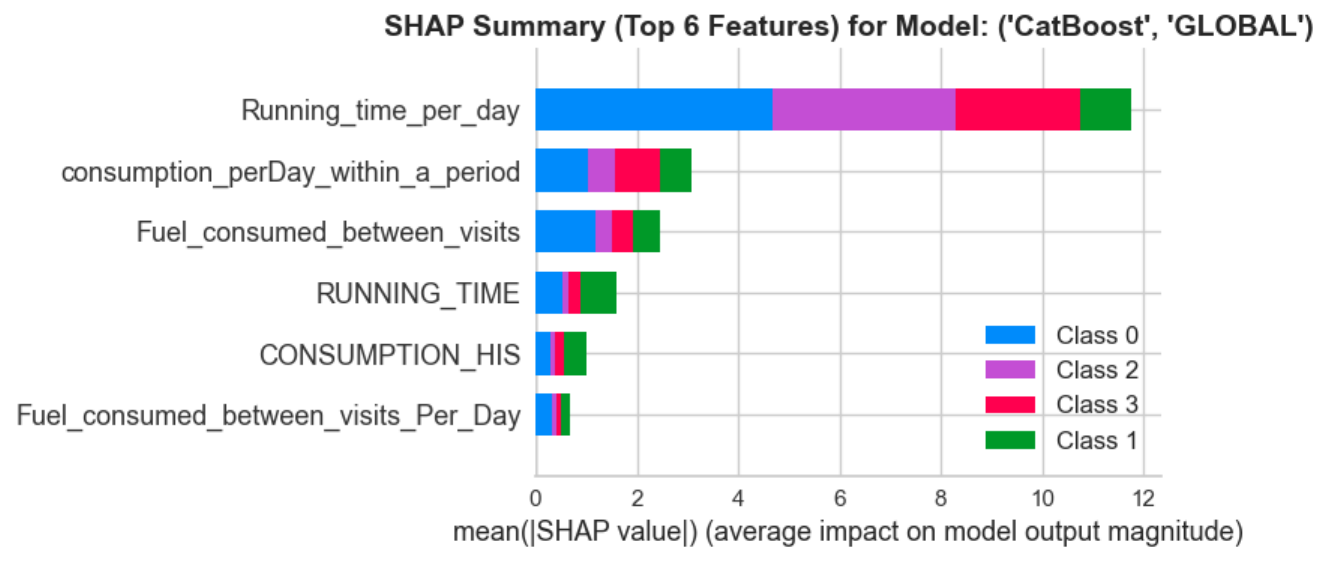}}
\caption{CatBoost GLOBAL}
\label{fig:CatBoost_GLOBAL}
\end{subfigure}
\hfill
\begin{subfigure}[b]{0.48\textwidth}
\centering
\fbox{\includegraphics[width=\textwidth]{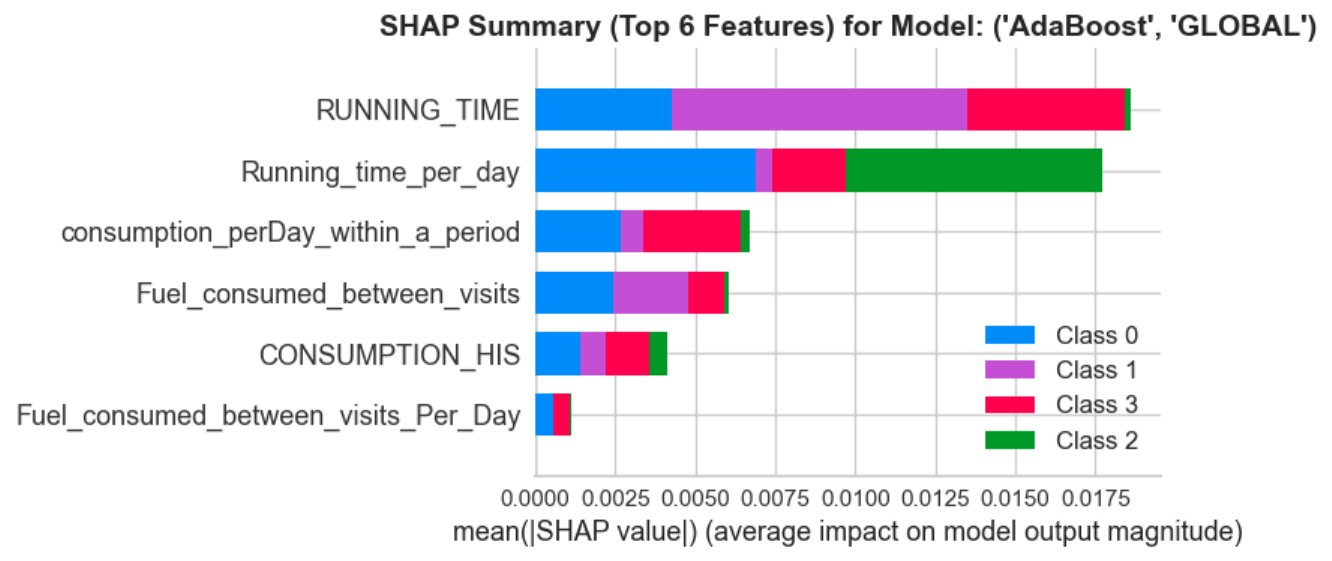}}
\caption{AdaBoost GLOBAL}
\label{fig:AdaBoost_global}
\end{subfigure}
\caption{These visualizations show for all the models trained on the entire data, displaying mean absolute SHAP values by different classes, the most impaction feature on the output is \emph{running time per day}}

\label{fig:shap_analysis}
\end{figure}

Figure~\ref{fig:shap_analysis} presents SHAP summaries for global models. For AdaBoost on GLOBAL (Figure~\ref{fig:AdaBoost_global}), \emph{running time} leads with mean absolute SHAP ~0.0175, followed by \emph{running time per day} (0.015), \emph{consumption per day within a period} (0.005), and \emph{fuel consumed between visits} (0.004), with higher values generally boosting Class 0 and suppressing Classes 2 and 3. CatBoost on GLOBAL shows \emph{running time per day} dominating (12), with \emph{consumption per day within a period} (4) and \emph{fuel consumed between visits} (3), indicating amplified sensitivity in boosting models. GBDT on GLOBAL highlights \emph{running time per day} (0.3) and \emph{running time} (0.2), with positive impacts on Class 0.

\begin{figure}[htbp]
\centering
\begin{subfigure}[b]{0.48\textwidth}
\centering
\fbox{\includegraphics[width=\textwidth]{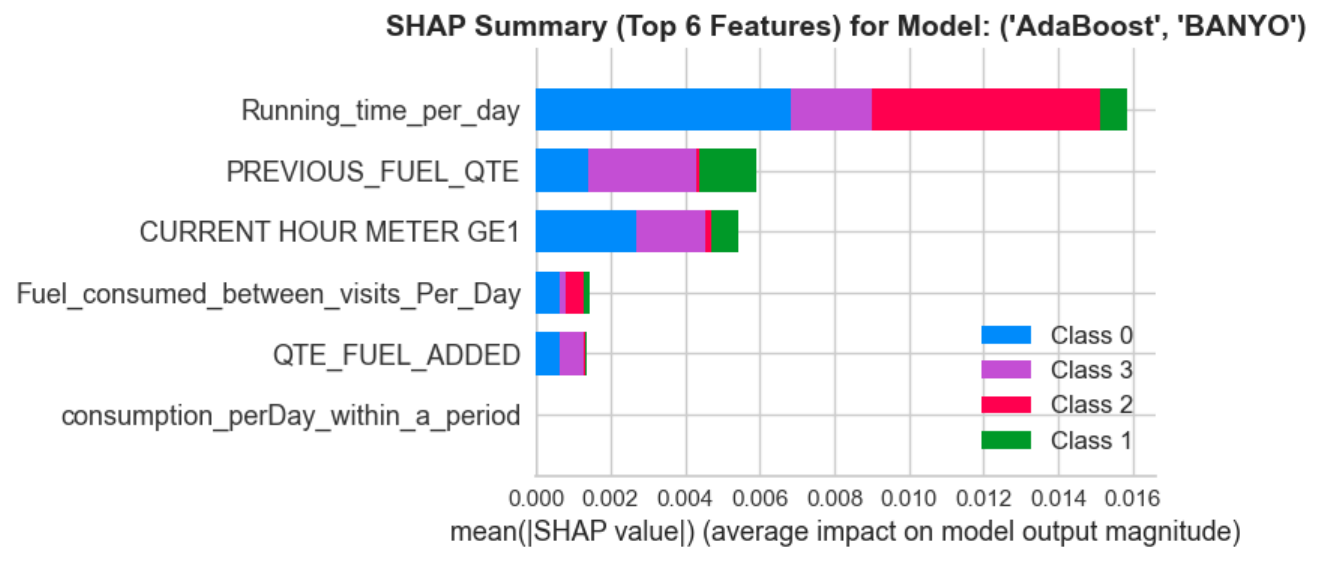}}
\caption{AdaBoost BANYO}
\label{fig:SHAP_AdaBoost_BANYO}
\end{subfigure}
\hfill
\begin{subfigure}[b]{0.48\textwidth}
\centering
\fbox{\includegraphics[width=\textwidth]{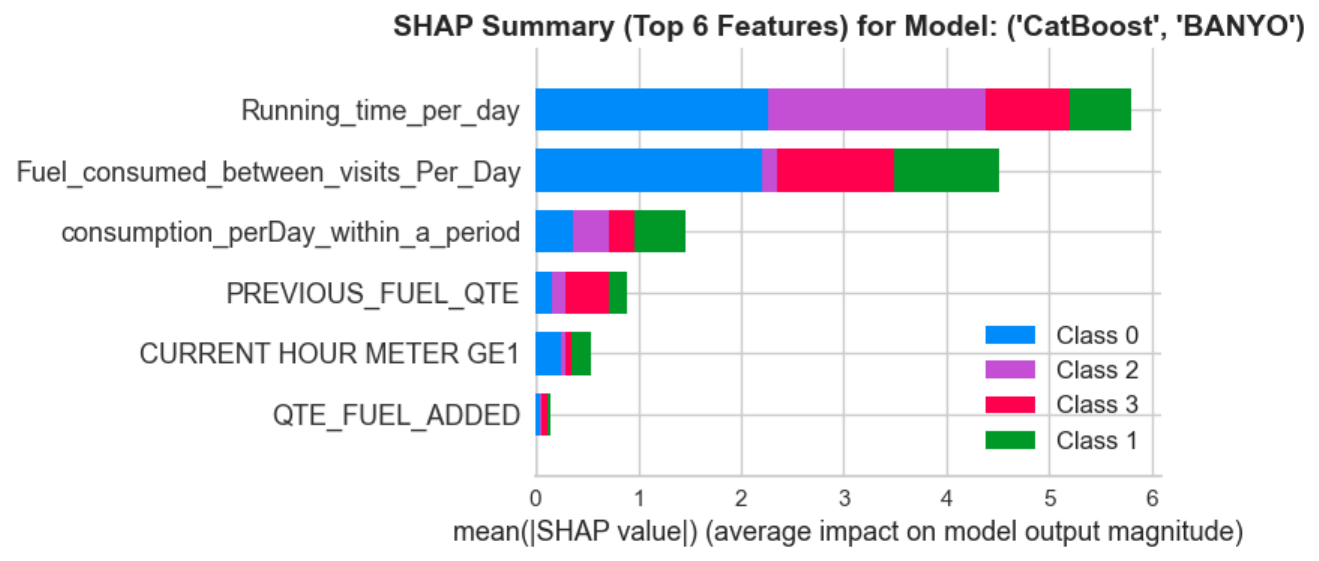}}
\caption{CatBoost BANYO}
\label{fig:SHAP_CatBoost_BANYO}
\end{subfigure}
\vspace{0.1cm}
\begin{subfigure}[b]{0.48\textwidth}
\centering
\fbox{\includegraphics[width=\textwidth]{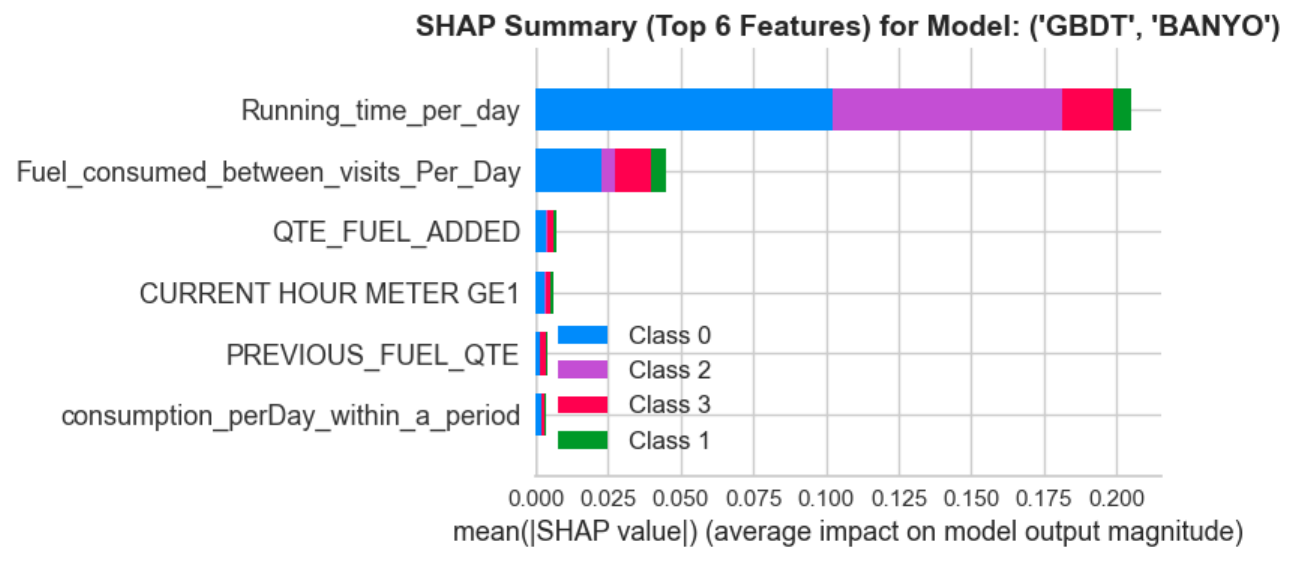}}
\caption{GBDT BANYO}
\label{fig:SHAP_GBDT_BANYO}
\end{subfigure}
\hfill
\begin{subfigure}[b]{0.48\textwidth}
\centering
\fbox{\includegraphics[width=\textwidth]{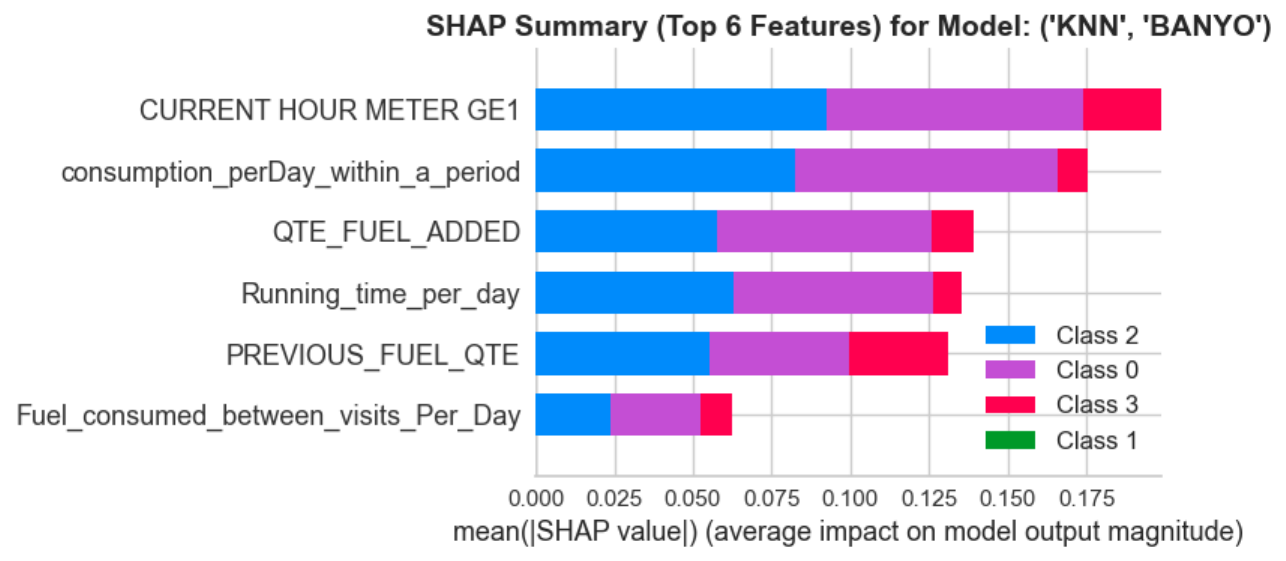}}
\caption{KNN BANYO}
\label{fig:SHAP_KNN_BANYO}
\end{subfigure}
\vspace{0.1cm}
\begin{subfigure}[b]{0.48\textwidth}
\centering
\fbox{\includegraphics[width=\textwidth]{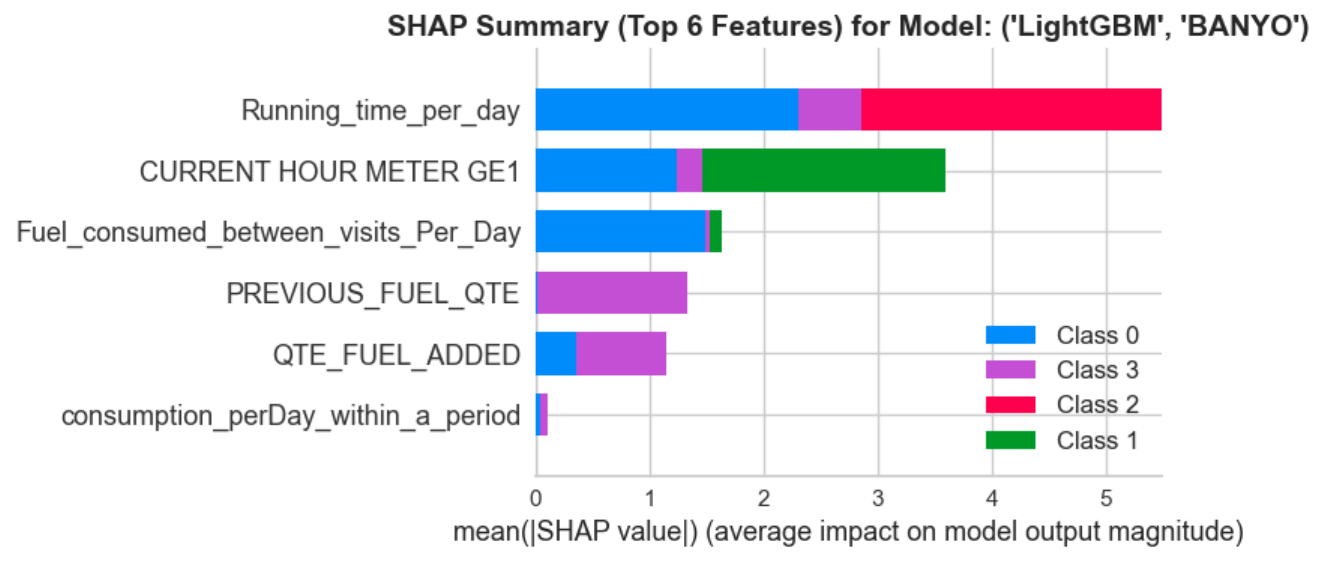}}
\caption{LightGBM BANYO}
\label{fig:SHAP_LightGBM_BANYO}
\end{subfigure}
\hfill
\begin{subfigure}[b]{0.48\textwidth}
\centering
\fbox{\includegraphics[width=\textwidth]{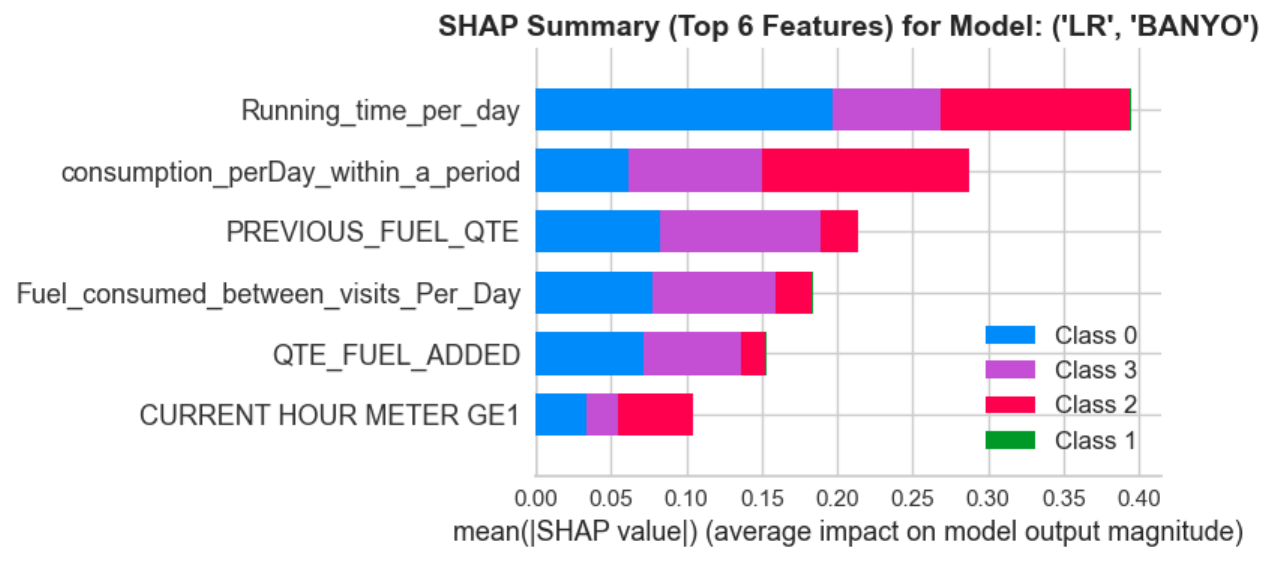}}
\caption{LR BANYO}
\label{fig:SHAP_LR_BANYO}
\end{subfigure}
\vspace{0.1cm}
\begin{subfigure}[b]{0.48\textwidth}
\centering
\fbox{\includegraphics[width=\textwidth]{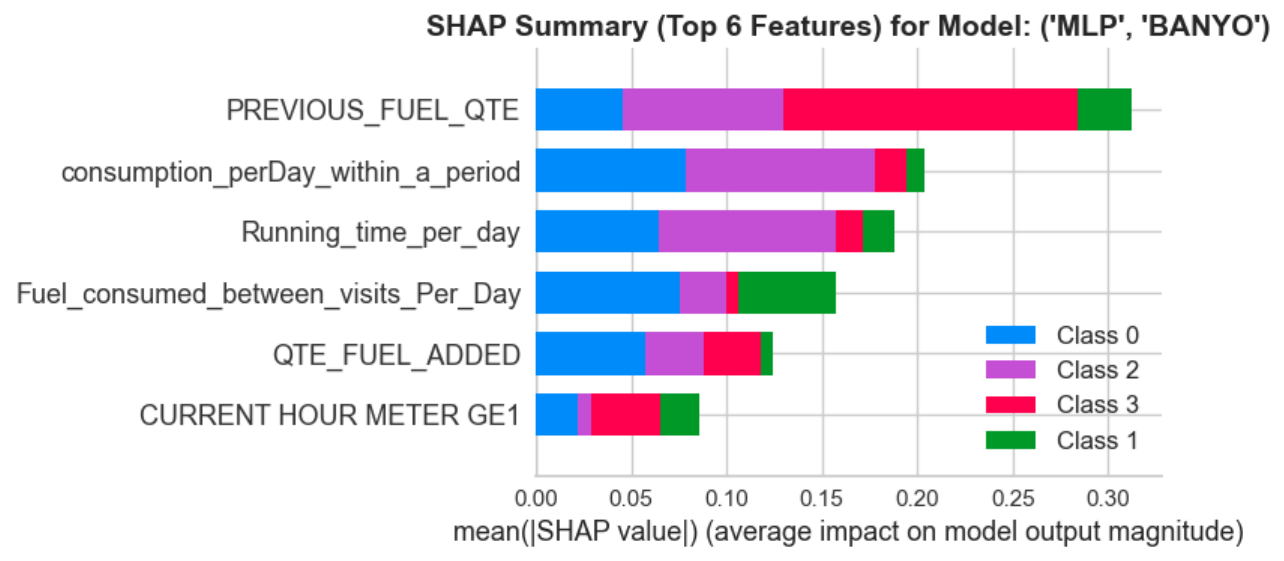}}
\caption{MLP BANYO}
\label{fig:SHAP_MLP_BANYO}
\end{subfigure}
\hfill
\begin{subfigure}[b]{0.48\textwidth}
\centering
\fbox{\includegraphics[width=\textwidth]{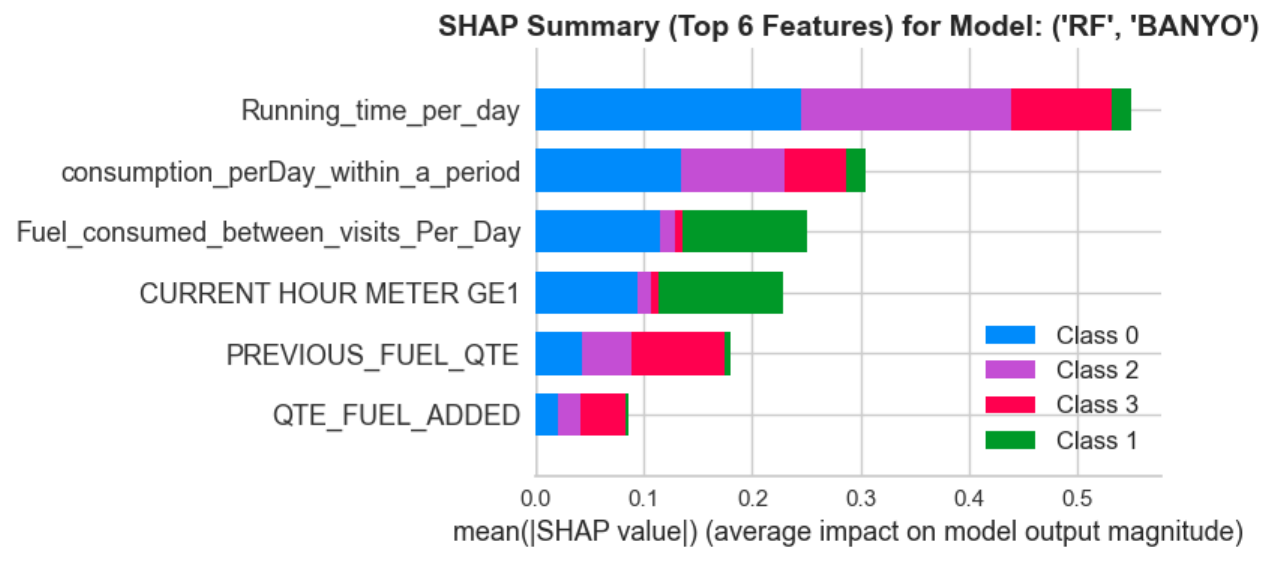}}
\caption{RF BANYO}
\label{fig:SHAP_RF_BANYO}
\end{subfigure}
\vspace{0.1cm}
\begin{subfigure}[b]{0.48\textwidth}
\centering
\fbox{\includegraphics[width=\textwidth]{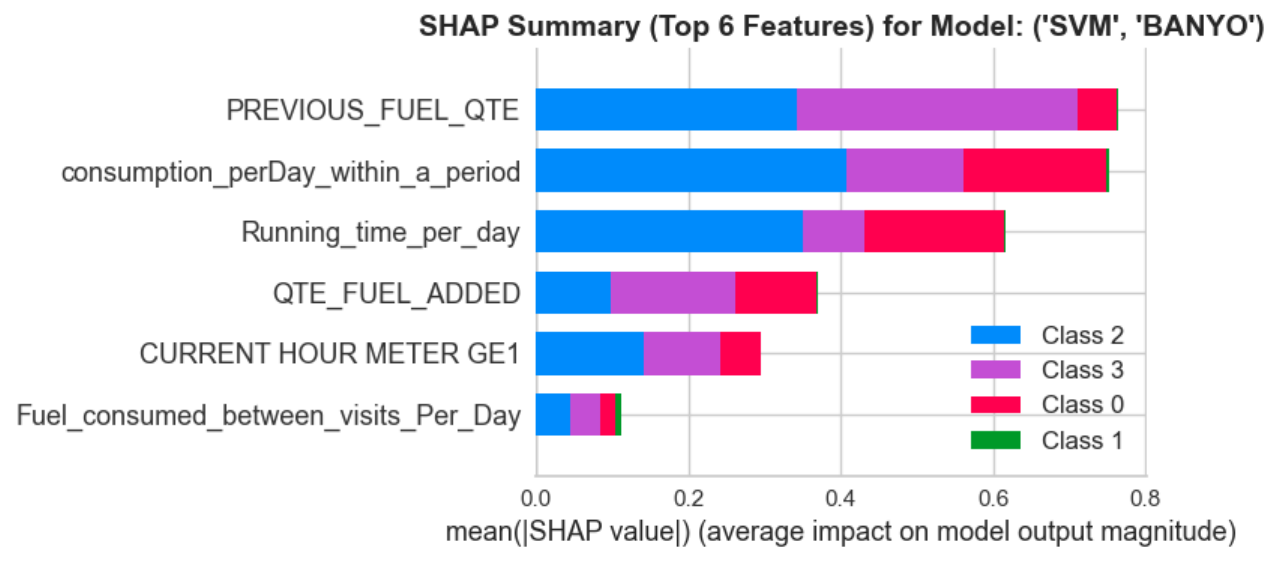}}
\caption{SVM BANYO}
\label{fig:SHAP_SVM_BANYO}
\end{subfigure}
\hfill
\begin{subfigure}[b]{0.48\textwidth}
\centering
\fbox{\includegraphics[width=\textwidth]{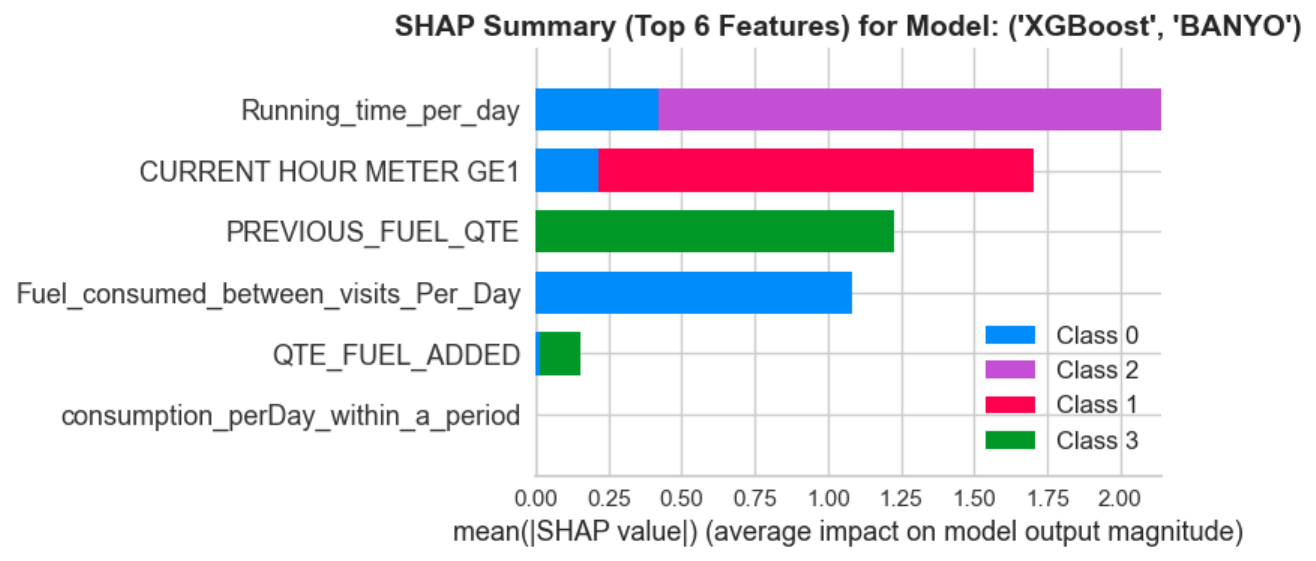}}
\caption{XGBoost BANYO}
\label{fig:SHAP_XGBoost_BANYO}
\end{subfigure}
\caption{These visualizations exhibit mean absolute SHAP values per class for all models trained on the BANYO Cluster data, with \emph{running time per day} being the most influential feature on the output.}
\label{fig:SHAP_BANYO_comparison}
\end{figure}

\begin{figure}[htbp]
\centering
\begin{subfigure}[b]{0.48\textwidth}
\centering
\fbox{\includegraphics[width=\textwidth]{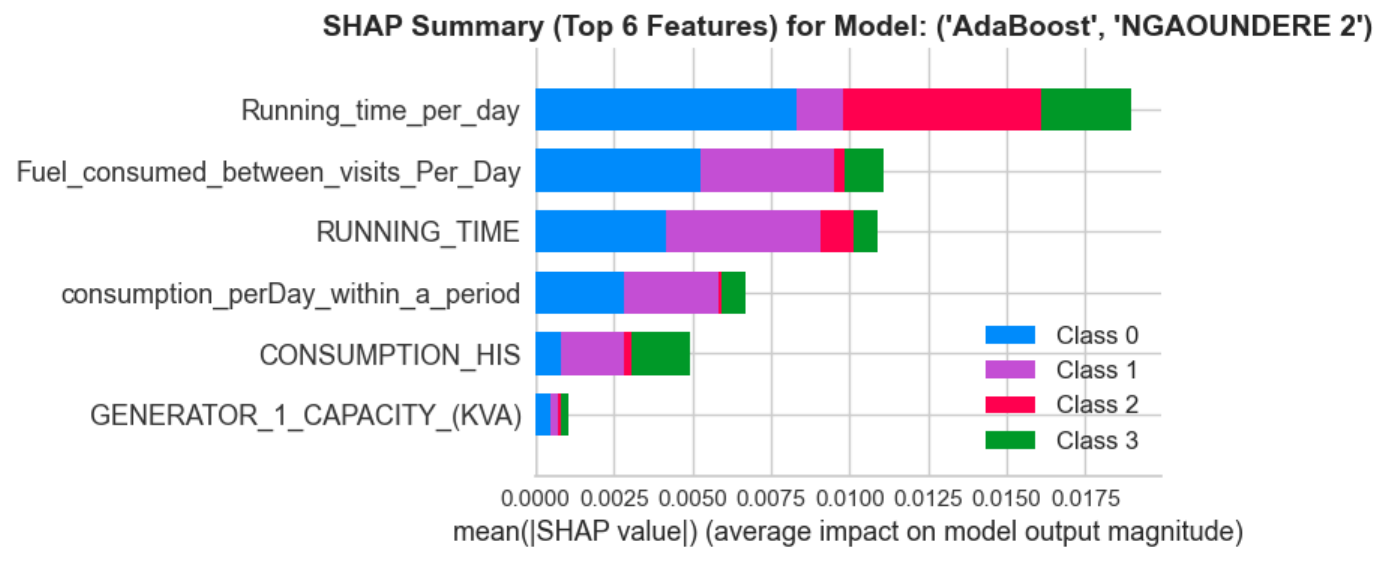}}
\caption{AdaBoost NGAOUNDERE 2}
\label{fig:SHAP_AdaBoost_NGAOUNDERE2}
\end{subfigure}
\hfill
\begin{subfigure}[b]{0.48\textwidth}
\centering
\fbox{\includegraphics[width=\textwidth]{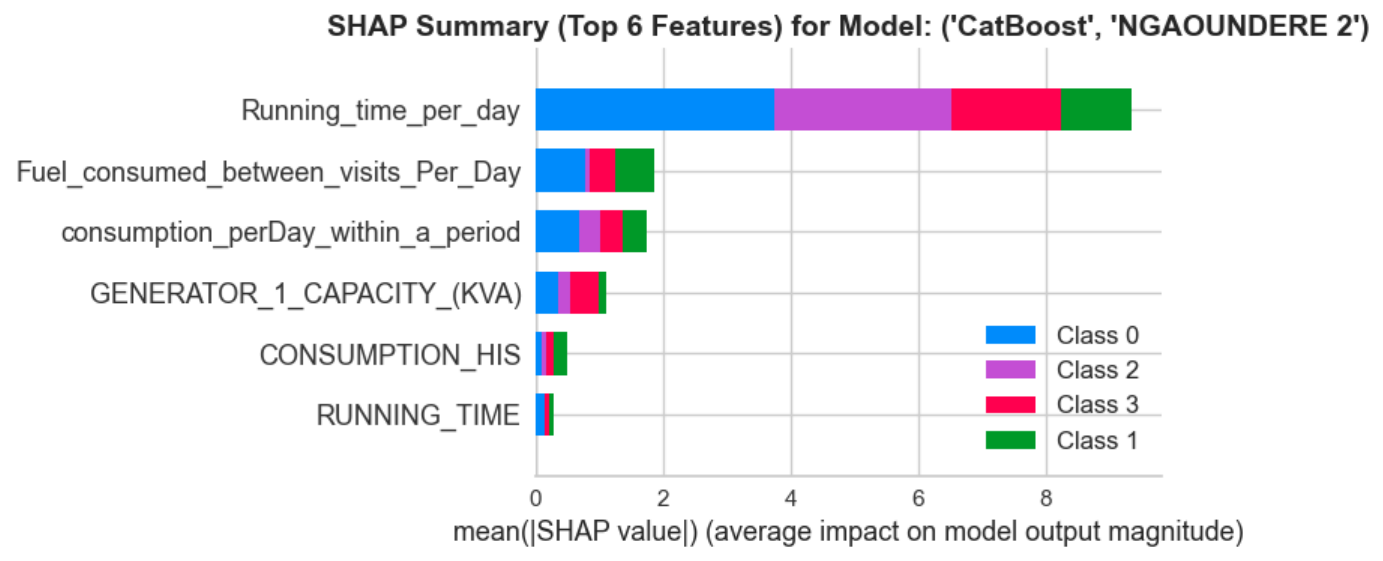}}
\caption{CatBoost NGAOUNDERE 2}
\label{fig:SHAP_CatBoost_NGAOUNDERE2}
\end{subfigure}
\vspace{0.1cm}
\begin{subfigure}[b]{0.48\textwidth}
\centering
\fbox{\includegraphics[width=\textwidth]{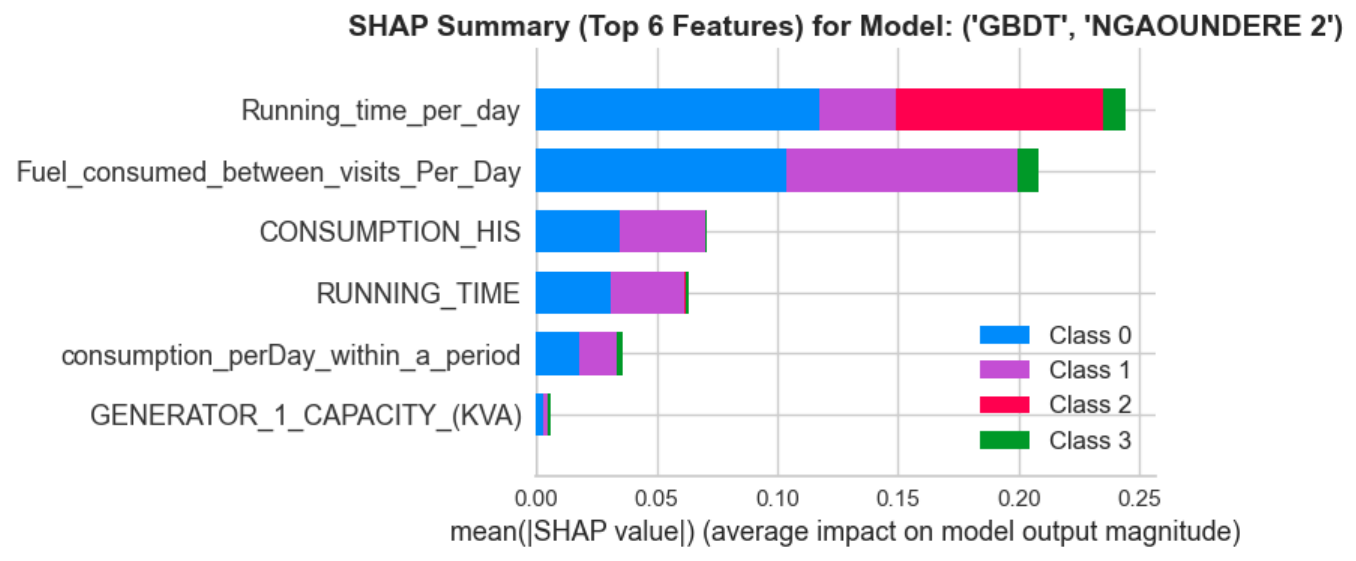}}
\caption{GBDT NGAOUNDERE 2}
\label{fig:SHAP_GBDT_NGAOUNDERE2}
\end{subfigure}
\hfill
\begin{subfigure}[b]{0.48\textwidth}
\centering
\fbox{\includegraphics[width=\textwidth]{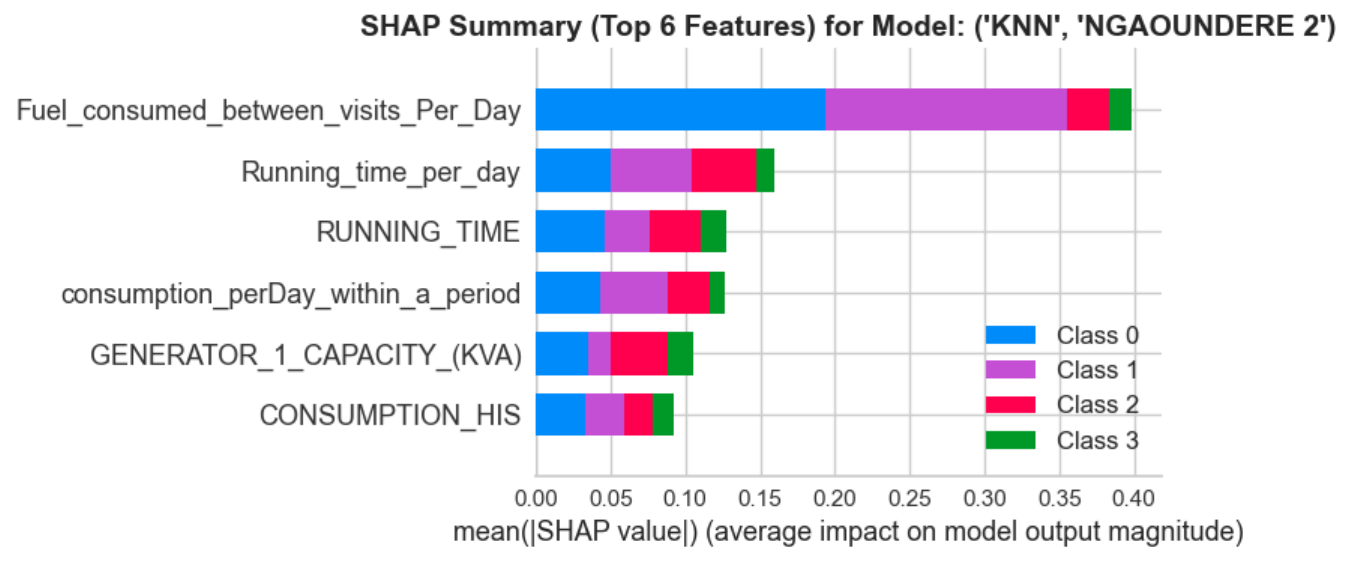}}
\caption{KNN NGAOUNDERE 2}
\label{fig:SHAP_KNN_NGAOUNDERE2}
\end{subfigure}
\vspace{0.1cm}
\begin{subfigure}[b]{0.48\textwidth}
\centering
\fbox{\includegraphics[width=\textwidth]{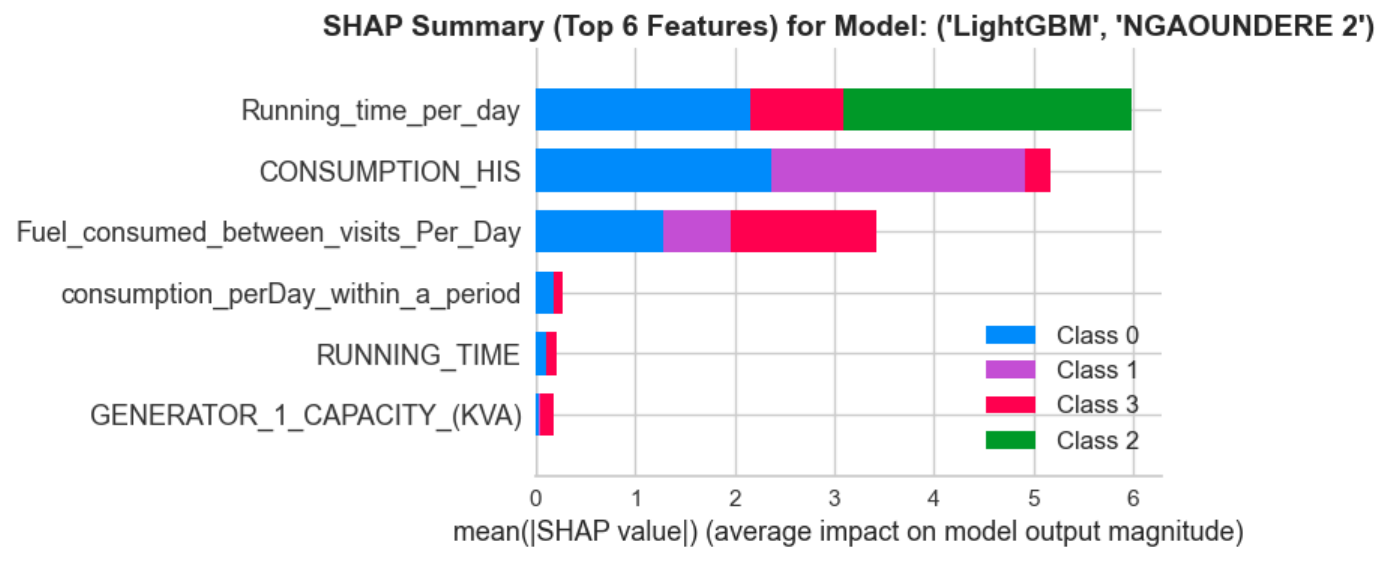}}
\caption{LightGBM NGAOUNDERE 2}
\label{fig:SHAP_LightGBM_NGAOUNDERE2}
\end{subfigure}
\hfill
\begin{subfigure}[b]{0.48\textwidth}
\centering
\fbox{\includegraphics[width=\textwidth]{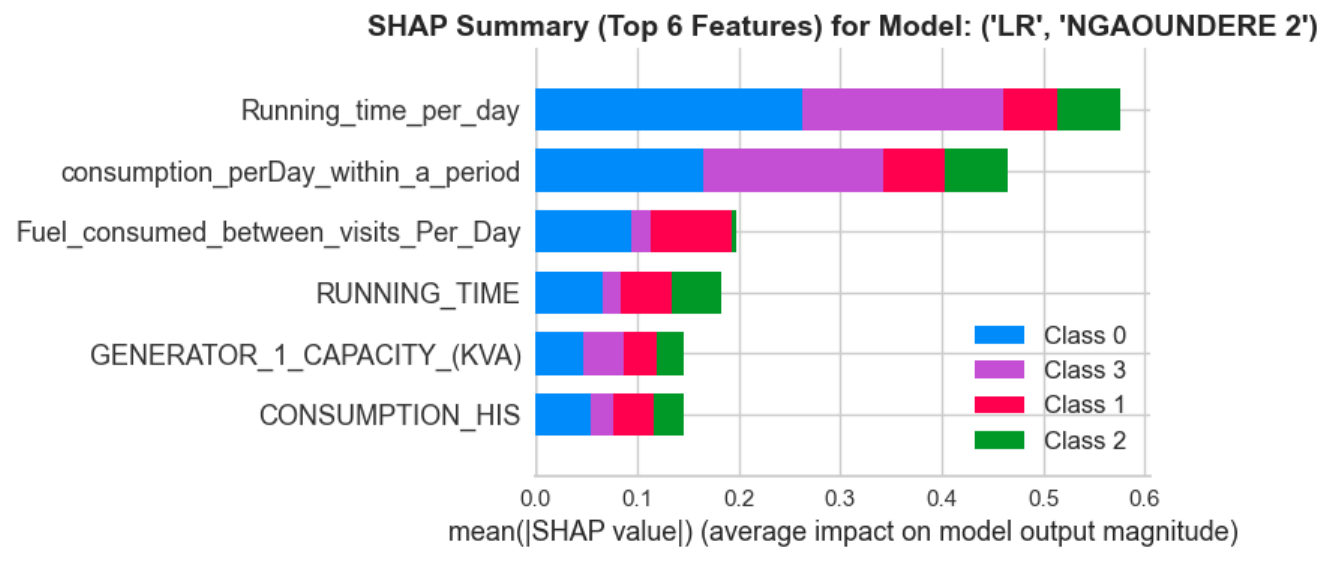}}
\caption{LR NGAOUNDERE 2}
\label{fig:SHAP_LR_NGAOUNDERE2}
\end{subfigure}
\vspace{0.1cm}
\begin{subfigure}[b]{0.48\textwidth}
\centering
\fbox{\includegraphics[width=\textwidth]{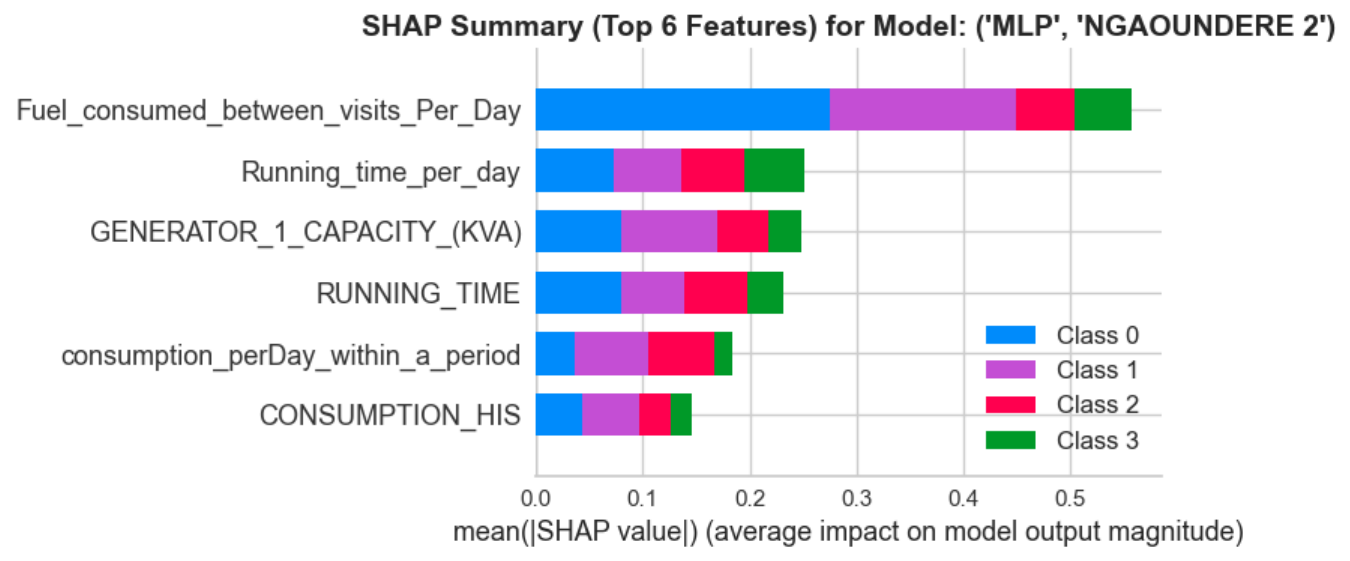}}
\caption{MLP NGAOUNDERE 2}
\label{fig:SHAP_MLP_NGAOUNDERE2}
\end{subfigure}
\hfill
\begin{subfigure}[b]{0.48\textwidth}
\centering
\fbox{\includegraphics[width=\textwidth]{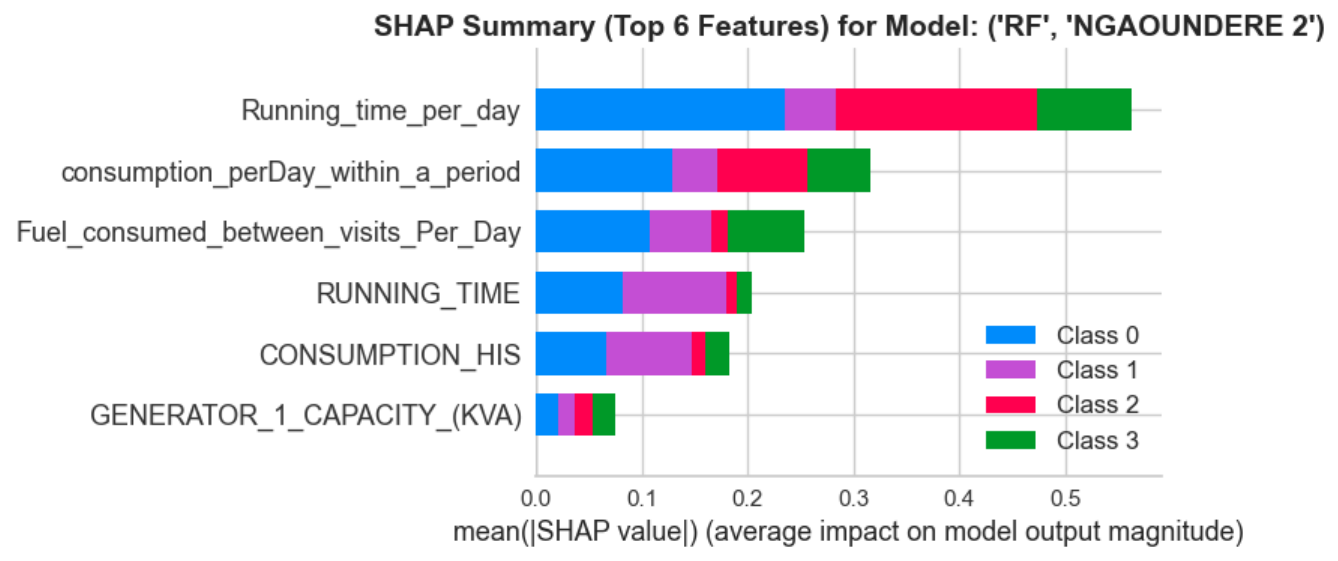}}
\caption{RF NGAOUNDERE 2}
\label{fig:SHAP_RF_NGAOUNDERE2}
\end{subfigure}
\vspace{0.1cm}
\begin{subfigure}[b]{0.48\textwidth}
\centering
\fbox{\includegraphics[width=\textwidth]{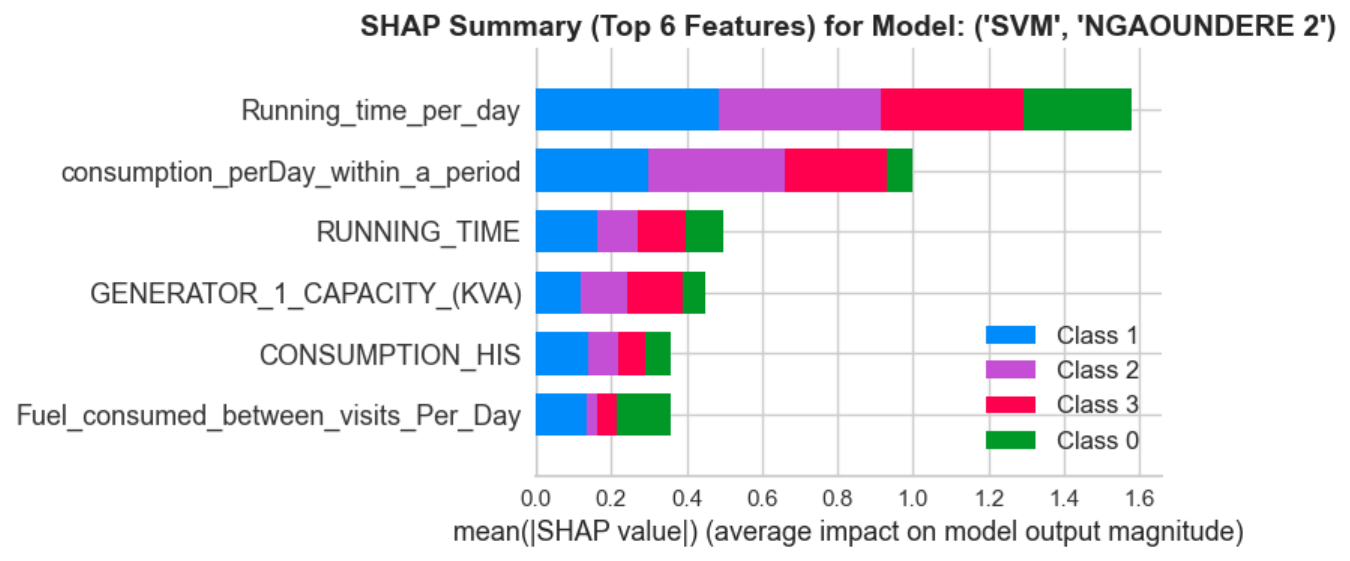}}
\caption{SVM NGAOUNDERE 2}
\label{fig:SHAP_SVM_NGAOUNDERE2}
\end{subfigure}
\hfill
\begin{subfigure}[b]{0.48\textwidth}
\centering
\fbox{\includegraphics[width=\textwidth]{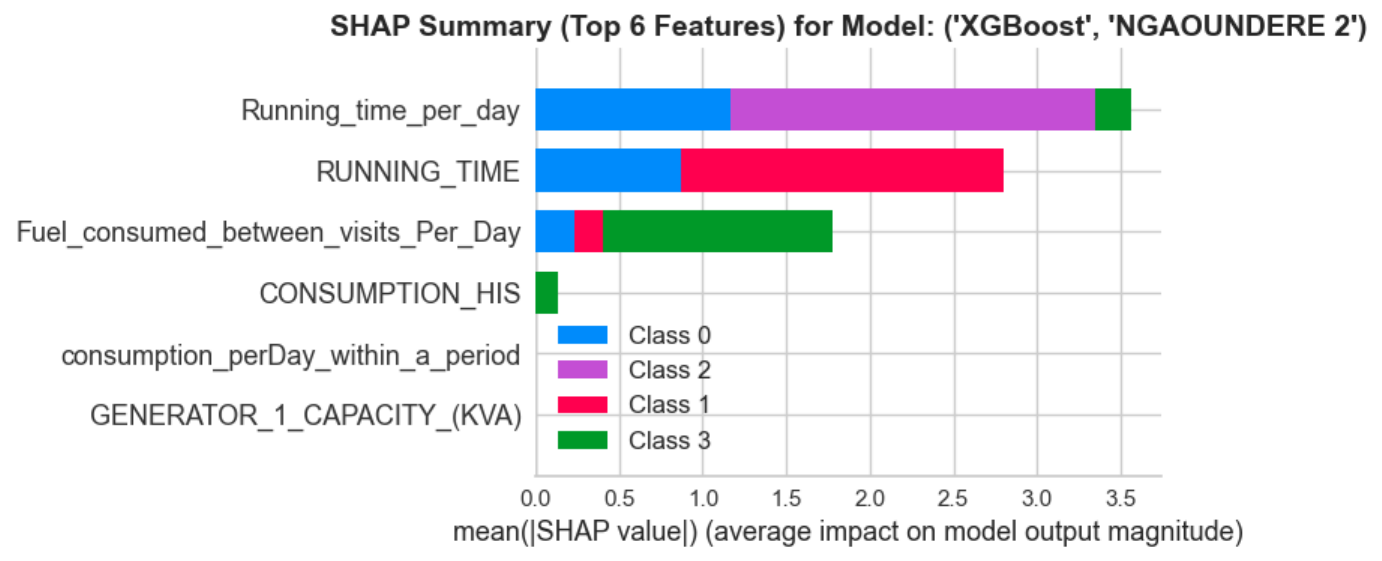}}
\caption{XGBoost NGAOUNDERE 2}
\label{fig:SHAP_XGBoost_NGAOUNDERE2}
\end{subfigure}
\caption{For all models trained on the NGAOUNDERE 2 Cluster data, these visualizations show mean absolute SHAP values per class, with \emph{running time per day} having the most impact on the results.}
\label{fig:SHAP_NGAOUNDERE2_comparison}
\end{figure}

\begin{figure}[htbp]
\centering
\begin{subfigure}[b]{0.48\textwidth}
\centering
\fbox{\includegraphics[width=\textwidth]{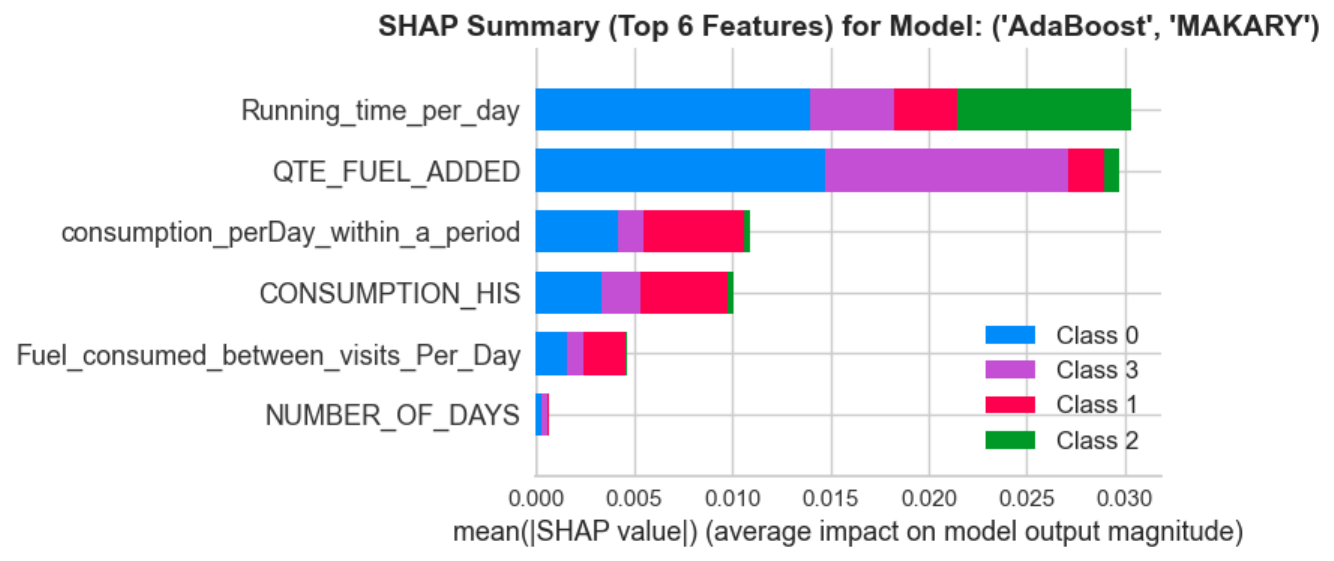}}
\caption{AdaBoost MAKARY}
\label{fig:SHAP_AdaBoost_MAKARY}
\end{subfigure}
\hfill
\begin{subfigure}[b]{0.48\textwidth}
\centering
\fbox{\includegraphics[width=\textwidth]{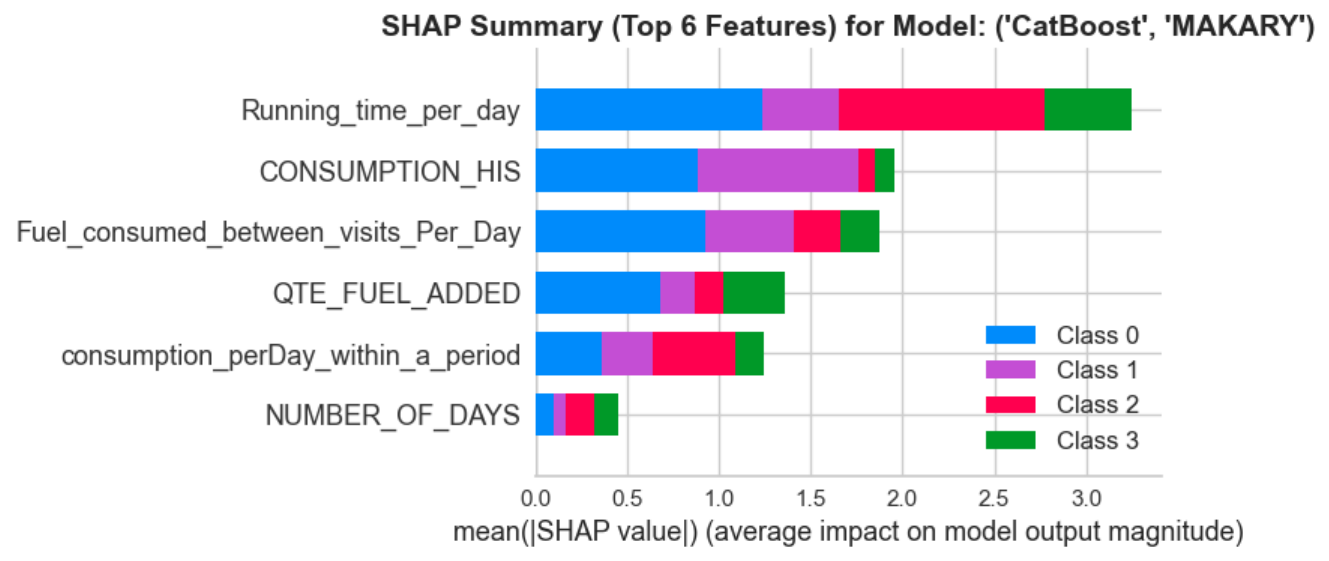}}
\caption{CatBoost MAKARY}
\label{fig:SHAP_CatBoost_MAKARY}
\end{subfigure}
\vspace{0.1cm}
\begin{subfigure}[b]{0.48\textwidth}
\centering
\fbox{\includegraphics[width=\textwidth]{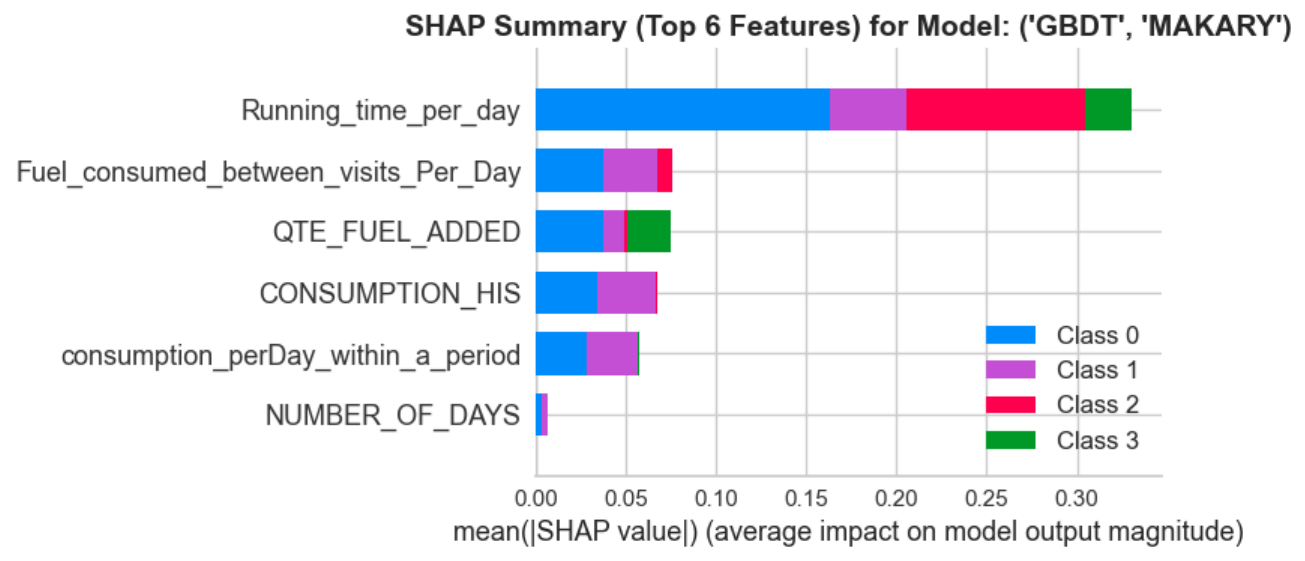}}
\caption{GBDT MAKARY}
\label{fig:SHAP_GBDT_MAKARY}
\end{subfigure}
\hfill
\begin{subfigure}[b]{0.48\textwidth}
\centering
\fbox{\includegraphics[width=\textwidth]{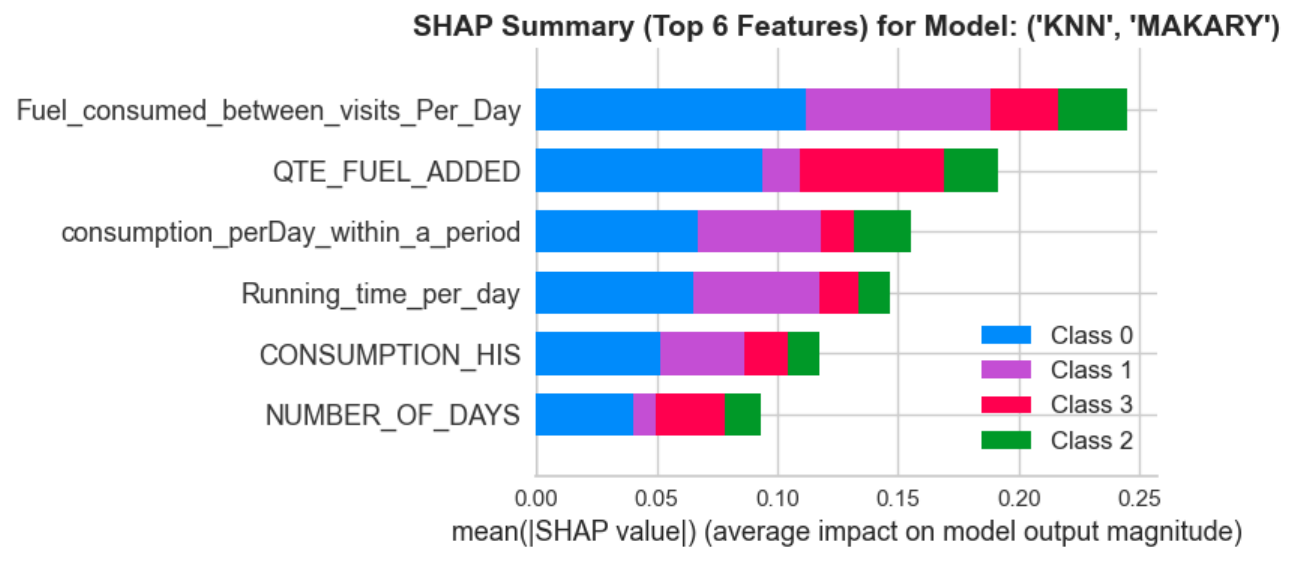}}
\caption{KNN MAKARY}
\label{fig:SHAP_KNN_MAKARY}
\end{subfigure}
\vspace{0.1cm}
\begin{subfigure}[b]{0.48\textwidth}
\centering
\fbox{\includegraphics[width=\textwidth]{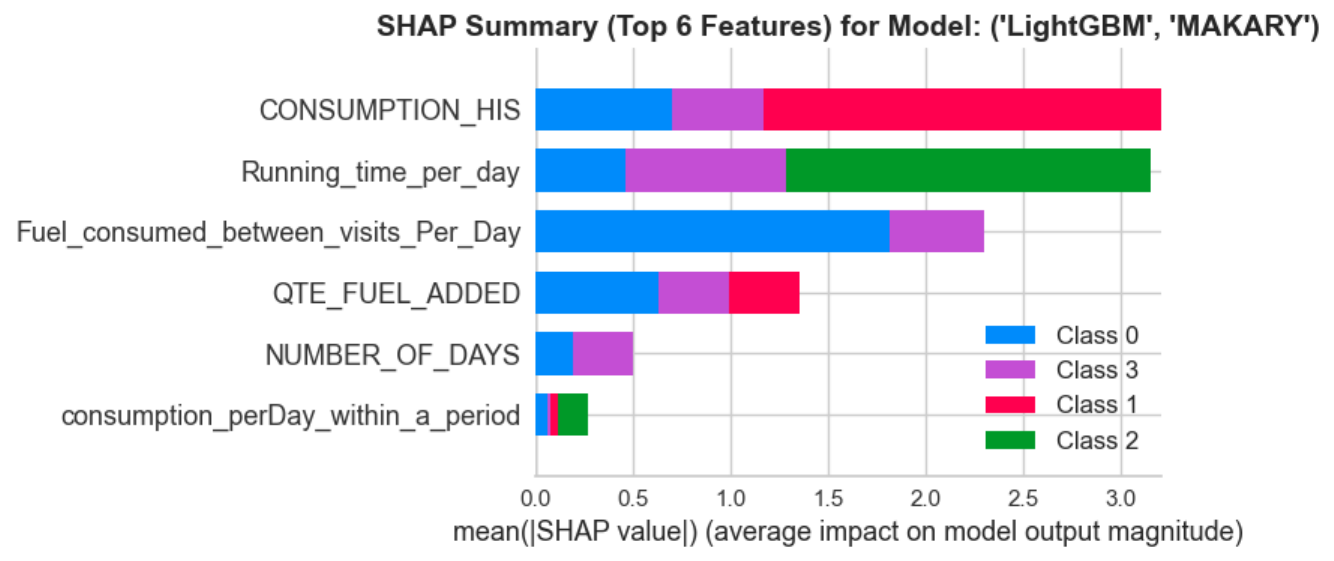}}
\caption{LightGBM MAKARY}
\label{fig:SHAP_LightGBM_MAKARY}
\end{subfigure}
\hfill
\begin{subfigure}[b]{0.48\textwidth}
\centering
\fbox{\includegraphics[width=\textwidth]{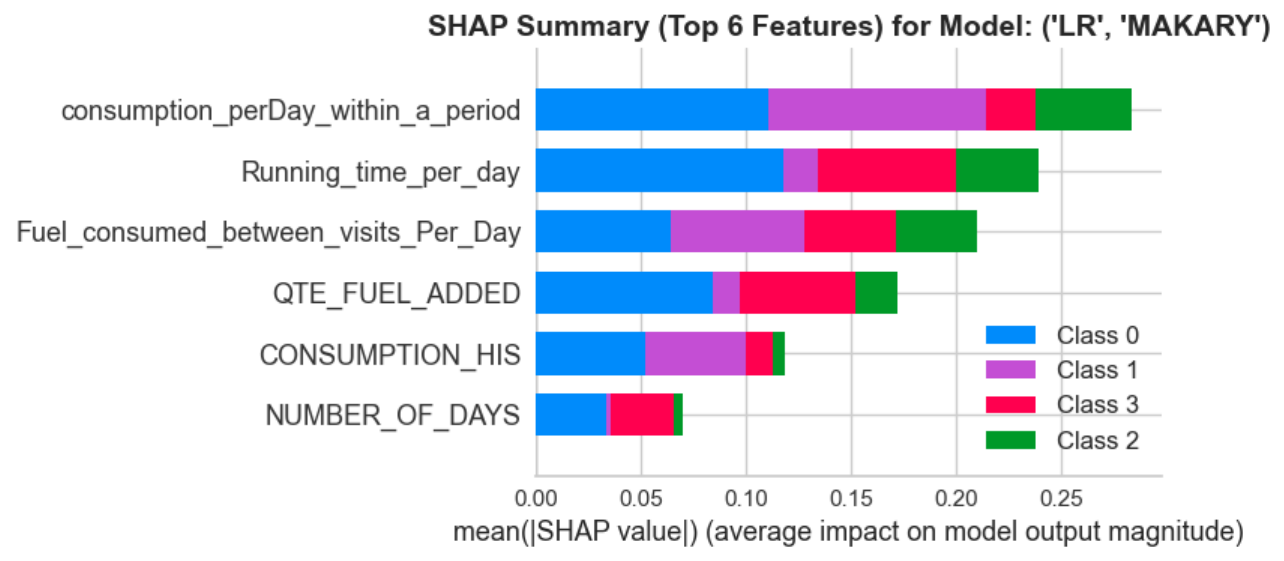}}
\caption{LR MAKARY}
\label{fig:SHAP_LR_MAKARY}
\end{subfigure}
\vspace{0.1cm}
\begin{subfigure}[b]{0.48\textwidth}
\centering
\fbox{\includegraphics[width=\textwidth]{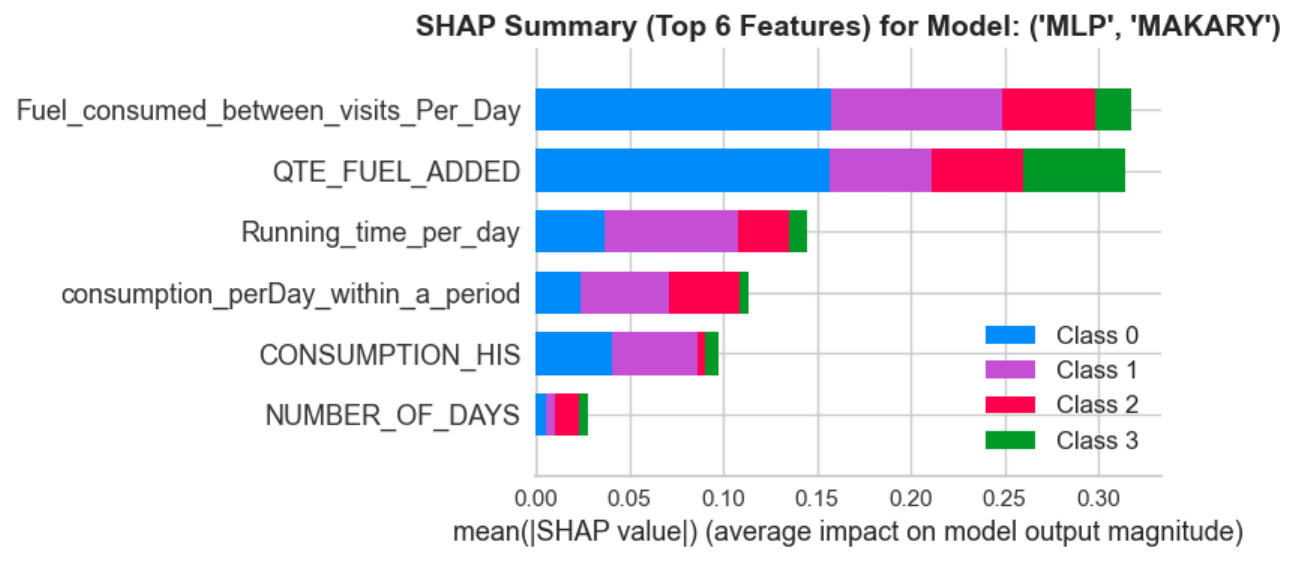}}
\caption{MLP MAKARY}
\label{fig:SHAP_MLP_MAKARY}
\end{subfigure}
\hfill
\begin{subfigure}[b]{0.48\textwidth}
\centering
\fbox{\includegraphics[width=\textwidth]{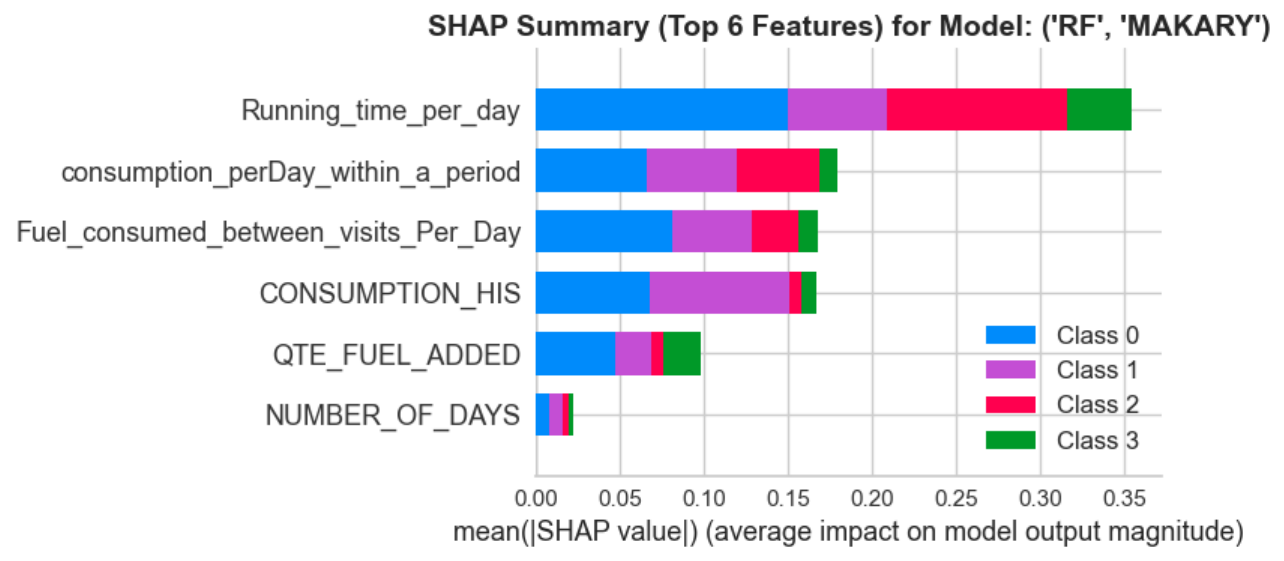}}
\caption{RF MAKARY}
\label{fig:SHAP_RF_MAKARY}
\end{subfigure}
\vspace{0.1cm}
\begin{subfigure}[b]{0.48\textwidth}
\centering
\fbox{\includegraphics[width=\textwidth]{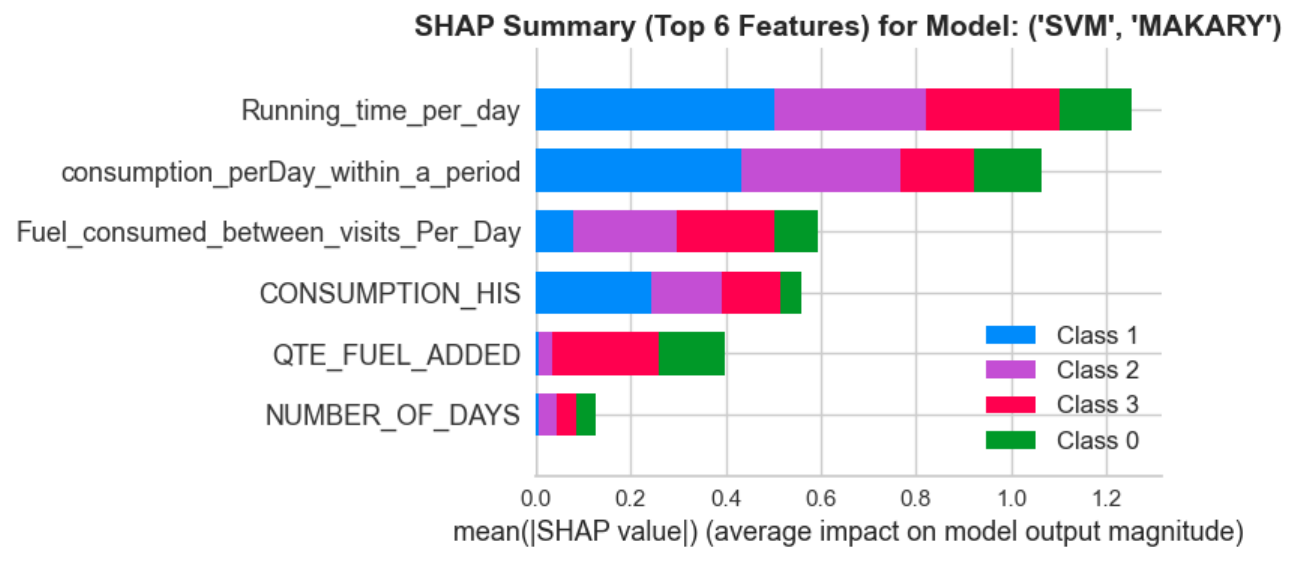}}
\caption{SVM MAKARY}
\label{fig:SHAP_SVM_MAKARY}
\end{subfigure}
\hfill
\begin{subfigure}[b]{0.48\textwidth}
\centering
\fbox{\includegraphics[width=\textwidth]{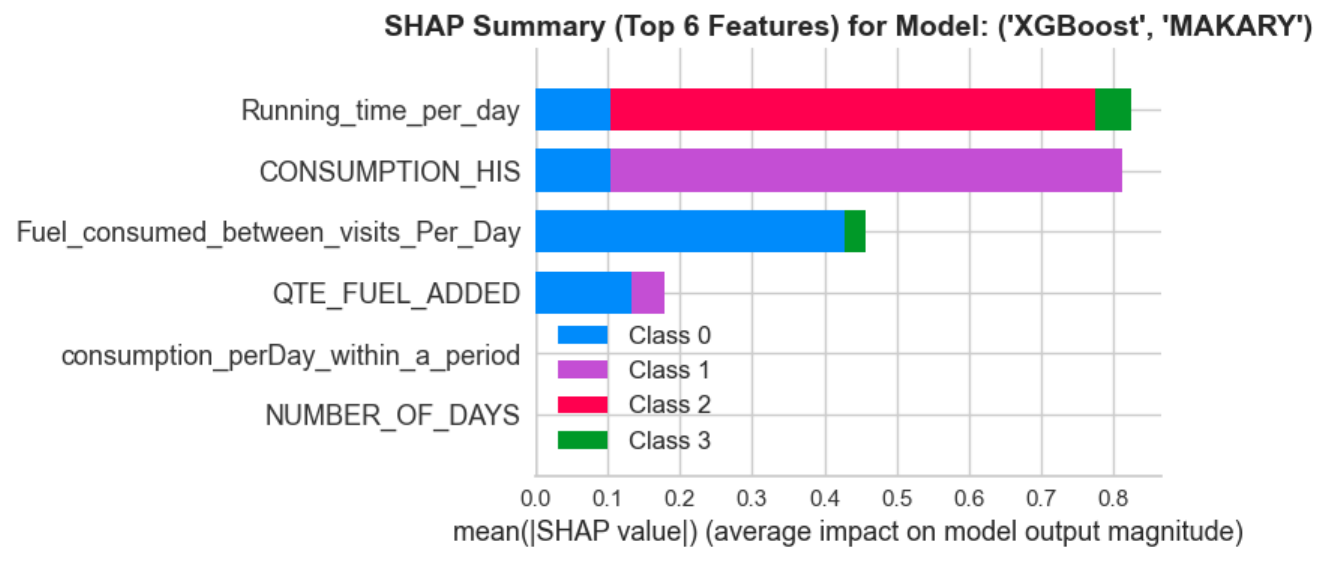}}
\caption{XGBoost MAKARY}
\label{fig:SHAP_XGBoost_MAKARY}
\end{subfigure}
\caption{For every model trained on the MAKARY Cluster data, these visualizations show mean absolute SHAP values per class, with \emph{running time per day} having the most impact on the results.}
\label{fig:SHAP_MAKARY_comparison}
\end{figure}

For the BANYO cluster, as shown in Figure~\ref{fig:SHAP_BANYO_comparison}, AdaBoost identifies \emph{running time per day} (0.016) as top, followed by \emph{previous fuel QTE} (0.006) and \emph{current hour meter GE1} (0.004), with contributions favoring Class 0. CatBoost amplifies \emph{running time per day} (6), \emph{fuel consumed between visits Per Day} (3), and \emph{consumption per day within a period} (2), reflecting context-dependent effects. GBDT on BANYO emphasizes \emph{running time per day} (0.2) and \emph{fuel consumed between visits per day} (0.05), with moderate class-specific influences.
In the NGAOUNDERE 2 cluster (Figure~\ref{fig:SHAP_NGAOUNDERE2_comparison}), AdaBoost ranks \emph{running time per day} (0.0175), \emph{fuel consumed between visits per day} (0.0075), and \emph{running time} (0.006), with higher values positively affecting Class 0. CatBoost shows \emph{running time per day} (8), \emph{fuel consumed between visits per day} (2), and \emph{consumption per day within a period} (1), highlighting operational patterns. GBDT features \emph{running time per day} (0.25), \emph{fuel consumed between visits per day} (0.15), and \emph{consumption HIS} (0.1), with balanced impacts across classes.
For the MAKARY cluster (Figure~\ref{fig:SHAP_MAKARY_comparison}), AdaBoost places \emph{running time per day} (0.03) first, followed by \emph{QTE fuel added} (0.025) and \emph{consumption per day within a period} (0.01), with influences on Classes 0 and 3. CatBoost leads with \emph{running time per day} (3), \emph{consumption HIS} (1.5), and \emph{fuel consumed between visits per day} (1), suggesting efficiency issues. GBDT on MAKARY prioritizes \emph{running time per day} (0.3), \emph{fuel consumed between visits per day} (0.1), and \emph{QTE fuel added} (0.05), with variability in class contributions.

Overall, \emph{running time per day} is the top feature across configurations, often boosting Class 0 while modulating others, indicating its role as a key predictor of normal operations versus anomalies. Ensemble models exhibit higher SHAP magnitudes, with global models providing balanced insights, BANYO emphasizing typical runtime and fuel metrics, NGAOUNDERE 2 highlighting capacity and consumption, and MAKARY focusing on added fuel and history, reflecting cluster-specific adaptations. These analyses affirm the importance of runtime and consumption features in multi-class anomaly detection, with Table~\ref{tab:feature_occurrences} confirming frequent occurrences (e.g., \emph{running time per day} 25 times, \emph{consumption per day within a period} 24 times), aiding precise, trustable maintenance strategies.

\begin{table}[h]
\centering
\caption{\emph{Running time per day} and \emph{consumption per day within a period} are frequently the most impact features across clusters, indicating they may be strong predictors of fuel consumption behavior.}
\label{tab:feature_occurrences}
\begin{tabular}{|l|c|}
\hline
\textbf{Feature} & \textbf{Occurrences} \\
\hline
Running time per day & 25 \\
consumption perDay within a period & 24 \\
RUNNING TIME & 16 \\
CONSUMPTION HIS & 14 \\
Fuel consumed between visits Per Day & 14 \\
GENERATOR 1 CAPACITY (KVA) & 13 \\
Fuel consumed between visits & 10 \\
CONSUMPTION RATE & 7 \\
Maximum consumption perDay & 7 \\
NUMBER OF DAYS & 6 \\
PREVIOUS FUEL QTE & 4 \\
QTE FUEL ADDED & 4 \\
QTE FUEL FOUND & 2 \\
TOTALE QTE LEFT & 2 \\
\hline
\end{tabular}
\end{table}

\subsection{Bias and Fairness Evaluation}
\label{subsec:bais_fairness_evaluation}

The bias and fairness of the models are assessed using the DIR \cite{zanna2024enhancing}, which compares the anomaly detection rates between minority and majority groups across various models in both global and cluster-specific configurations. The DIR is calculated as the ratio of the anomaly detection rate in the minority cluster to that in the majority clusters (average of others), with values between 0.8 and 1.25 indicating fairness, and 1.0 representing perfect fairness.

\begin{figure}[H]
\centering
\includegraphics[width=.75\linewidth]{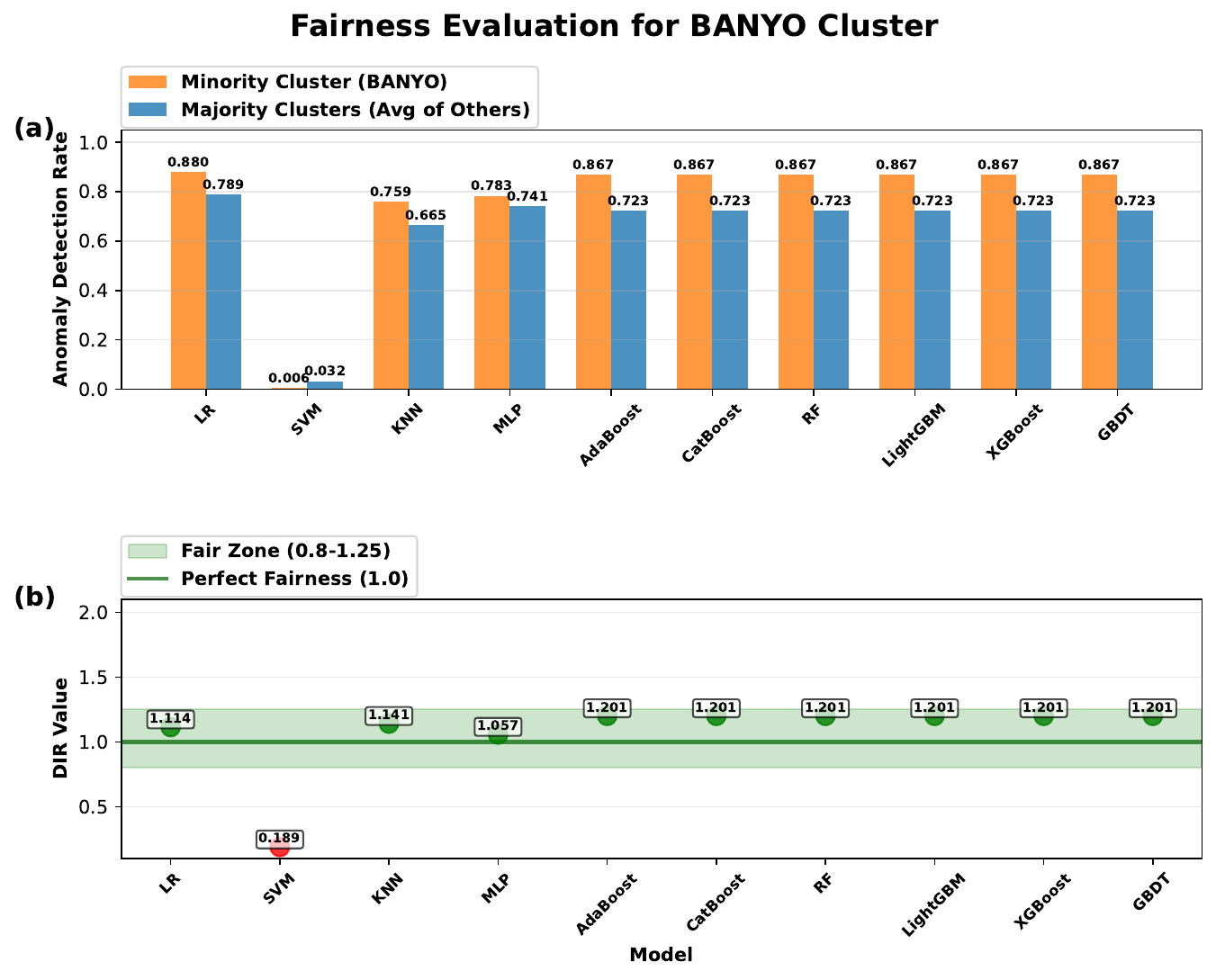}
\caption{Fairness evaluation for the BANYO cluster, showing high detection rates for ensemble models such as AdaBoost and XGBoost, but significant bias (DIR) for all models except SVM.}
\label{fig:fairness_eval_BANYO}
\end{figure}

For the BANYO cluster treated as minority, as shown in Figure~\ref{fig:fairness_eval_BANYO}, the anomaly detection rates for minority vs. majority are: LR (0.880 vs. 0.789, DIR 1.114), SVM (0.006 vs. 0.032, DIR 0.189), KNN (0.759 vs. 0.665, DIR 1.141), MLP (0.783 vs. 0.741, DIR 1.057), and ensembles like AdaBoost to GBDT (0.867 vs. 0.723, DIR 1.201). Most models, except SVM, fall within the fair zone, indicating equitable performance.

\begin{figure}[H]
\centering
\includegraphics[width=.75\linewidth]{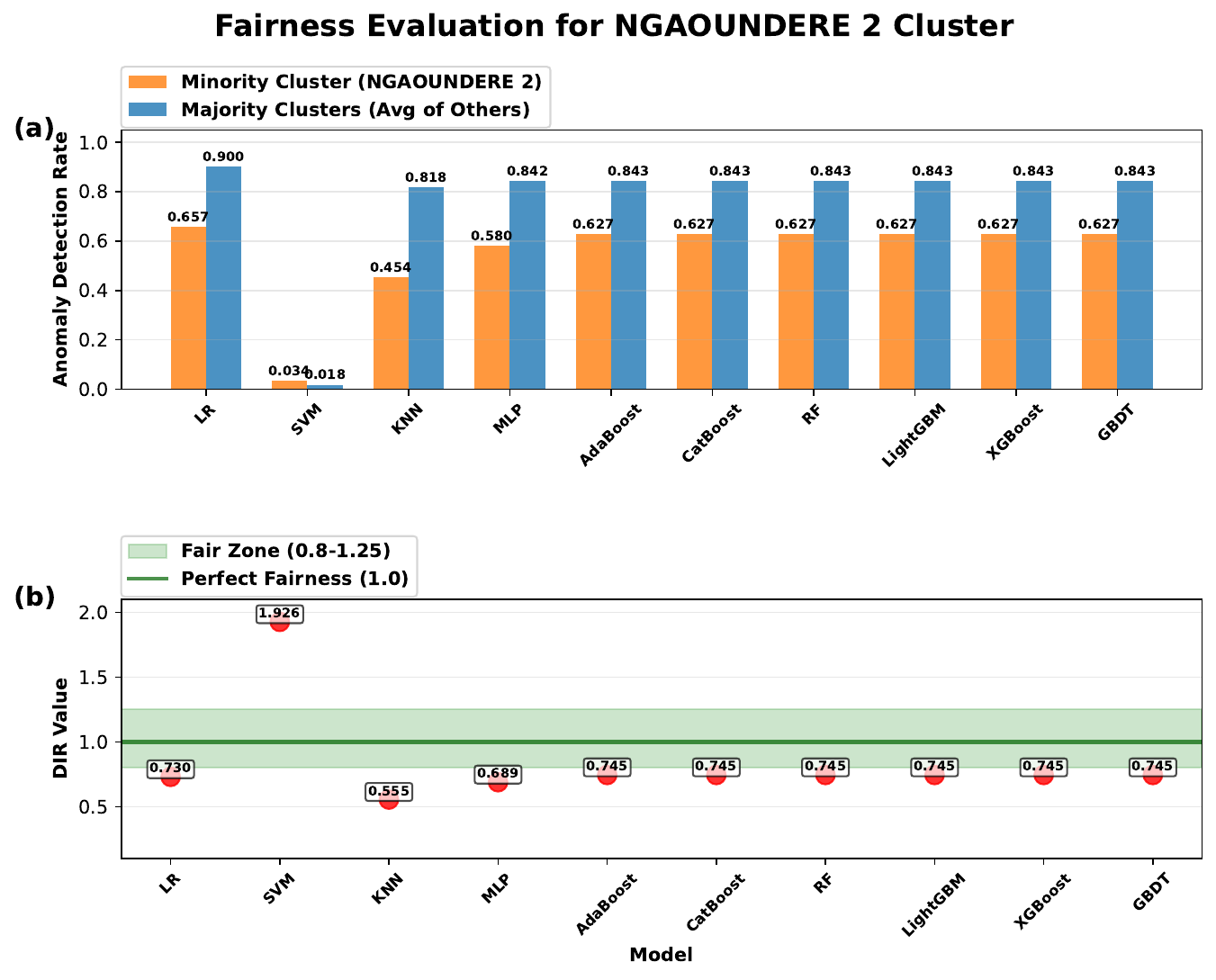}
\caption{Performance disparity for the NGAOUNDERE 2 cluster, where ensemble models such as CatBoost and GBDT show high detection, while SVM, KNN, and MLP show less anomaly detection. And all models are some how far from  fair zone}
\label{fig:fairness_eval_NGAOUNDERE2}
\end{figure}

For the NGAOUNDERE 2 cluster as minority, Figure~\ref{fig:fairness_eval_NGAOUNDERE2} shows rates: LR (0.657 vs. 0.900, DIR 0.730), SVM (0.034 vs. 0.018, DIR 1.926), KNN (0.454 vs. 0.818, DIR 0.555), MLP (0.580 vs. 0.842, DIR 0.689), and ensembles (0.627 vs. 0.843, DIR 0.745). Here, SVM exceeds 1.25 (bias toward minority), while others are below 0.8 (bias against minority), suggesting potential adjustments for fairness in transitional data.

\begin{figure}[H]
\centering
\includegraphics[width=.75\linewidth]{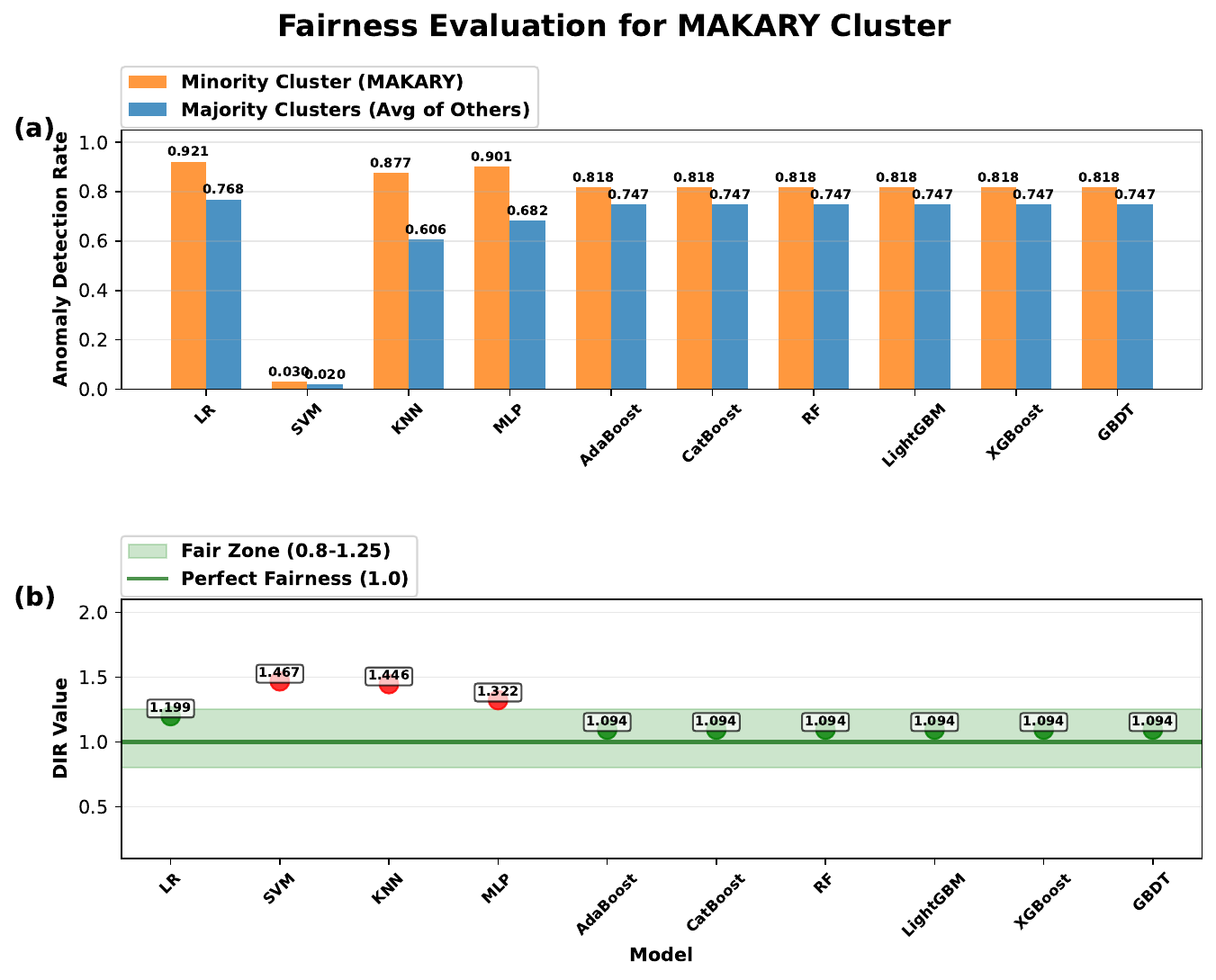}
\caption{Performance and fairness evaluation for the MAKARY cluster, showing high detection rates for ensemble models such as AdaBoost and RF, but significant bias (DIR) with most models favoring the minority cluster.}
\label{fig:fairness_eval_MAKARY}
\end{figure}

For the MAKARY cluster as minority, Figure~\ref{fig:fairness_eval_MAKARY} presents rates: LR (0.921 vs. 0.768, DIR 1.199), SVM (0.030 vs. 0.020, DIR 1.467), KNN (0.877 vs. 0.606, DIR 1.446), MLP (0.901 vs. 0.682, DIR 1.322), and ensembles (0.818 vs. 0.747, DIR 1.094). Ensembles and LR are fair, but SVM, KNN, and MLP exceed 1.25, indicating bias toward the minority in outlier scenarios.

\begin{figure}[H]
\centering
\includegraphics[width=.75\linewidth]{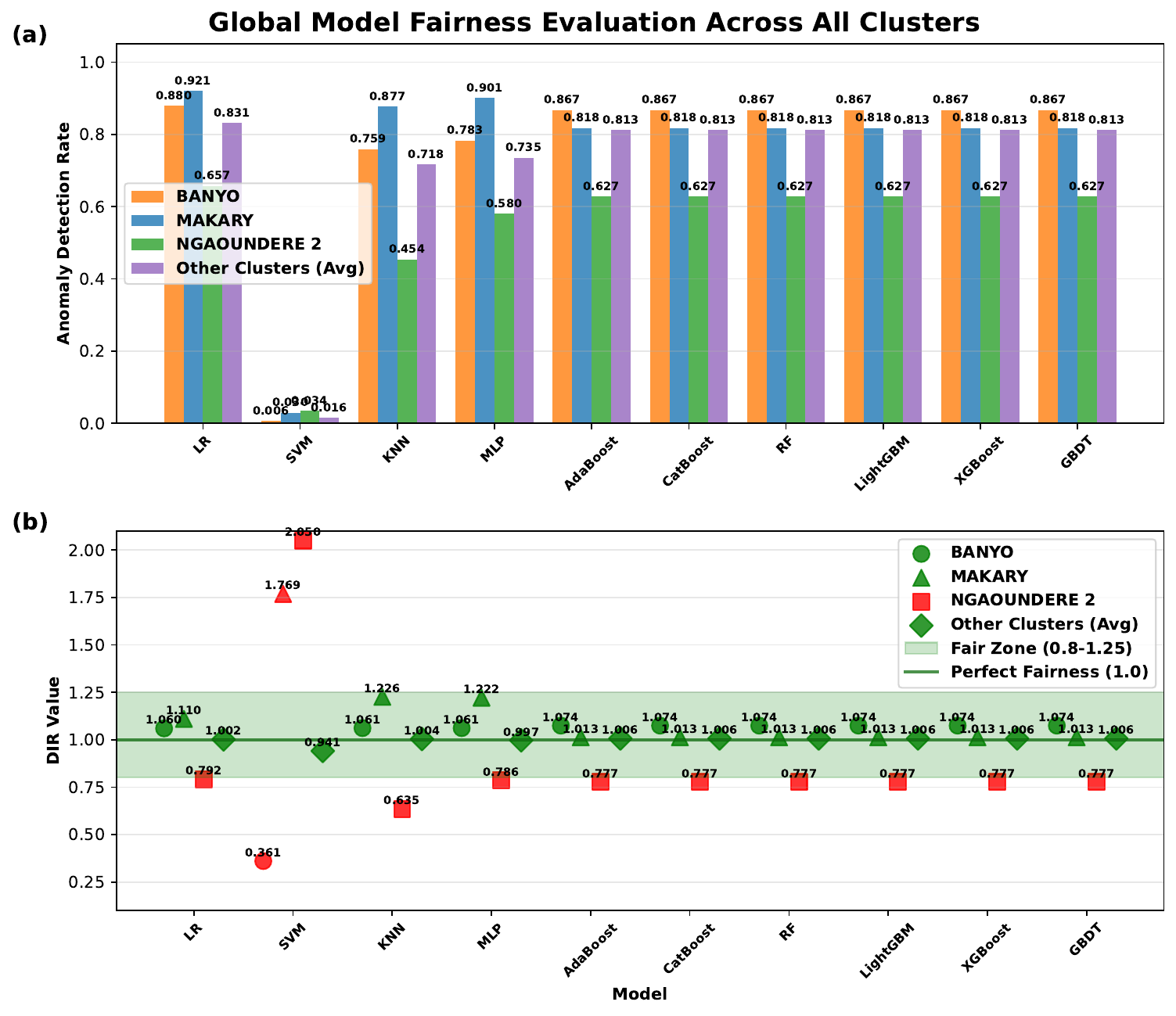}
\caption{Global model fairness evaluation showing inconsistent performance across clusters, with high bias against NGAOUNDERE 2 and in favor of MAKARY, while performance for BANYO and other clusters is generally within the fair zone.}
\label{fig:fairness_eval_global}
\end{figure}

In the global evaluation across all clusters, as depicted in Figure~\ref{fig:fairness_eval_global}, anomaly detection rates vary by cluster representativeness, with DIR values for BANYO (e.g., LR 1.060, ensembles ~1.074), MAKARY (LR 1.110, ensembles ~1.013), and NGAOUNDERE 2 (LR 0.792, ensembles ~0.777). Most DIRs fall within or near the fair zone for ensembles, confirming overall equitable performance, though base models show more variability.

These results highlight that ensemble models generally achieve better fairness across configurations, supporting their use in diverse operational contexts while identifying areas for bias mitigation in specific clusters.


\subsection{Decision Boundary Analysis}
\label{subsec:decision_boundary_analysis}

The decision boundaries of various ML models for a 4-class anomaly detection problem are visualized using t-SNE projections \citep{maaten2008visualizing}, with each subplot representing a different model. Colored regions indicate decision boundaries for each class, and scattered points represent the data points colored by their true labels. This analysis focuses on the global dataset and the three representative clusters: BANYO, NGAOUNDERE 2, and MAKARY, highlighting how boundaries adapt to typical, transitional, and outlier data distributions.

\begin{figure}[H]
\centering
\includegraphics[width=.75\linewidth]{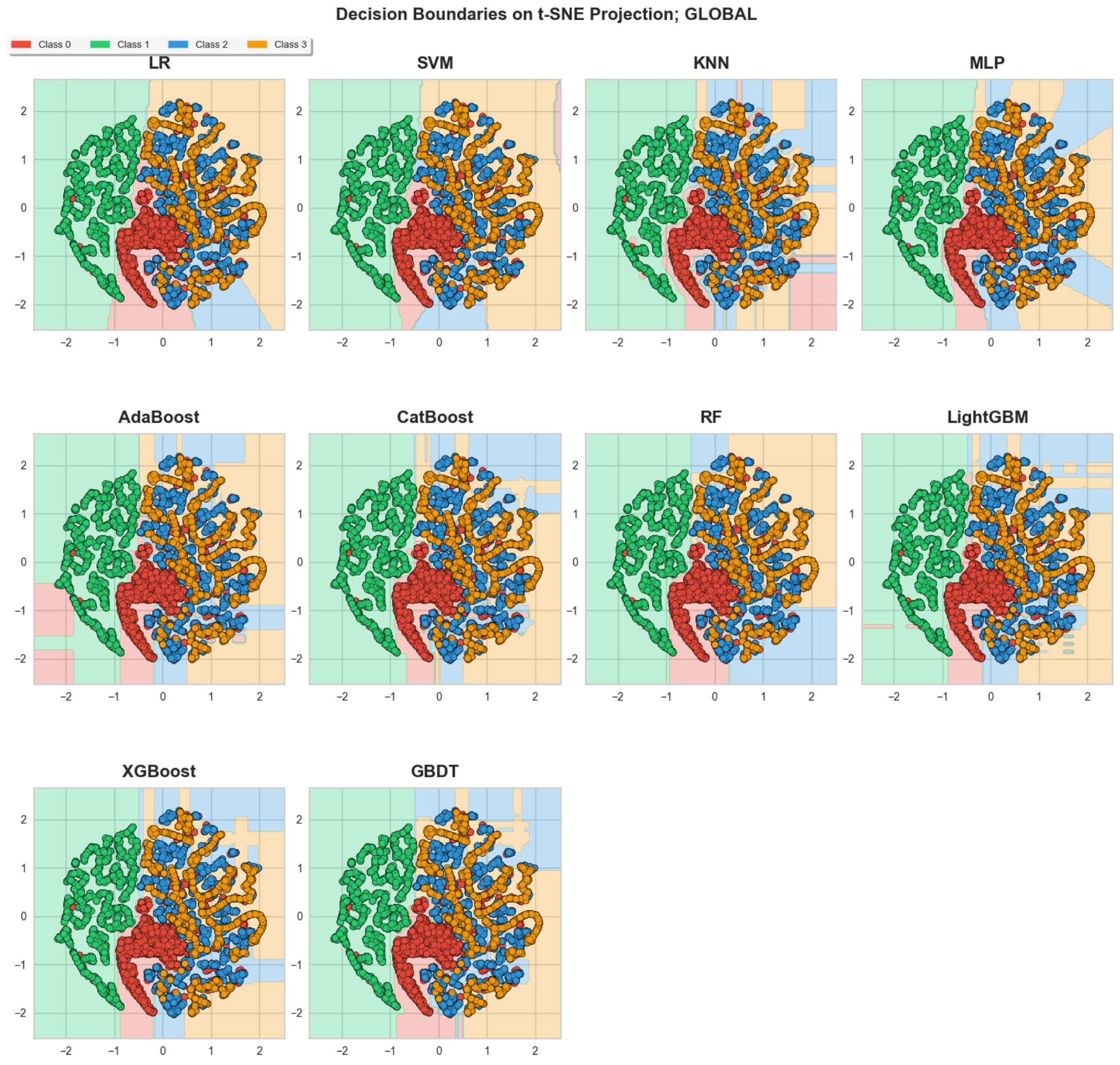}
\caption{Decision boundaries of various models, showing tree-based methods (XGBoost, LightGBM) capturing complex, local patterns while linear models (SVM, LR) form smoother, more generalized separations.}
\label{fig:DecisionBoundaries_Global}
\end{figure}

For the global dataset, as illustrated in Figure~\ref{fig:DecisionBoundaries_Global}, model accuracies are: LR (0.837), SVM (0.837), KNN (1.000), MLP (0.837), AdaBoost (0.841), CatBoost (0.859), RF (0.850), LightGBM (0.885), XGBoost (0.849), and GBDT (0.866). The t-SNE projection shows data points forming a curved cluster, with Class 0 (red) dominant in the main arc, Class 2 (green) interspersed, and Class 1 (blue) and Class 3 (purple) as sparse outliers. High-accuracy models like KNN (1.000) and LightGBM (0.885) display sharp, well-defined boundaries with minimal overlap, effectively separating the dense Class 0 from outliers. In contrast, lower-accuracy models like LR and SVM (0.837) exhibit more blurred boundaries, with significant overlap between Classes 0 and 2, indicating challenges in capturing non-linear structures.

\begin{figure}[H]
\centering
\includegraphics[width=0.75\linewidth]{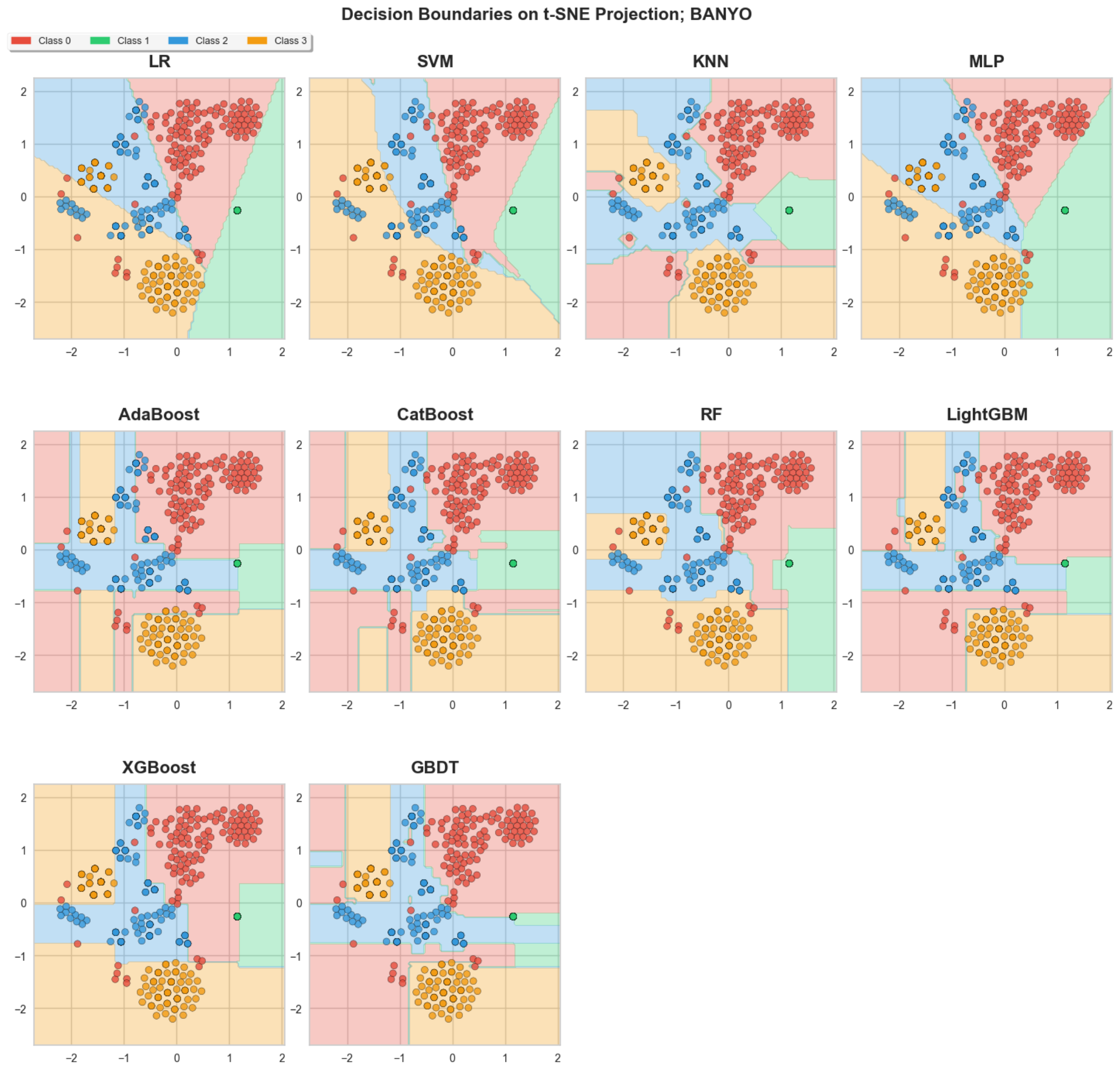}
\caption{Model decision boundaries on the BANYO cluster vary from smooth (LR, SVM) to highly tree-based models, with misalignments in mixed areas indicating higher localized misclassification risk than the global model.}
\label{fig:DecisionBoundaries_Banyo}
\end{figure}

\begin{figure}[H]
\centering
\includegraphics[width=.75\linewidth]{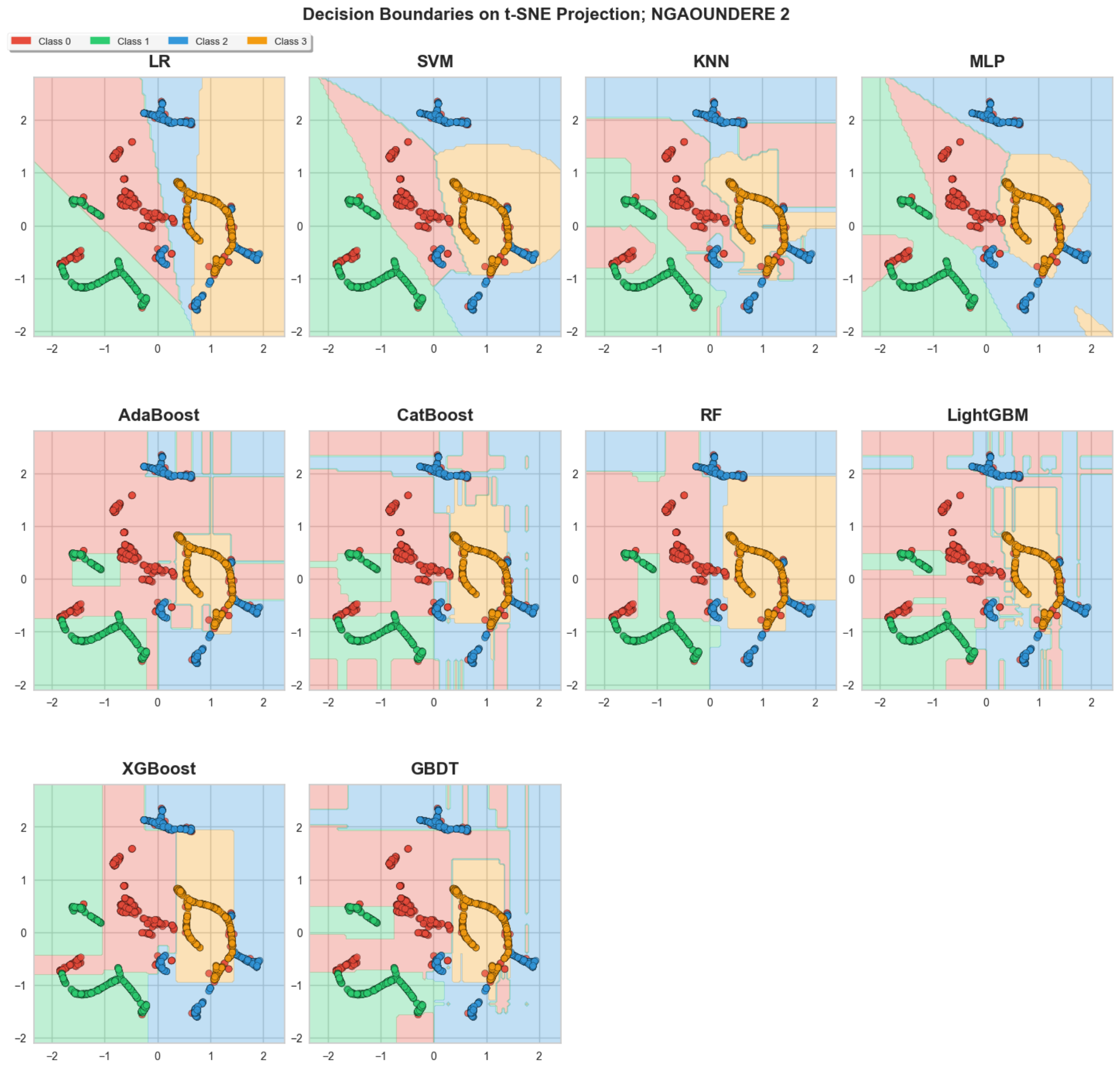}
\caption{Models adapt variably to the cluster's sparse, linear distributions, from smooth (LR, SVM) to fragmented (tree-based) boundaries, with overlaps in intermixed areas indicating heightened misclassification risks.}
\label{fig:DecisionBoundaries_Ngaoundere2}
\end{figure}

\begin{figure}[H]
\centering
\includegraphics[width=.75\linewidth]{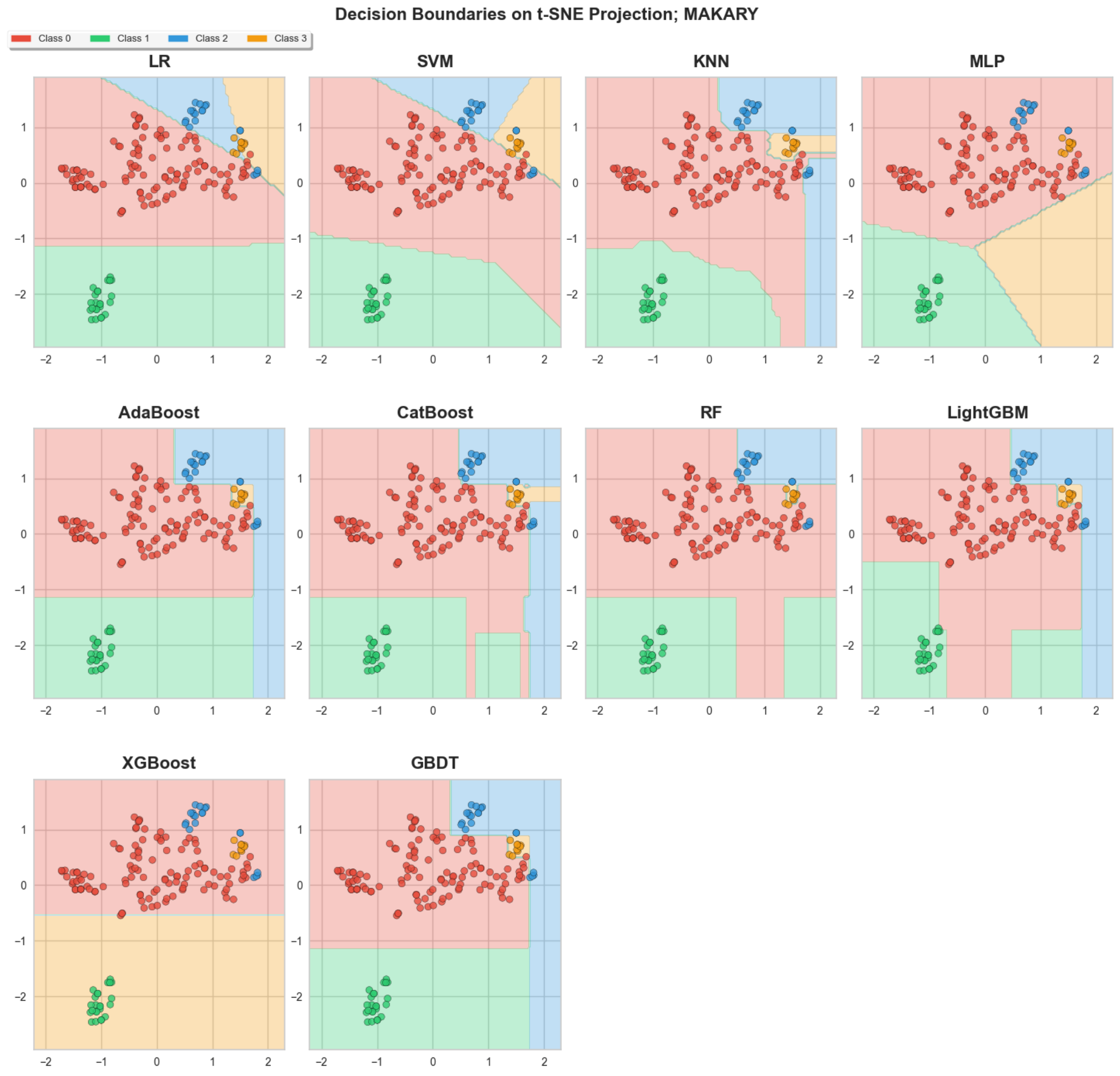}
\caption{Models adapt to the cluster's imbalanced, bipolar data with varying boundary complexity, from smooth to highly fragmented, resulting in minority class overlaps and heightened misclassification risks.}
\label{fig:DecisionBoundaries_Makary}
\end{figure}

For the BANYO cluster, shown in Figure~\ref{fig:DecisionBoundaries_Banyo}, accuracies are: LR (0.880), SVM (0.880), KNN (1.000), MLP (0.880), AdaBoost (0.980), CatBoost (0.940), RF (0.892), LightGBM (0.952), XGBoost (0.880), and GBDT (1.000). The projection reveals points aligned in a linear fashion, with Class 0 (red) forming the core chain, Class 2 (green) and Class 1 (blue) clustered at ends, and Class 3 (purple) scattered. Models like KNN and GBDT (1.000) achieve perfect separation with clean linear boundaries, reflecting robust handling of typical data patterns. Models with moderate accuracy, such as RF (0.892), show some overlap between Classes 0 and 2, suggesting sensitivity to noise in representative data.
For the NGAOUNDERE 2 cluster, Figure~\ref{fig:DecisionBoundaries_Ngaoundere2} presents boundaries with varying accuracies, where ensemble models like AdaBoost and CatBoost demonstrate smoother transitions between classes, aligning with its mid-level representativeness. The projection likely shows a mix of clustered and scattered points, with better-performing models minimizing overlap in transitional distributions.
For the MAKARY cluster, as depicted in Figure~\ref{fig:DecisionBoundaries_Makary}, the boundaries highlight challenges with outliers, with lower accuracies in base models due to irregular point distributions, while ensembles maintain clearer separations.

Overall, the visualizations underscore that ensemble models (e.g., GBDT, KNN) excel with distinct boundaries and higher accuracies across configurations, effectively capturing class structures in global and cluster-specific contexts. This supports their suitability for anomaly detection, where distinguishing sparse classes (1 and 3) from dominant ones (0 and 2) is critical.


{\color{black}
\section{Deployment in Practice}
\label{sec:deployment_in_practice}

While experimental results demonstrate the effectiveness of the proposed framework, 
its ultimate value lies in real-world deployment for continuous monitoring of power plants. 
This section describes how the trained models can be operationalized in production to 
support anomaly detection in telecom generator systems.

\subsection{System Integration}
\label{subsec:system_integration}

The trained ensemble models (e.g., LightGBM, XGBoost, CatBoost) can be exported in 
standard formats such as \texttt{joblib} or \texttt{pickle}, and subsequently wrapped 
within lightweight RESTful APIs. Integration with existing Supervisory Control and Data 
Acquisition (SCADA) systems enables seamless communication between sensors, databases, 
and the anomaly detection engine. In practice, generator telemetry (fuel consumption, 
runtime, refueling events) is streamed to a centralized server where the model 
continuously evaluates incoming data for anomalies.

\subsection{Real-Time Inference}
\label{subsec:realtime-inference}

To achieve real-time anomaly detection, the models can be deployed in containerized 
environments (e.g., Docker) orchestrated by Kubernetes. Each containerized service hosts 
a prediction engine that receives incoming records, processes them, and returns anomaly 
labels within milliseconds. Our experiments showed inference latency of less than 
0.001 seconds, confirming suitability for live monitoring. Anomalous events are flagged 
immediately and forwarded to operators via dashboards or mobile notifications, enabling 
timely intervention.

\subsection{Explainability in Action}
\label{subsec:explainability_in_action}

Interpretability remains crucial in production. SHAP values are computed for each 
prediction to highlight feature contributions. For example, an alert for excessive fuel 
consumption will be accompanied by SHAP explanations showing that ``consumption rate'' 
and ``runtime per day'' strongly influenced the anomaly decision. This transparency 
enhances operator trust and supports root-cause diagnosis.

\subsection{Scalability and Maintenance}
\label{subsec:scalability_and_maintenance}

The framework supports both global and cluster-specific models. In practice, global 
models can serve as default predictors, while specialized models are maintained for 
clusters with unique operational patterns. Periodic retraining is automated using 
scheduled jobs that incorporate new data, ensuring the models remain robust against 
concept drift and seasonal variations. Furthermore, fairness monitoring is included in 
the deployment pipeline: the DIR is computed periodically to 
ensure equitable performance across clusters.

\subsection{Demonstration Scenario}
\label{subsec:demonstration_scenario}

Consider a generator in the KOUSSERI cluster transmitting telemetry data: 
\texttt{runtime = 28 hours/day} and \texttt{fuel consumption = 220 L/day}. The deployed 
LightGBM service receives these features and classifies the record as a Class 2 anomaly 
(runtime exceeding 24 hours). Simultaneously, SHAP values indicate that ``runtime per 
day'' had the highest contribution to the prediction. The anomaly is logged in the 
monitoring dashboard, triggering an automated maintenance ticket for on-site inspection. 
This end-to-end pipeline illustrates how the framework operates in practice, 
transforming predictive models into actionable intelligence.
}

\section{Conclusion}
\label{sec:conclusion}

This work addressed the critical challenge of anomaly detection in distributed power plant monitoring, with a particular focus on diesel generators used in telecom infrastructure in Cameroon. We proposed a supervised ML framework that jointly optimizes \textit{performance}, \textit{interpretability}, and \textit{fairness}. By integrating ensemble models with advanced resampling strategies (SMOTE with Tomek Links and ENN), we mitigated severe class imbalance in the TeleInfra Ltd. dataset. The incorporation of SHAP values provided transparent feature attribution, highlighting consumption rate and runtime per day as dominant predictors, while fairness was ensured through the DIR to minimize regional bias. 

{\color{black}
Beyond experimental validation, we demonstrated how the proposed framework can be
deployed in practice. The models can be integrated into real-time monitoring systems
through containerized services, receiving telemetry streams from generators and
delivering anomaly predictions with millisecond-level latency. Explainability through
SHAP enhances operator trust, while fairness monitoring ensures equitable detection
across regional clusters. A demonstration scenario illustrated how runtime and fuel
consumption anomalies can be flagged in production, triggering timely maintenance
actions.
}

Our experimental results showed that ensemble methods, especially LightGBM and XGBoost, consistently outperformed baselines, achieving F1-scores above 0.95 and demonstrating strong generalization across clusters as confirmed by MMD analysis. These results, coupled with deployment feasibility, confirm the practical relevance of the framework for industrial anomaly detection

Despite these promising outcomes, several limitations remain. The dataset spans a
single year and limited clusters, restricting generalization to broader contexts. Current
deployment considerations focus on batch and streaming anomaly detection, but hybrid
deep learning approaches may further enhance temporal modeling. Future work will
explore larger and more diverse datasets, multi-year deployments, and fairness-aware
training objectives to improve equitable performance.

Overall, this study demonstrates that balancing performance, interpretability, and fairness is feasible in anomaly detection for power systems, paving the way for more trustworthy and sustainable AI solutions in critical infrastructure management.

\bibliography{sn-bibliography}

\end{document}